%% file: mnist_paper.tex
\pdfoutput=1
\documentclass[nohyperref]{article}

\input{preamble_packages}

\input{preamble_settings}

\begin{document}

\twocolumn[
\icmltitle{{Equivariance versus Augmentation for Spherical Images}}



\icmlsetsymbol{equal}{*}

\begin{icmlauthorlist}
    \icmlauthor{Jan E.\ Gerken}{chalmers,tuberlin,bifold}
    \icmlauthor{Oscar Carlsson}{chalmers}
    \icmlauthor{Hampus Linander}{gu}
    \icmlauthor{Fredrik Ohlsson}{ume}
    \icmlauthor{Christoffer Petersson}{zenseact,chalmers}
    \icmlauthor{Daniel Persson}{chalmers}
\end{icmlauthorlist}

\icmlaffiliation{chalmers}{Department of Mathematical Sciences, Chalmers University of Technology, Gothenburg, Sweden}
\icmlaffiliation{gu}{Department of Physics, University of Gothenburg, Gothenburg, Sweden}
\icmlaffiliation{tuberlin}{Machine Learning Group at Berlin Institute of Technology, Berlin, Germany}
\icmlaffiliation{bifold}{Berlin Institute for the Foundations of Learning and Data (BIFOLD), Berlin, Germany}
\icmlaffiliation{ume}{Department of Mathematics and Mathematical Statistics, Umeå University, Umeå, Sweden}
\icmlaffiliation{zenseact}{Zenseact, Gothenburg, Sweden}

%\icmlcorrespondingauthor{Jan Gerken}{gerken@tu-berlin.de}
\icmlcorrespondingauthor{Daniel Persson}{daniel.persson@chalmers.se}

\icmlkeywords{semantic segmentation, classification, cnn, data augmentation, spherical cnn}

\vskip 0.3in

]



\printAffiliationsAndNotice{}  % leave blank if no need to mention equal contribution

\begin{abstract}
We analyze the role of rotational equivariance in convolutional neural networks (CNNs) applied to spherical images. We compare the performance of the group equivariant networks known as S2CNNs and standard non-equivariant CNNs trained with an increasing amount of data augmentation. The chosen architectures can be considered baseline references for the respective design paradigms. Our models are trained and evaluated on single or multiple items from the MNIST or FashionMNIST dataset projected onto the sphere. For the task of image classification, which is inherently rotationally {\it invariant}, we find that by considerably increasing the amount of data augmentation and the size of the networks, it is possible for the standard CNNs to reach at least the same performance as the equivariant network. In contrast, for the inherently {\it equivariant} task of semantic segmentation, the non-equivariant networks are consistently outperformed by the equivariant networks with significantly fewer parameters. We also analyze and compare the inference latency and training times of the different networks, enabling detailed tradeoff considerations between equivariant architectures and data augmentation for practical problems. The equivariant spherical networks used in the experiments are available at \url{https://github.com/JanEGerken/sem_seg_s2cnn}.
\end{abstract}

\input{section_introduction}

\input{section_theory}

\input{section_spherical_case}

\input{section_conclusions}

{\small
\bibliographystyle{icml2022}
\bibliography{cites}
}

\appendix

\input{appendix_proj_so3_s2}

\input{appendix_model_generation}

\input{appendix_spherical_expmts}

\input{appendix_profiling}

\end{document}

%% file: preamble_packages.tex
\usepackage{microtype}
\usepackage{graphicx}
\usepackage{booktabs} % for professional tables

\usepackage[accepted]{icml2022}

\usepackage{mathtools}

\makeatletter
\@namedef{ver@everyshi.sty}{}
\makeatother

\usepackage{tikz}
\usepackage{pgfplots}
\pgfplotsset{compat=newest}
\DeclareUnicodeCharacter{2212}{−}
\usepgfplotslibrary{groupplots,dateplot}
\usetikzlibrary{patterns,shapes.arrows}
\usetikzlibrary{backgrounds}
\usetikzlibrary{arrows}
\usetikzlibrary{shapes,shapes.geometric,shapes.misc}

\tikzstyle{tikzfig}=[baseline=-0.25em,scale=0.5]

\pgfkeys{/tikz/tikzit fill/.initial=0}
\pgfkeys{/tikz/tikzit draw/.initial=0}
\pgfkeys{/tikz/tikzit shape/.initial=0}
\pgfkeys{/tikz/tikzit category/.initial=0}

\pgfdeclarelayer{edgelayer}
\pgfdeclarelayer{nodelayer}
\pgfsetlayers{background,edgelayer,nodelayer,main}

\tikzstyle{red_dot}=[fill=red, draw=black, shape=circle]
\tikzstyle{green_dot}=[fill=green, draw=black, shape=circle]

\tikzstyle{right_arrow}=[->]
\tikzstyle{left_arrow}=[<-]
\tikzstyle{thick_right_arrow}=[->, thick]
\tikzstyle{thick_left_arrow}=[<-, thick]
\tikzstyle{filled_path_edge}=[-, fill={rgb,255: red,209; green,209; blue,209}]

\tikzstyle{none}=[inner sep=0mm]

\usepackage{graphicx}
\usepackage[all]{xy}
\usepackage{mathrsfs}
\usepackage{marvosym}
\usepackage{stmaryrd}
\usepackage{srcltx}

\usepackage[shadow,textwidth=1.8cm]{todonotes}

\usepackage{pstricks}

\usepackage{amsfonts,amsmath,amssymb,amscd,amsthm}
\usepackage{enumerate}
\usepackage[all]{xy}
\usepackage{mathtools}
\usepackage{mathrsfs}
\usepackage{xparse}
\usepackage{float}
\usepackage{mathrsfs}
\usepackage{physics}
\usepackage{dsfont}
\usepackage{bbm}
\usepackage{bm}
\usepackage{tabu}
\usepackage{booktabs}
\usepackage{tensor}
\usepackage{textcomp}
\usepackage{subcaption}
\usepackage[group-separator={,}]{siunitx}

\usepackage{lipsum}

\usepackage[T1]{fontenc}

\usepackage{ifthen}
\usepackage[pagebackref=true,breaklinks=true,letterpaper=true,colorlinks,bookmarks=false]{hyperref}

\usepackage[capitalize,noabbrev]{cleveref}

%% file: preamble_settings.tex
\newtheorem*{namedtheorem}{\theoremname}
\newcommand{\theoremname}{testing}

\theoremstyle{definition}

\theoremstyle{remark}

\numberwithin{equation}{section}

\DeclareMathOperator{\down}{down}
\DeclareMathOperator{\up}{up}
\DeclareMathOperator{\conv}{conv}
\DeclareMathOperator{\maxPool}{maxPool}
\DeclareMathOperator{\tConv}{tConv}
\DeclareMathOperator{\upsample}{upsample}
\DeclareMathOperator{\StwoSOthreeconv}{S2SO3conv}
\DeclareMathOperator{\SOthreeconv}{SO3conv}
\DeclareMathOperator{\SOthreeStwoconv}{SO3S2conv}

\newcommand{\SO}{\mathrm{SO}}

\newcounter{todocounter}

\colorlet{jgcolor}{green!40!white}

\newcommand{\jginline}[2][]{
  \ifthenelse { \equal {#1} {} }
    { \def\temp {#2} }  % if #1 == blank
    { \def\temp {#1} }   % else (not blank)
  \refstepcounter{todocounter}\todo[color=jgcolor,inline,caption={\textbf{\thetodocounter. JG} \temp}]{\textbf{\thetodocounter. JG:} #2}{}}
\newcommand{\jgblock}[2][{}]{
  \ifthenelse { \equal {#1} {} }
  { \def\templist {\emph{block comment}}
    \def\tempheader {}}  % if #1 == blank
  { \def\templist {#1}
    \def\tempheader {#1}}   % else (not blank)
  \refstepcounter{todocounter}\todo[color=jgcolor,inline,caption={\textbf{\thetodocounter. JG} \templist}]{\textbf{\thetodocounter. JG: \tempheader}\\\begin{minipage}{\textwidth}#2\end{minipage}}{}}

\colorlet{jacolor}{blue!20!white}

\newcommand{\jainline}[2][]{
  \ifthenelse { \equal {#1} {} }
    { \def\temp {#2} }  % if #1 == blank
    { \def\temp {#1} }   % else (not blank)
  \refstepcounter{todocounter}\todo[color=jacolor,inline,caption={\textbf{\thetodocounter. JA} \temp}]{\textbf{\thetodocounter. JA:} #2}{}}
\newcommand{\jablock}[2][{}]{
  \ifthenelse { \equal {#1} {} }
  { \def\templist {\emph{block comment}}
    \def\tempheader {}}  % if #1 == blank
  { \def\templist {#1}
    \def\tempheader {#1}}   % else (not blank)
  \refstepcounter{todocounter}\todo[color=jacolor,inline,caption={\textbf{\thetodocounter. JA} \templist}]{\textbf{\thetodocounter. JA: \tempheader}\\\begin{minipage}{\textwidth}#2\end{minipage}}{}}

\colorlet{dpcolor}{yellow!40!white}

\newcommand{\dpinline}[2][]{
  \ifthenelse { \equal {#1} {} }
    { \def\temp {#2} }  % if #1 == blank
    { \def\temp {#1} }   % else (not blank)
  \refstepcounter{todocounter}\todo[color=dpcolor,inline,caption={\textbf{\thetodocounter. DP} \temp}]{\textbf{\thetodocounter. DP:} #2}{}}
\newcommand{\dpblock}[2][{}]{
  \ifthenelse { \equal {#1} {} }
  { \def\templist {\emph{block comment}}
    \def\tempheader {}}  % if #1 == blank
  { \def\templist {#1}
    \def\tempheader {#1}}   % else (not blank)
  \refstepcounter{todocounter}\todo[color=dpcolor,inline,caption={\textbf{\thetodocounter. DP} \templist}]{\textbf{\thetodocounter. DP: \tempheader}\\\begin{minipage}{\textwidth}#2\end{minipage}}{}}

\colorlet{cpcolor}{red!40!white}

\newcommand{\cpinline}[2][]{
  \ifthenelse { \equal {#1} {} }
    { \def\temp {#2} }  % if #1 == blank
    { \def\temp {#1} }   % else (not blank)
  \refstepcounter{todocounter}\todo[color=cpcolor,inline,caption={\textbf{\thetodocounter. CP} \temp}]{\textbf{\thetodocounter. CP:} #2}{}}
\newcommand{\cpblock}[2][{}]{
  \ifthenelse { \equal {#1} {} }
  { \def\templist {\emph{block comment}}
    \def\tempheader {}}  % if #1 == blank
  { \def\templist {#1}
    \def\tempheader {#1}}   % else (not blank)
  \refstepcounter{todocounter}\todo[color=cpcolor,inline,caption={\textbf{\thetodocounter. CP} \templist}]{\textbf{\thetodocounter. CP: \tempheader}\\\begin{minipage}{\textwidth}#2\end{minipage}}{}}

\colorlet{hlcolor}{purple!40!white}

\newcommand{\hlinline}[2][]{
  \ifthenelse { \equal {#1} {} }
    { \def\temp {#2} }  % if #1 == blank
    { \def\temp {#1} }   % else (not blank)
  \refstepcounter{todocounter}\todo[color=hlcolor,inline,caption={\textbf{\thetodocounter. HL} \temp}]{\textbf{\thetodocounter. HL:} #2}{}}
\newcommand{\hlblock}[2][{}]{
  \ifthenelse { \equal {#1} {} }
  { \def\templist {\emph{block comment}}
    \def\tempheader {}}  % if #1 == blank
  { \def\templist {#1}
    \def\tempheader {#1}}   % else (not blank)
  \refstepcounter{todocounter}\todo[color=hlcolor,inline,caption={\textbf{\thetodocounter. HL} \templist}]{\textbf{\thetodocounter. HL: \tempheader}\\\begin{minipage}{\textwidth}#2\end{minipage}}{}}

\colorlet{focolor}{orange!40!white}

\newcommand{\foinline}[2][]{
  \ifthenelse { \equal {#1} {} }
    { \def\temp {#2} }  % if #1 == blank
    { \def\temp {#1} }   % else (not blank)
  \refstepcounter{todocounter}\todo[color=focolor,inline,caption={\textbf{\thetodocounter. FO} \temp}]{\textbf{\thetodocounter. FO:} #2}{}}
\newcommand{\foblock}[2][{}]{
  \ifthenelse { \equal {#1} {} }
  { \def\templist {\emph{block comment}}
    \def\tempheader {}}  % if #1 == blank
  { \def\templist {#1}
    \def\tempheader {#1}}   % else (not blank)
  \refstepcounter{todocounter}\todo[color=focolor,inline,caption={\textbf{\thetodocounter. FO} \templist}]{\textbf{\thetodocounter. FO: \tempheader}\\\begin{minipage}{\textwidth}#2\end{minipage}}{}}

\colorlet{occolor}{purple!40!white}

\newcommand{\ocinline}[2][]{
    \ifthenelse { \equal {#1} {} }
    { \def\temp {#2} }  % if #1 == blank
    { \def\temp {#1} }   % else (not blank)
    \refstepcounter{todocounter}
    \todo[color=occolor,
          inline,
          caption={\textbf{\thetodocounter. OC} \temp}
          ]{\textbf{\thetodocounter. OC:} #2}{}
}
\newcommand{\ocblock}[2][{}]{
  \ifthenelse { \equal {#1} {} }
  { \def\templist {\emph{block comment}}
    \def\tempheader {}}  % if #1 == blank
  { \def\templist {#1}
    \def\tempheader {#1}}   % else (not blank)
  \refstepcounter{todocounter}\todo[color=occolor,inline,caption={\textbf{\thetodocounter. OC} \templist}]{\textbf{\thetodocounter. OC: \tempheader}\\\begin{minipage}{\textwidth}#2\end{minipage}}{}}

%% file: section_introduction.tex
\section{Introduction}
\label{sec:introduction}

\begin{figure}
  \centering
  \includegraphics[width=0.45\columnwidth]{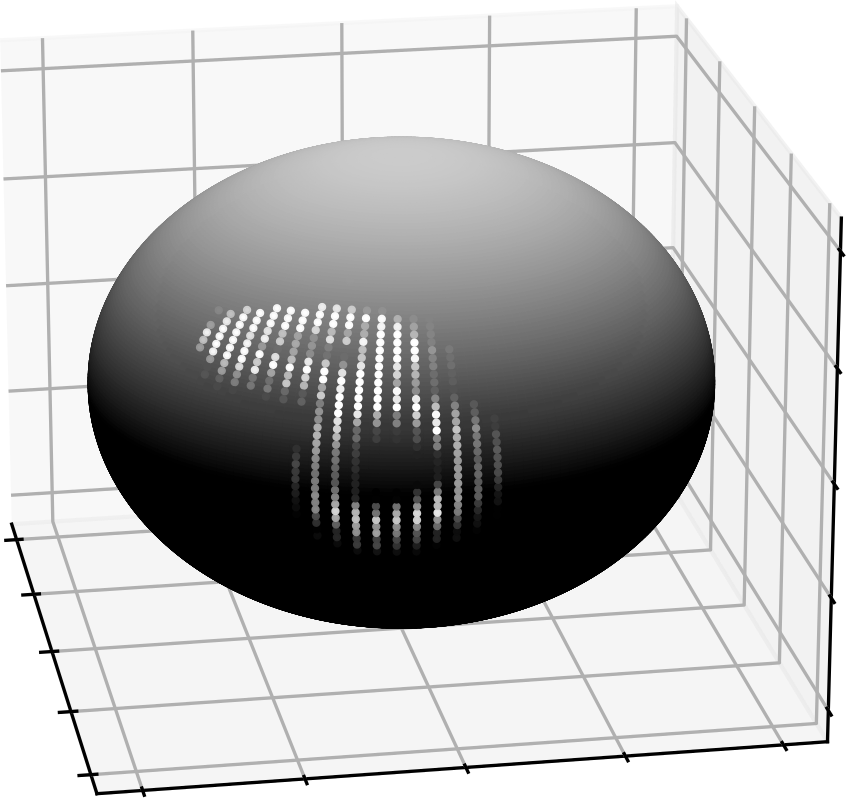}
  \hfill
  \includegraphics[width=0.45\columnwidth]{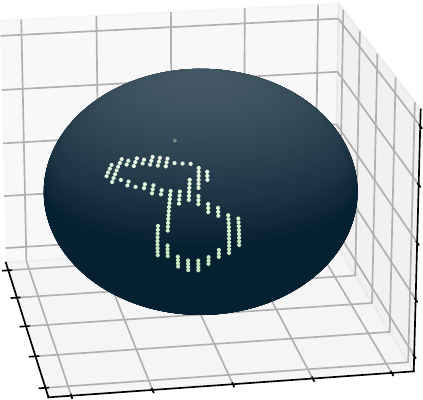}
  \caption{Sample from the spherical MNIST dataset used for semantic segmentation. Left: input data. Right: segmentation mask.}
  \vspace*{-0.5cm}
  \label{fig:sphericalMNIST}
\end{figure}
\noindent In virtually all computer vision tasks, convolutional neural networks (CNNs) consistently outperform fully connected architectures. This performance gain can be attributed to the weight sharing and translational equivariance of the CNN layers: The network does not have to learn to identify translated versions of an image, since the inductive bias of equivariance already implies this identification. In contrast, fully connected layers require many more training samples to learn an effective form of equivariance. One way of ameliorating this problem is to supply the network with translated copies of the original training images, a form of data augmentation. However, training with this kind of data augmentation requires much longer training times and the performance of CNNs may not be reached in this way.

\begin{figure}
  \centering
  \resizebox{\columnwidth}{!}{\input{imgs/spherical_data_augmentation_1_digit.tex}}
  \caption{\textit{Semantic Segmentation on Spherical MNIST.} The performance of equivariant (S2CNN) and non-equivariant (CNN) semantic segmentation models for various amounts of data augmented spherical MNIST single digit training images. Performance is measured in terms of mean intersection over union (mIoU) for the non-background classes. The numbers in the model names refer to the number of trainable parameters. The non-equivariant models are trained on randomly rotated samples, whereas the equivariant models are trained on unrotated samples.}
  \label{fig:data_augm_1_digit}
\end{figure}
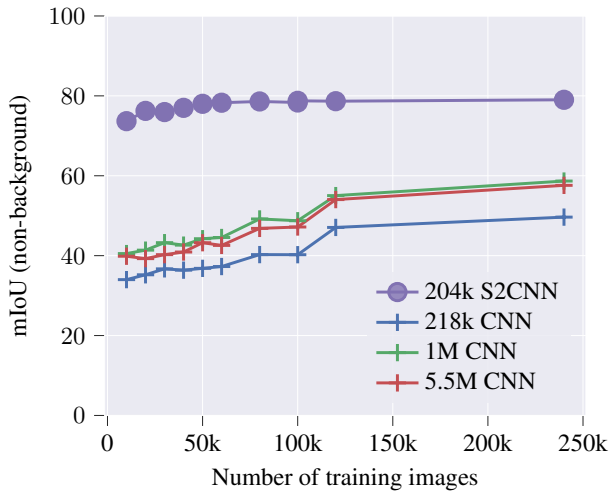

A crucial point is that in the equivariant setting, we are viewing the input image not only as a vector, but as a signal defined on a two-dimensional grid. The translational symmetry then has a geometric origin: it is a symmetry of the grid. From this perspective it is  natural to envision generalizations of CNNs where not only translational symmetries are implemented but also more general transformations, such as rotations. Examples of networks that realize this property are group equivariant CNNs which are equivariant with respect to rotations on the plane \cite{Cohen2016}, or spherical CNNs which are equivariant with respect to rotations of the sphere \cite{cohen2018b}. Similarly to the case of translations discussed above, these symmetry properties of the data can be learned approximately by a non-equivariant model using data augmentation.

In the present paper, we evaluate the performance of equivariant classification and semantic segmentation models (S2CNNs \cite{cohen2018b}) on spherical image data subject to rotational symmetry (see Figure~\ref{fig:sphericalMNIST} for an example), and compare it to the performance obtained using rotational data augmentation in ordinary CNNs (see Figure~\ref{fig:data_augm_1_digit}). We conduct comparisons for several datasets to elucidate the influence of the type and complexity of the task on the performance. Our overall aim is to investigate whether equivariant architectures offer an inherent advantage over data augmentation in non-equivariant CNN models.

Our choice of using an ordinary CNN, without additional structure to make it more compatible with spherical data, is motivated by the desire to isolate the limits of geometric data augmentation.

% Review addition, 2022-06-02, FO
While our present investigation is concerned with rotational equivariance in spherical data, group equivariant networks generalize to any homogeneous space with a global action of a symmetry transformation. In fact, the theoretical development in Section~\ref{sec:theory} applies to any homogeneous space, and the question of equivariance versus data augmentation is relevant also in the general setting.

\subsection{Summary of contributions and outline}
\label{sec:summary_of_results}
Here we list the main contributions of the paper, both theoretical and experimental.

\begin{itemize}
\item We define a new group equivariant CNN layer (Equation~(\ref{eqn:SO3S2_conv})), designed for the task of semantic segmentation for arbitrary Lie groups $G$. This generalizes previous results in \cite{linmans2018}.
    \item We extend the S2CNN architecture by adding a layer which allows for equivariant outputs on the sphere, as required for semantic segmentation tasks, and present a detailed proof of equivariance  (Appendix~\ref{sec:projection-so3-s2}).

    \item We demonstrate that non-equivariant classification models require considerable data augmentation to reach the performance of smaller equivariant networks (Section~\ref{sec:classification}).
    \item We show that the performance of non-equivariant semantic segmentation models saturates well below that of equivariant models as the amount of data augmentation is increased (Section~\ref{sec:sem-seg}). We confirm that this result still holds when the complexity of the original segmentation task is increased (Section~\ref{sec:data-complexity}).
    \item We perform detailed inference time profiling of the GPU-based implementation of the equivariant models, and measure the throughput overhead compared to the non-equivariant models. Our results indicate that most inference time is spent in the last network layers processing the largest $\SO(3)$ tensors, suggesting possible avenues towards optimizing the S2CNN architecture (Section~\ref{sec:profiling}).
    \item We show that the total training time for an equivariant model is shorter compared to a non-equivariant model at matched performance (Section~\ref{sec:profiling}).
\end{itemize}

Appendix~\ref{sec:projection-so3-s2} contains  mathematical details about our new final layer used for semantic segmentation. Details of the model generation for the equivariant and non-equivariant networks can be found in Appendix~\ref{sec:spherical-model-gen}.
Appendix~\ref{sec:spherical-expmts} contains details of the datasets and augmentation. Finally, the latency profiling is summarized in Appendix~\ref{sec:latency_profiling}.

\subsection{Related literature}
\label{sec:related-literature}

\noindent The theory of group equivariant neural networks was first developed in \cite{Kondor2018, cohen2018, esteves2020}, generalizing the ordinary planar convolution of CNNs. An implementation for spherical data with a rotational $\mathrm{SO}(3)$ symmetry was introduced in \cite{cohen2018b,kondor2018a,cobb2020}.

There are two main approaches to equivariance in semantic segmentation tasks. In the first, one utilizes the methods available for ordinary flat CNNs by modifying the shape of convolutional kernels to compensate for spherical distortion \cite{tatenoDistortionAwareConvolutionalFilters2018}, by cutting the sphere in smaller pieces or otherwise preprocessing the data to be able to apply an ordinary CNN without much distortion \cite{zhangOrientationAwareSemanticSegmentation2019, leeSpherePHDApplyingCNNs2019,haim2019surface,eder2019convolutions,duSphericalTransformerAdapting2021,pmlr-v139-shakerinava21a}. In the second approach, one avoids spherical distortions by Fourier decomposing signals and filters on the sphere \cite{esteves2020b}. Semantic segmentation for group equivariant CNNs is analyzed in \cite{linmans2018}.

Data augmentation in the context of symmetric tasks was studied previously in \cite{gandikotaTrainingArchitectureHow2021}, where a method to align input data is presented and compared to data augmentation. Closest to our work is a comparison between an equivariant model and a non-equivariant model for reduced training data sizes for an MRI application in \cite{muller2021}. In contrast, we systematically compare data augmentation with increased training data sizes for different tasks, datasets and several non-equivariant models with an equivariant architecture.

\begin{figure*}[t]
    \centering

    \input{imgs/so3_to_s2_figure_v5.tikz}

    \caption{The final layer \eqref{eq:6} of our fully equivariant architecture takes Fourier coefficients on $\mathrm{SO}(3)$ and sums them over $n$ to yield Fourier coefficients on the sphere, representing a segmentation mask.}
    \label{fig:so3tos2}
\end{figure*}
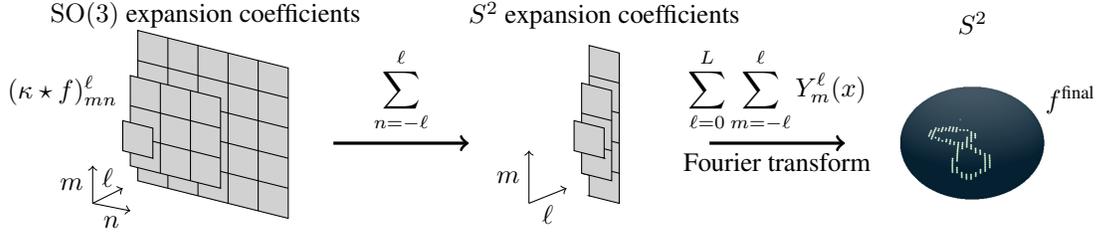

%%% Local Variables:
%%% mode: latex
%%% TeX-master: "mnist_paper"
%%% End:

%% file: imgs/spherical_data_augmentation_1_digit.tex
% This file was created with tikzplotlib v0.9.12.
\begin{tikzpicture}

\definecolor{color0}{rgb}{0.917647058823529,0.917647058823529,0.949019607843137}
\definecolor{color1}{rgb}{0.505882352941176,0.447058823529412,0.698039215686274}
\definecolor{color2}{rgb}{0.298039215686275,0.447058823529412,0.690196078431373}
\definecolor{color3}{rgb}{0.333333333333333,0.658823529411765,0.407843137254902}
\definecolor{color4}{rgb}{0.768627450980392,0.305882352941176,0.32156862745098}

\begin{axis}[
axis background/.style={fill=color0},
axis line style={white},
legend cell align={left},
legend style={
  fill opacity=0.8,
  draw opacity=1,
  text opacity=1,
  at={(0.97,0.03)},
  anchor=south east,
  draw=none,
  fill=color0
},
tick align=outside,
tick pos=left,
x grid style={white},
xlabel={Number of training images},
xmajorgrids,
xmin=-1.5, xmax=251.5,
xtick style={color=white!15!black},
xtick={0,50,100,150,200,250},
xticklabels={0,50k,100k,150k,200k,250k},
y grid style={white},
ylabel={mIoU (non-background)},
ymajorgrids,
ymin=0, ymax=100,
ytick style={color=white!15!black}
]
\addplot [very thick, color1, mark=*, mark size=3.5, mark options={solid}]
table {%
10 73.6672228087486
20 76.2396656837297
30 75.9440592307601
40 77.0006140071276
50 78.0121633327085
60 78.2821354107209
80 78.6005328484791
100 78.3440466722201
100 78.7591833919464
120 78.6653772927548
240 79.0172031795914
};
\addlegendentry{204k S2CNN}
\addplot [very thick, color2, mark=+, mark size=3.5, mark options={solid}]
table {%
10 33.9979413230564
20 35.2098953469084
30 36.6968152116464
40 36.3691463001431
50 36.8220698817118
60 37.2769826449922
80 40.2635431161527
100 40.2390027250844
120 47.0656347978214
240 49.6463416406717
};
\addlegendentry{218k CNN}
\addplot [very thick, color3, mark=+, mark size=3.5, mark options={solid}]
table {%
10 40.5212931678065
20 41.377917930675
30 43.251758735901
40 42.6359169287997
50 44.2412785935146
60 44.5472077883157
80 49.1886897951291
100 48.7415286074443
120 55.0321060520411
240 58.6775255398797
};
\addlegendentry{1M CNN}
\addplot [very thick, color4, mark=+, mark size=3.5, mark options={solid}]
table {%
10 39.8560257098186
20 39.2542836381079
30 40.2511386267728
40 40.9095744779148
50 43.239519380981
60 42.5646759801113
80 46.8256498871211
100 47.1734684919231
120 54.0339057646098
240 57.5987507390389
};
\addlegendentry{5.5M CNN}
\end{axis}

\end{tikzpicture}

%% file: imgs/so3_to_s2_figure_v5.tikz
\begin{tikzpicture}[scale=0.4]
	\begin{pgfonlayer}{nodelayer}
% 		\node [style=none] (0) at (-3.5, 3.25) {};
% 		\node [style=none] (1) at (-2.5, 3) {};
% 		\node [style=none] (2) at (-2.5, 4) {};
		\node [style=none] (0) at (-3.5, 3.00) {};
		\node [style=none] (1) at (-2.5, 2.75) {};
		\node [style=none] (2) at (-2.5, 3.75) {};
		\node [style=none] (4) at (-3.25, 5.5) {};
		\node [style=none] (5) at (-0.25, 4.75) {};
		\node [style=none] (6) at (-0.25, 1.75) {};
		\node [style=none] (7) at (-3.25, 2.5) {};
		\node [style=none] (8) at (-3.5, 4.00) {};
		\node [style=none] (9) at (2, 0.75) {};
		\node [style=none] (10) at (2, 5.75) {};
		\node [style=none] (11) at (-3, 7) {};
		\node [style=none] (12) at (-3, 2) {};
		\node [style=none] (13) at (-4.5, 1.25) {};
		\node [style=none] (14) at (-4.5, 2.5) {};
		\node [style=none] (15) at (-3.5, 1.75) {};
		\node [style=none] (16) at (-3.25, 1) {};
		\node [style=none] (18) at (-0.75, 7.5) {$\mathrm{SO}(3)$  expansion coefficients};
		\node [style=none] (19) at (3.5, 3.25) {};
		\node [style=none] (20) at (8, 3.25) {};
		\node [style=none] (21) at (11.5, 3) {};
		\node [style=none] (22) at (12.5, 2.75) {};
		\node [style=none] (23) at (12.5, 3.75) {};
		\node [style=none] (24) at (11.5, 4) {};
		\node [style=none] (25) at (11.75, 2.25) {};
		\node [style=none] (26) at (12.75, 2) {};
		\node [style=none] (27) at (12.75, 5) {};
		\node [style=none] (28) at (11.75, 5.25) {};
		\node [style=none] (29) at (12, 1.5) {};
		\node [style=none] (30) at (13, 1.25) {};
		\node [style=none] (31) at (13, 6.25) {};
		\node [style=none] (32) at (12, 6.5) {};
		\node [style=none] (33) at (10, 1.25) {};
		\node [style=none] (34) at (10, 3) {};
		\node [style=none] (35) at (11.25, 1.75) {};
		\node [style=none] (36) at (12.5, 7.5) {$S^2$ expansion coefficients};
		\node [style=none] (37) at (-5.5, 5) {$(\kappa\star f)^\ell_{mn}$};
		\node [style=none] (38) at (-2.25, 5.25) {};
		\node [style=none] (39) at (-1.25, 5) {};
		\node [style=none] (40) at (-2.25, 2.25) {};
		\node [style=none] (41) at (-1.25, 2) {};
		\node [style=none] (42) at (-3.25, 4.5) {};
		\node [style=none] (43) at (-3.25, 3.5) {};
		\node [style=none] (44) at (-0.25, 2.75) {};
		\node [style=none] (45) at (-0.25, 3.75) {};
		\node [style=none] (46) at (1, 1) {};
		\node [style=none] (47) at (0, 1.25) {};
		\node [style=none] (48) at (-1, 1.5) {};
		\node [style=none] (49) at (-2, 1.75) {};
		\node [style=none] (50) at (1, 6) {};
		\node [style=none] (51) at (0, 6.25) {};
		\node [style=none] (52) at (-1, 6.5) {};
		\node [style=none] (53) at (-2, 6.75) {};
		\node [style=none] (54) at (2, 4.75) {};
		\node [style=none] (55) at (2, 3.75) {};
		\node [style=none] (56) at (2, 2.75) {};
		\node [style=none] (57) at (2, 1.75) {};
		\node [style=none] (58) at (-3, 3) {};
		\node [style=none] (59) at (-3, 4) {};
		\node [style=none] (60) at (-3, 5) {};
		\node [style=none] (61) at (-3, 6) {};
		\node [style=none] (62) at (11.75, 3.25) {};
		\node [style=none] (63) at (12.75, 3) {};
		\node [style=none] (64) at (11.75, 4.25) {};
		\node [style=none] (65) at (12.75, 4) {};
		\node [style=none] (66) at (12, 2.5) {};
		\node [style=none] (67) at (13, 2.25) {};
		\node [style=none] (68) at (12, 3.5) {};
		\node [style=none] (69) at (13, 3.25) {};
		\node [style=none] (70) at (12, 4.5) {};
		\node [style=none] (71) at (13, 4.25) {};
		\node [style=none] (72) at (12, 5.5) {};
		\node [style=none] (73) at (13, 5.25) {};
		\node [style=none] (74) at (16, 3.25) {};
		\node [style=none] (75) at (20.5, 3.25) {};
		\node [style=none] (78) at (24.75, 7.25) {$S^2$};
		\node [style=none] (79) at (24.75, 3.25) {\includegraphics[width=0.15\textwidth]{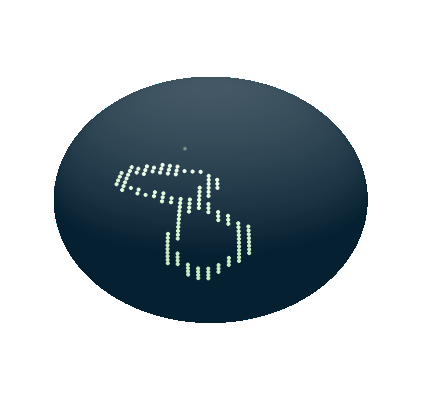}};
		\node [style=none] (80) at (28,4.7) {$f^{\text{final}}$};
	\end{pgfonlayer}
	\begin{pgfonlayer}{edgelayer}
		\draw [style={right_arrow}] (13.center) to node [midway, above] {$\ell$} (15.center);
		\draw [style={right_arrow}] (13.center) to node [midway, left] {$m$} (14.center);
		\draw [style={right_arrow}] (13.center) to node [midway, below] {$n$} (16.center);
		\draw [style={right_arrow}, very thick] (19.center) to node [midway, above] {$\displaystyle\sum_{n=-\ell}^\ell$} (20.center);
		\draw [style={right_arrow}] (33.center) to node [midway, below] {$\ell$} (35.center);
		\draw [style={right_arrow}] (33.center) to node [midway, left] {$m$} (34.center);
		\draw [style={filled_path_edge}] (25.center) to (26.center);
		\draw [style={right_arrow}, very thick] (74.center) to node [midway, above] {$\displaystyle\sum_{\ell=0}^L\displaystyle\sum_{m=-\ell}^\ell Y^\ell_m(x)$} node [midway, below] {Fourier transform} (75.center);
		\draw [style={filled_path_edge}] (11.center)
			 to (12.center)
			 to (9.center)
			 to (10.center)
			 to cycle;
		\draw (53.center) to (49.center);
		\draw (52.center) to (48.center);
		\draw (51.center) to (47.center);
		\draw (50.center) to (46.center);
		\draw (61.center) to (54.center);
		\draw (60.center) to (55.center);
		\draw (59.center) to (56.center);
		\draw (58.center) to (57.center);
		\draw [style={filled_path_edge}] (5.center)
			 to (4.center)
			 to (7.center)
			 to (6.center)
			 to cycle;
		\draw (39.center) to (41.center);
		\draw (38.center) to (40.center);
		\draw (42.center) to (45.center);
		\draw (43.center) to (44.center);
		\draw [style={filled_path_edge}] (1.center)
			 to (2.center)
			 to (8.center)
			 to (0.center)
			 to cycle;
		\draw [style={filled_path_edge}] (29.center)
			 to (30.center)
			 to (31.center)
			 to (32.center)
			 to cycle;
		\draw [style={filled_path_edge}] (66.center) to (67.center);
		\draw [style={filled_path_edge}] (68.center) to (69.center);
		\draw [style={filled_path_edge}] (70.center) to (71.center);
		\draw [style={filled_path_edge}] (72.center) to (73.center);
		\draw [style={filled_path_edge}] (28.center)
			 to (25.center)
			 to (26.center)
			 to (27.center)
			 to cycle;
		\draw [style={filled_path_edge}] (64.center) to (65.center);
		\draw [style={filled_path_edge}] (62.center) to (63.center);
		\draw [style={filled_path_edge}] (22.center)
			 to (23.center)
			 to (24.center)
			 to (21.center)
			 to cycle;
	    %\shadedraw[shading=ball,ball color=gray, white] (79) circle (2);

	\end{pgfonlayer}
\end{tikzpicture}

%% file: section_theory.tex
\section{Theory}
\label{sec:theory}
\subsection{Group equivariant networks}
\label{sec:group-equiv-conv-nets}

In this section we introduce the basic mathematical structure of GCNNs. Let $G$ be a group and $H\subset G$ a closed subgroup\footnote{In this paper, we will always consider the case where $G$ is a Lie group and $H$ is a compact subgroup.}. The input space of the network is the homogeneous space $\mathcal{M} = G/H$, meaning that a feature map in the first layer is a map
\begin{equation}
\label{eqn:input_feature_map}
f\, : G/H \to \mathbb{R}^K,
\end{equation}
where $K$ denotes the number of channels. This feature map represents the input data, e.g. the pixel values of an image represented on the homogeneous space $G/H$. Features of subsequent layers are obtained using the convolution between the feature map $f$ and a filter $\kappa : G/H \to \mathbb{R}^{K',K}$ defined by
\begin{equation}
\label{eqn:conv_first_layer}
(\kappa\star f)(g)=\int_{G/H} \kappa(g^{-1}x)f(x)\dd{x},
\end{equation}
where $g \in G$ and $dx$ is the invariant  measure on $G/H$ induced from the Haar measure on $G$.

The action of $G$ extends from $\mathcal{M}=G/H$ to $G$ itself and the convolution~(\ref{eqn:conv_first_layer}) is equivariant with respect to this action, $g (\kappa \star f) = \kappa \star (gf), \, g \in G$, meaning that the result obtained by transforming the convoluted feature map $\kappa \star f$ by an element $g\in G$ is identical to that obtained by convolving the transformed feature map $gf$. In other words, the transformation $g \in G$ commutes with the convolution.

The networks considered here all have scalar features throughout the architecture. In general, however, the group $G$ can act through non-trivial representations on both the manifold $G/H$ and the feature maps (see e.g. \cite{Kondor2018,cohen2018,aronsson2021,gerkenGeometricDeepLearning2021}).

Note that the output of the convolution~(\ref{eqn:conv_first_layer}) is a function on $G$, rather than on the homogeneous space $G/H$. Limiting feature maps to convolutions on the original manifold $G/H$ is in general far too restrictive, as was first noticed in the context of spherical signals by \cite{makadia2007}. In order to keep the network as general as possible, and maximize the expressiveness of each individual layer, subsequent convolutions will all be taken on the group $G$,
\begin{equation}
\label{eqn:conv_subseq_layer}
(\kappa\star f)(g)=\int_{G} \kappa(g'^{-1}g)f(g^{\prime})\dd{g^{\prime}},
\end{equation}
$g \in G$, except for the last one which will be discussed in more detail below.

\subsection{Semantic segmentation and equivariance}
\label{sec:sem-seg-equiv}

In the final layers we must take into account the desired task that the network is designed to perform. If we are interested in a classification task, we want the entire network to be invariant with respect to $G$. This can be achieved by integrating over $G$ following the final convolution
\begin{equation}
    f^{\text{final}} = \int_{G} f(g) \dd{g} \,.
\end{equation}
For semantic segmentation, however, we are classifying individual pixels into a segmentation mask. The network output should therefore be equivariant with respect to transformations of the input.  One way to achieve this aim is to define a final convolution
\begin{equation}
\label{eqn:SO3S2_conv}
f^{\text{final}}(x)=\int_G \kappa(g'^{-1}g_x)f(g^{\prime})\dd{g^{\prime}},
\end{equation}
where $g_x\in G$ is a representative in $G$ corresponding to the point $x$ in the coset space $G/H$ and $\kappa$ is required to be an $H$-invariant kernel on $G$. This generalizes the convolution defined in \cite{linmans2018} for finite roto-translations on the plane. The output~(\ref{eqn:SO3S2_conv}) is then a signal on $G/H$, rather than $G$, and equivariance ensures that a transformation by $G$ on the input image will result in the same transformation on the output.

Alternatively, similarly to the classification case, an output segmentation mask on $G/H$ can be obtained by integrating along the orbits of the subgroup $H$
\begin{equation}
    f^{\text{final}}(x) = \int_{H} f(g_xh) \dd{h} \,,\label{eq:10}
\end{equation}
with $g_x$ defined as above. This segmentation mask is equivariant with respect to the action of $G$ by construction.

%%% Local Variables:
%%% mode: latex
%%% TeX-master: "mnist_paper"
%%% End:

%% file: section_spherical_case.tex
\section{Data augmentation for classification and semantic segmentation}
\label{sec:spherical_case}

In this section, we investigate several aspects of the performance difference between equivariant and non-equivariant models, applied to both classification and segmentation tasks on spherical images. Consequently, we consider input data in the form of grayscale images defined on the sphere $\mathcal{M} = S^2$ with continuous rotational symmetry $G=\mathrm{SO}(3)$. Taking $H=\mathrm{SO}(2)$ to be the isotropy subgroup of a point $x \in S^2$, the sphere can be expressed as a homogeneous space $S^2 = \mathrm{SO}(3)/\mathrm{SO}(2)$.

\subsection{Equivariant model}
\label{sec:equivariant-model}

As discussed in Section~\ref{sec:theory}, the equivariant network architecture for spherical signals necessitates a redefinition of the convolution compared to ordinary CNNs. For our implementation we will use the S2CNN architecture of \cite{cohen2018b}, in the implementation available at \cite{kohlerSphericalCNNs2022}, adapted in the case of semantic segmentation.\footnote{Our adaptation is  available at \url{https://github.com/JanEGerken/sem_seg_s2cnn}.}

For S2CNNs the first layer takes an input feature map $f$ defined on $S^{2}$, convolves it with a kernel $\kappa$ defined on $\mathrm{SO}(3)$ and outputs a feature map defined on $\mathrm{SO}(3)$ according to
\begin{align}
  (\kappa \star f)(R) = \int_{S^{2}}\kappa(R^{-1}x)f(x)\,\dd{x}\,,\label{eq:4}
\end{align}
where $R \in \mathrm{SO}(3)$. In subsequent layers, the input feature map $f$ is also defined on $\mathrm{SO}(3)$ and the convolution takes the form
\begin{align}
  (\kappa \star f)(R) = \int_{\SO(3)} \kappa(S^{-1}R) f(S) \,\dd S\,.\label{eq:5}
\end{align}
In the S2CNN architecture, these convolutions are computed in the respective Fourier domain of $S^{2}$ and $\mathrm{SO}(3)$. For $S^{2}$, this amounts to an expansion in spherical harmonics $Y^{\ell}_{m}$; for $\mathrm{SO}(3)$, to an expansion in Wigner matrices $\mathcal{D}^{\ell}_{mn}$ with $\ell=0,\dots,L$ and $m,n=-\ell\dots\ell$ for some bandlimit $L$.\footnote{Note that following the original S2CNN code \cite{cohen2018b}, we use Wigner matrices for the Fourier transform and complex conjugated Wigner matrices for the inverse Fourier transform, contrary to the usual convention.}

The original S2CNN architecture introduced in \cite{cohen2018b} was used for classification tasks and hence in the last convolutional layer the feature map was integrated over $\mathrm{SO}(3)$ to render the output invariant under rotations of the input. We use the same setup for the classification task discussed below.

In contrast, for semantic segmentation, we need an equivariant network, as detailed in Section~\ref{sec:sem-seg-equiv} above. To this end, instead of the Fourier-back-transform on $\mathrm{SO}(3)$, we use for the last convolution
\begin{align}
  f^{\text{final}}(x)=\sum_{\ell=0}^{L}\sum_{m=-\ell}^{\ell}\sum_{n=-\ell}^{\ell}(\kappa\star f)^{\ell}_{mn}Y^{\ell}_{m}(x)\,,\label{eq:6}
\end{align}
where $\kappa\star f$ is as in~\eqref{eq:5} on $\mathrm{SO}(3)$. The sum over $n$ corresponds roughly to the Fourier space version of the integral over $H$ presented in Equation~\eqref{eq:10} and ensures that the output $f^{\text{final}}$ lives on $S^2$, as illustrated in Figure~\ref{fig:so3tos2}. We provide more mathematical details in Appendix~\ref{sec:projection-so3-s2}.

After a softmax nonlinearity, we interpret $f^{\text{final}}$ as the predicted segmentation mask and use a point-wise cross entropy loss for training.

To facilitate the segmentation task, we use a sequence of downsampling $\mathrm{SO}(3)$ convolutions followed by a sequence of upsampling convolutions. These networks do not contain any fully connected layers. For the classification task, we only use the downsampling layers, which are, after the integration over $\mathrm{SO}(3)$,  followed by three fully connected layers, alternated with batch-normalization layers.

The downsampling layers are realized in the S2CNN framework by choosing the bandlimit for the Fourier-back-transform to be lower than the bandlimit of the Fourier transform, i.e.\ we are dropping higher order modes in the result of the convolution. This is similar to strides used in downsampling layers of ordinary CNNs. Since the Fourier transforms of the feature maps are computed using the FFT algorithm, this also decreases the spatial resolution of the feature maps.

For the upsampling layers, we select the Fourier-back-transform to have higher bandlimit than the Fourier transform. The missing Fourier modes are here filled with zeros, yielding a feature map of higher spatial resolution (but of course with still lower resolution in the Fourier domain). This is similar to upsampling by interpolation.

To fix the precise architectures and hyperparameters for the experiments without biasing for or against the equivariant models, we generated architectures at random. For the equivariant models, we generated 20 models at random for each parameter range and selected the one performing best on a reference task, as detailed for the semantic segmentation models in Appendix~\ref{sec:equiv-model-gen}. This resulted in two equivariant models with 204k and 820k parameters, respectively for the semantic segmentation tasks and one model with 150k parameters for the classification task.

For the semantic segmentation tasks, we generated three non-equivariant models as a baseline with 218k, 1M and 5.5M parameters, respectively. Each of these performed best out of 20 randomly generated models with a fixed parameter budget. The non-equivariant models have (ordinary) convolutional layers for downsampling and transpose convolutions for upsampling layers which mirror the image dimensions and channels of the downsampling layers exactly. We add skip connections over each (transpose) convolution, resulting in a ResNet-like architecture. For details, cf.~Appendix~\ref{sec:non-equiv-model-gen}. The precise architectures of the models used in our experiments are summarized in Tables~\ref{tab:non_equiv_models} and \ref{tab:equiv_models} in the appendix.

Similarly, for the classification task, we generated three non-invariant models with 114k, 0.5M and 2.5M parameters, respectively. These models only have downsampling layers, followed by an average pooling layer and three fully connected layers alternated with batch-normalization layers. We did not use skip connections in this case.

\subsection{Classification}
\label{sec:classification}

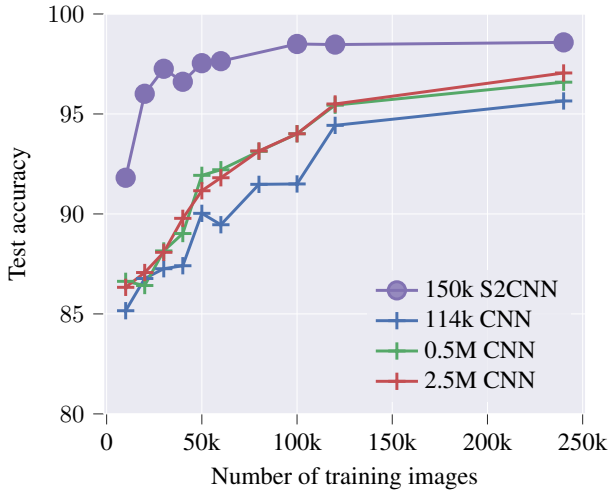
\begin{figure}
  \centering
  \resizebox{\columnwidth}{!}{\input{imgs/spherical_classification_data_augmentation_1_digit}}
  \caption{\textit{Image Classification on Spherical MNIST}. Accuracy of invariant and non-invariant classification models for various amounts of data augmentation as in Figure~\ref{fig:data_augm_1_digit}. Training data for the equivariant model is sampled randomly with replacement from the available 60k training samples.}
  \vspace{-0.5cm}
  \label{fig:spherical_classification}
\end{figure}

The primary question we want to investigate in this work is whether data augmentation can make up for the benefits of equivariant network architectures. We first study this question in the context of classification by training equivariant and non-equivariant models on training data with different amounts of data augmentation, as depicted in Figure~\ref{fig:spherical_classification}. The input data consists of single digits sampled with replacement from the MNIST~\cite{lecun1998a} dataset which are projected onto the sphere and labeled with the classes of the digits. The spherical pictures are rotated by a random rotation matrix in $\mathrm{SO}(3)$. A sample is depicted in the left panel of Figure~\ref{fig:sphericalMNIST}. For dataset larger than the 60k original MNIST dataset, digits are necessarily repeated but have different rotations, hence these correspond to data augmentation as compared to the original dataset. The equivariant networks were trained on unrotated spherical images with the digits being projected on the southern hemisphere, sampled randomly with replacement from the original 60k training samples of MNIST.\footnote{Note that we did not take special precautions to treat ambiguous cases such as ``6'' vs.\ ``9''.}

Figure~\ref{fig:spherical_classification} shows that for the smaller data regimes the accuracy of the equivariant model dominates the non-equivariant models, even though it uses fewer parameters than the non-equivariant models. Whereas the non-equivariant models continue to benefit from the increased dataset size, the equivariant model does not improve beyond the original 60k training samples of MNIST as expected. Looking at the trend of the non-equivariant models in Figure~\ref{fig:spherical_classification} it is unclear if they would eventually match the equivariant model for large enough augmented datasets. It turns out that for the task of spherical MNIST classification, a large enough non-equivariant CNN trained on augmented data can achieve similar performance to an equivariant model, see Figure~\ref{fig:saturation} left.

A priori, it is not clear whether increasing model size and data augmentation is always sufficient for non-equivariant models to match the performance of equivariant models. What is clear is that spherical MNIST classification leaves much to be desired in terms of task and dataset complexity.

\subsection{Semantic segmentation}
\label{sec:sem-seg}
\begin{figure*}[h]
  \centering
  \resizebox{0.49\textwidth}{!}{\input{imgs/spherical_data_augmentation_non_rot_projected_on_grid_center_non_rot_eval.tex}}
  \hfill
  \resizebox{0.49\textwidth}{!}{\input{imgs/spherical_data_augmentation_non_rot_projected_on_grid_center_rot_eval.tex}}
  \caption{\textit{Training on Unrotated Images}. Performance of equivariant and non-equivariant models in semantic segmentation for various amounts of data augmentation for models trained on unrotated data with one digit. Performance  is  measured  in terms of mIoU for the non-background classes. Left: evaluated on unrotated test data. Right: evaluated on rotated test data.}
  \label{fig:data_augm_non_rot}
\end{figure*}
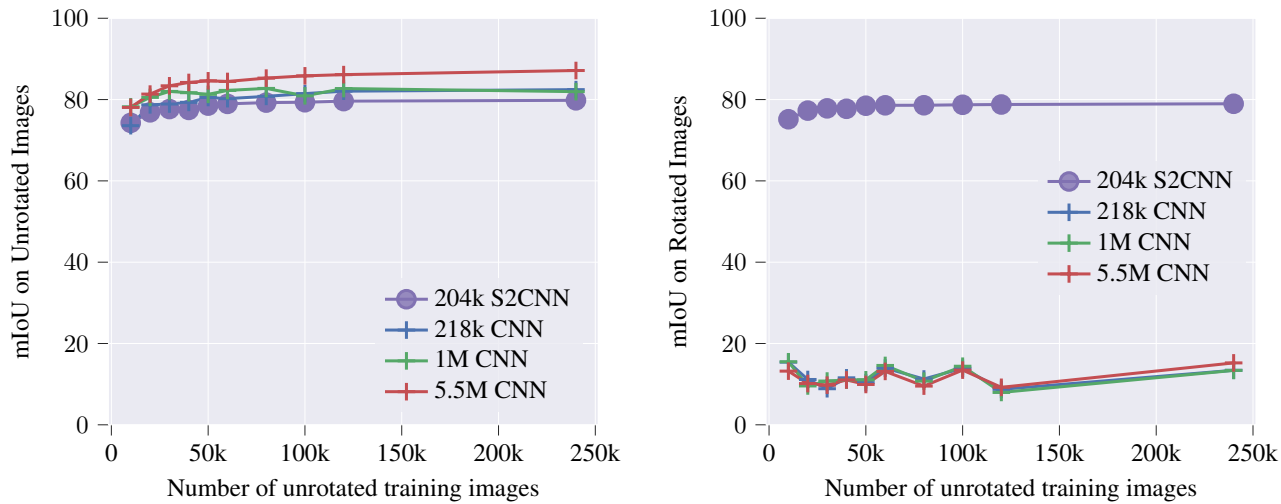
Semantic segmentation is an interesting example of a task where the output is equivariant, in contrast to the invariance of classification output. This example lets us investigate if data augmentation on large enough non-equivariant models can match the equivariant models not only in classification tasks, but also in more difficult and truly equivariant tasks.

For our experiments, we replaced the classification labels from the spherical MNIST dataset described above by segmentation masks. A sample mask is depicted in the right panel of Figure~\ref{fig:sphericalMNIST}, further details can be found in Appendix~\ref{sec:spherical-expmts}.

In order to investigate whether data augmentation can push the performance of non-equivariant networks to match that of equivariant architectures, we trained the non-equivariant networks on rotated samples and compare to equivariant networks trained only on non-rotated samples. For evaluation, we use the mean intersection over union (mIoU) where we drop the background class. More technical details can be found in Appendix~\ref{sec:spherical-expmts}.

The plot in Figure~\ref{fig:data_augm_1_digit} shows the results of this experiment. Sample predictions for the same task with four digits are shown in Appendix~\ref{sec:sample-predictions}. As expected, more training data (and hence stronger data augmentation) increases the performance of the non-equivariant models. However, the equivariant models outperform the non-equivariant models even for copious amounts of data augmentation. As is also shown in Figure~\ref{fig:data_augm_1_digit}, larger models outperform smaller models, as expected. However, there seems to be a saturation of this effect as the 5.5M parameter model performs on par (within statistical fluctuations) with the 1M parameter model. Notably though, even the largest model trained on data augmented with a factor of 20 cannot outperform the equivariant models. We also trained the non-equivariant models on even larger datasets with up to 1.2M data points, but their performance saturates at a level comparable to what is reached at 240k train data points, see Figure~\ref{fig:saturation} right.

We also trained larger spherical models to see if we could push performance even further. However, even the 820k parameter model specified in Table~\ref{tab:equiv_models} in the appendix performed only at $76.71\%$ non-background mIoU for 60k training data points. This is on par with the performance of $78.28\%$ non-background mIoU that the 204k spherical model reached for this dataset and hence suggests that the smaller model already exhausts the chosen architecture for this problem. Since the larger model requires much more compute, we performed all the remaining experiments only with the 204k spherical model.

In order to verify that our non-equivariant models are in principle expressive enough to learn the given datasets, we also trained them on unrotated data, as we did for the spherical models. From the plot in the left panel of Figure~\ref{fig:data_augm_non_rot}, it is clear that all the models perform well when evaluated on the same data on which they were trained. As shown in the right panel of Figure~\ref{fig:data_augm_non_rot}, performance deteriorates to almost random guessing (as expected) if the models are evaluated on rotated test data. The performance of the equivariant models is identical for both cases. For the task of training and evaluation on unrotated data, which contains no symmetries, the considerably larger non-equivariant models slightly outperform the equivariant model. The performance of the smallest non-equivariant model is on par with that of the equivariant model.

\begin{figure}[t]
  \centering
  \includegraphics[width=0.4\columnwidth]{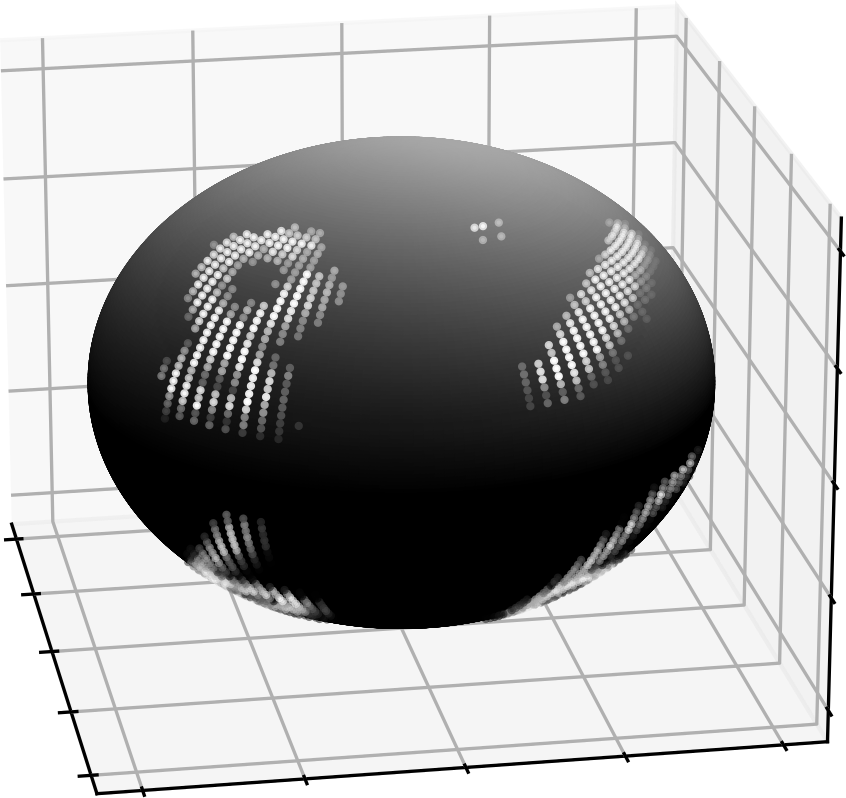}
  \hfill
  \includegraphics[width=0.4\columnwidth]{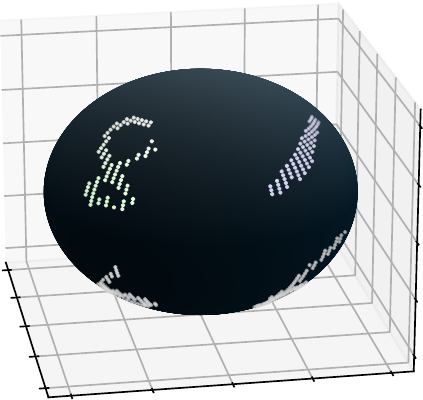}
  \caption{Sample from the spherical MNIST dataset with four digits projected onto the sphere. Left: input data. Right: segmentation mask.}
  \label{fig:sphericalMNIST_4_digits}
\end{figure}
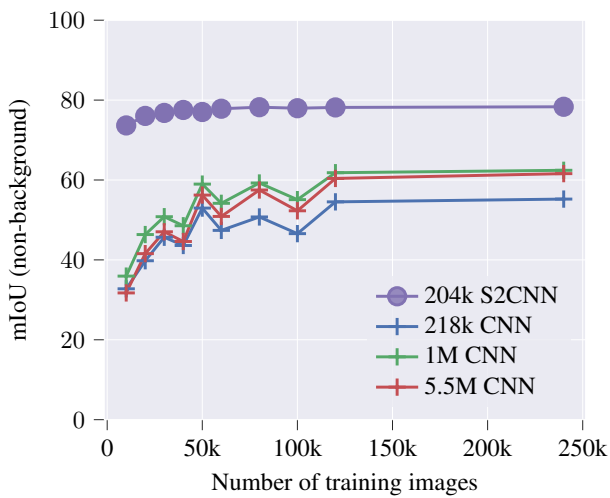
\begin{figure}[t]
  \centering
  \resizebox{\columnwidth}{!}{\input{imgs/spherical_data_augmentation_4_digits.tex}}
  \caption{\textit{Segmentation on four digit Spherical MNIST}. Performance of equivariant and non-equivariant models in semantic segmentation for various amounts of data augmentation as in Figure~\ref{fig:data_augm_1_digit}, but with four digits projected onto the sphere.}
  \label{fig:data_augm_4_digits}
\end{figure}

Note that the performance of the non-equivariant models furthermore crucially depends on where on the spherical grid the digits are projected. For the runs depicted in Figure~\ref{fig:data_augm_non_rot}, the digits were projected in the center of the Driscoll-Healy grid, so that the ordinary CNNs could benefit maximally from their translation equivariance and distortions were minimal. In Appendix~\ref{sec:infl-proj-point} we show that performance is greatly reduced if the digits are projected closer to the pole of the sphere, where distortions in the equirectangular projection of the Driscoll-Healy grid are maximal.

\subsection{Dataset complexity}
\label{sec:data-complexity}
The experiments described in the previous section used a very simple dataset, and since only one class needed to be predicted, all non-zero pixels belonged to the same foreground class. We have therefore investigated if the observed performance gain of the equivariant models persists for more complex datasets.

The first modification we performed on the dataset depicted in Figure~\ref{fig:sphericalMNIST} is to project four MNIST digits onto the same sphere and construct a corresponding segmentation mask. A sample from this dataset is shown in Figure~\ref{fig:sphericalMNIST_4_digits}. The results of these experiments are summarized in Figure~\ref{fig:data_augm_4_digits}, sample predictions for the best equivariant and non-equivariant models are shown in Appendix~\ref{sec:sample-predictions}. Note that the non-monotonic increase in performance in Figure~\ref{fig:data_augm_4_digits} is due to sampling effects during the data generation. We have explicitly verified that these features are within the range of the statistical fluctuations of this sampling and since this affects all models equally, these features are irrelevant for the model comparison. More details on this point are given in Appendix~\ref{sec:perf-vari-data-sampl}.

Moreover, for an increased number of digits, we observe a clear benefit of the equivariant architectures. Again, increasing model sizes beyond a certain parameter count does not translate into an increase in performance. In comparison to Figure~\ref{fig:data_augm_1_digit}, we see that all models in Figure~\ref{fig:data_augm_4_digits} benefited from seeing more samples during training, with a steeper increase in performance with number of training samples and earlier saturation.

\begin{figure}[t]
  \centering
  \hfill
  \includegraphics[width=0.4\columnwidth]{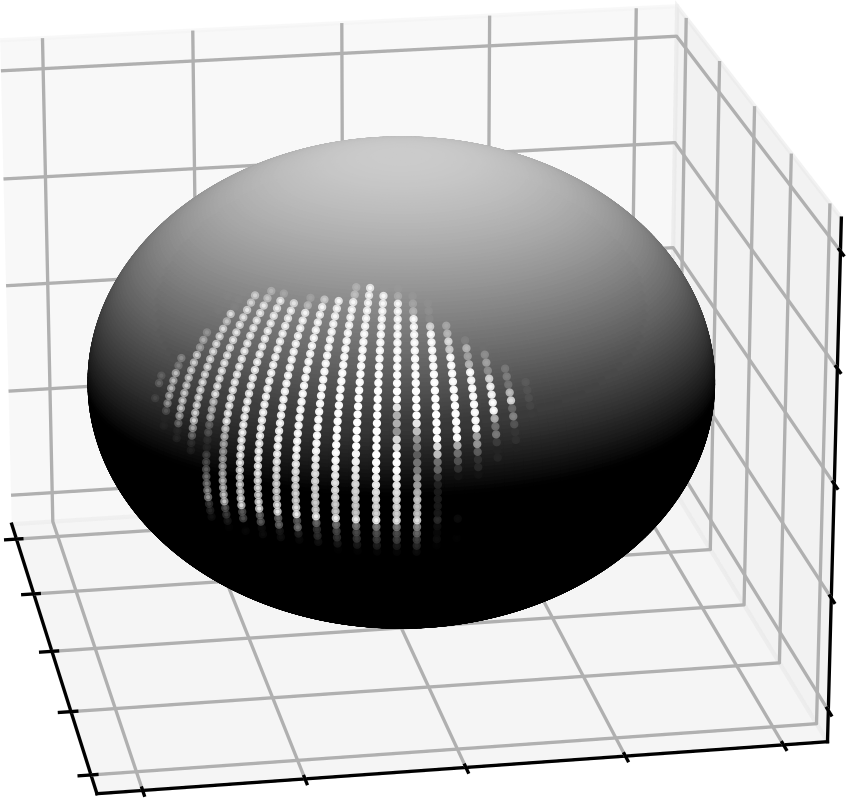}
  \hfill
  \includegraphics[width=0.4\columnwidth]{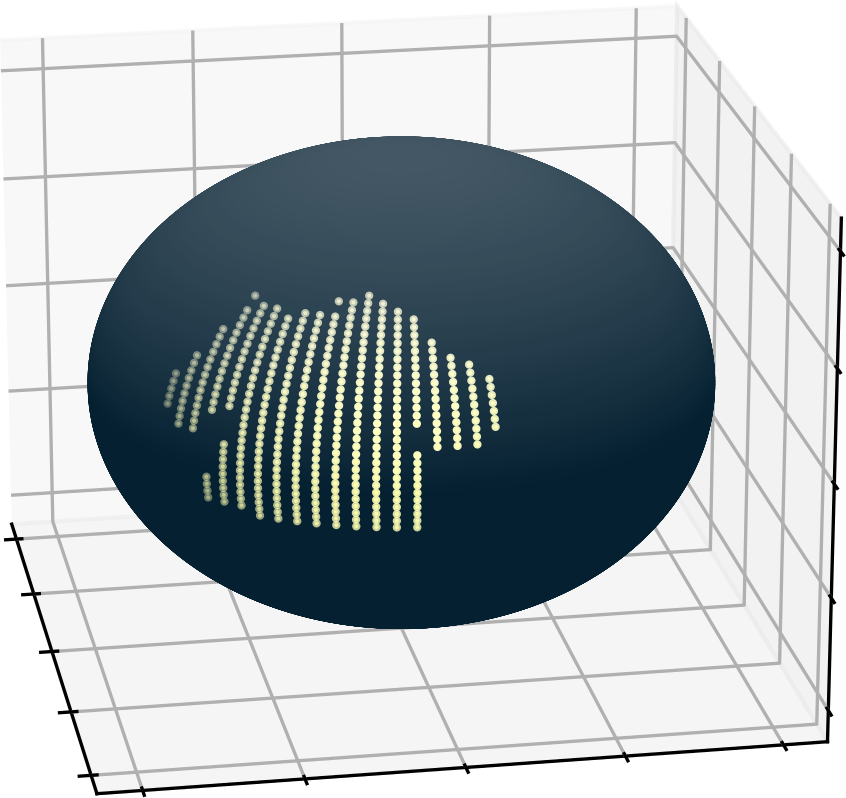}
  \hspace{0.3cm}
  \caption{Sample from the spherical FashionMNIST dataset. Left: input data. Right: segmentation mask.}
  \label{fig:sphericalMNIST_fashion}
\end{figure}
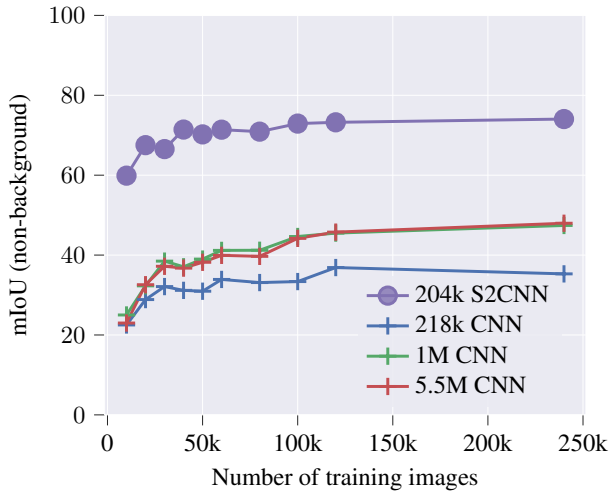
\begin{figure}[t]
  \centering
  \resizebox{\columnwidth}{!}{\input{imgs/spherical_data_augmentation_fashion.tex}}
  \caption{\textit{Segmentation on Spherical FashionMNIST}. Performance of equivariant and non-equivariant models in semantic segmentation for various amounts of data augmentation as in Figure~\ref{fig:data_augm_1_digit}, but with items of clothing from the FashionMNIST dataset projected onto the sphere.}
  \label{fig:data_augm_fashion}
\end{figure}

In addition to increasing the number of MNIST digits in each datapoint, we have also made the object recognition aspect more difficult by swapping the MNIST digits for items of clothing from the FasionMNIST dataset~\cite{xiao2017}. A sample datapoint is depicted in Figure~\ref{fig:sphericalMNIST_fashion}. The results shown in Figure~\ref{fig:data_augm_fashion} reflect the higher difficulty of this task: The performance of all models is lower compared to Figure~\ref{fig:data_augm_1_digit}. This is partly due to the difficulty of constructing segmentation masks for the items of clothing, as discussed in Appendix~\ref{sec:spherical-expmts}.
However, the equivariant models still outperform the non-equivariant models for all training data sizes that we tried by a large margin.

\subsection{Non-equivariant performance saturation}
\label{sec:satur-non-equiv}
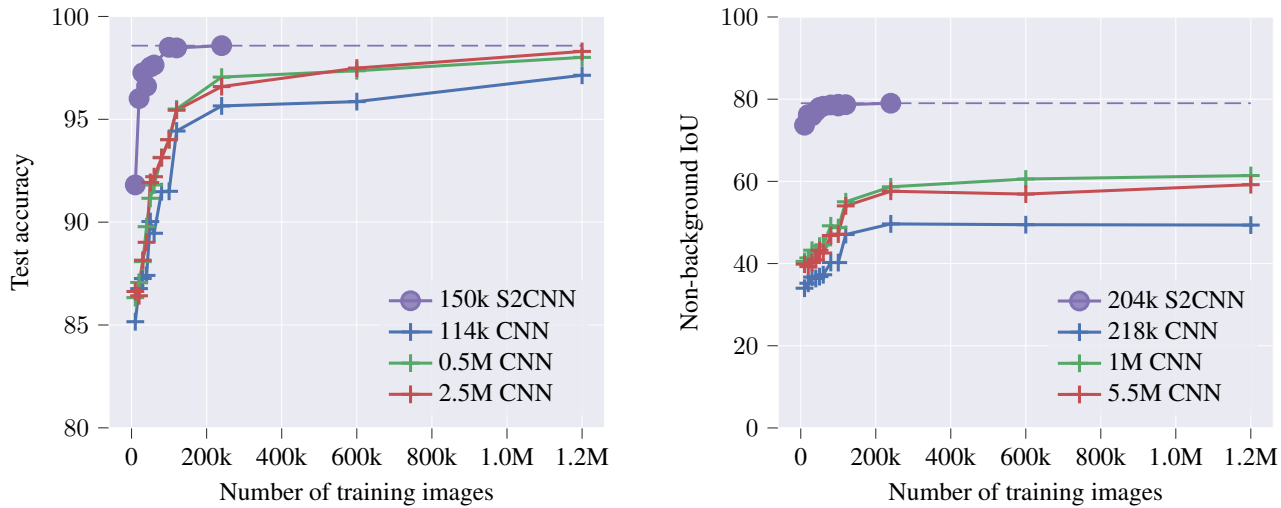
\begin{figure*}[t]
    \centering
    \resizebox{\columnwidth}{!}{\input{imgs/spherical_classification_data_augmentation_1_digit_full}}
    \hfill
    \resizebox{\columnwidth}{!}{\input{imgs/spherical_data_augmentation_1_digit_saturation.tex}}
    \caption{\textit{Non-equivariant performance saturation for segmentation}. Left: For classification of spherical MNIST as in Figure~\ref{fig:spherical_classification}, the non-equivariant models reach the test accuracy of the equivariant models for very large amounts of data augmentation. Right: For semantic segmentation of one-digit spherical MNIST as in Figure~\ref{fig:data_augm_1_digit}, the non-background IoU of the non-equivariant models saturates well below the performance of the equivariant model even for moderately high amounts of data augmentation.}
    \label{fig:saturation}
\end{figure*}

To check whether the non-equivariant models could be pushed to the performance of the equivariant models by extending the training data even further, we trained our non-equivariant models on larger datasets.

For spherical MNIST classification the non-equivariant models can match the accuracy of the smaller equivariant model with enough augmented data, cf.\ Figure~\ref{fig:saturation} left. On the other hand, for semantic segmentation,  as shown in Figure~\ref{fig:saturation} right, even these much larger training datasets did not improve the test performance of the non-equivariant models.

These experiments support the intuition that for equivariant tasks, equivariant models perform so much better than non-equivariant models that even very large amounts of data augmentation cannot compensate for this advantage. In contrast, although equivariant models (which were made invariant only in the last layer) still show higher performance for invariant tasks, non-equivariant models can ultimately reach the same performance if enough data augmentation is applied.

\subsection{Inference latency and training times}
\label{sec:profiling}
At similar parameter counts, the GPU based implementation \cite{cohen2018b} of the equivariant layers have an order of magnitude higher inference latency as can be seen in Tables~\ref{tab:spherical_performance} and \ref{tab:CNN_performance}.

A detailed profiling of the model and CUDA implementation shows that the bulk of the inference time is spent in the later upsampling layers which compute the largest $\SO(3)$ tensors in the network. In terms of operations, almost half of the time is spent in the custom implementation of the complex matrix multiplication of $\SO(3)$ tensors. There is also a significant overhead in the transformation back and forth to the Fourier domain. See Appendix~\ref{sec:latency_profiling} for the details on the profiling of layers and operations.

The backpropagation latency of the equivariant models mirrors the inference latency resulting in about an order of magnitude slower training as compared to the non-equivariant CNNs. Note that even though the training is slower, because of the increase in data augmentation needed for the non-equivariant models, at a fixed performance goal the equivariant model actually trains faster, cf.\ Table~\ref{tab:classification_training_times} for training times for the classification task at about 97.5\% accuracy level. A similar comparison for semantic segmentation is less relevant as even large amount of data augmentation leaves a big performance gap compared to the equivariant model.

\begin{table}[t]
\caption{Runtime latency and throughput for the equivariant semantic segmentation model (204k S2CNN in Appendix Table~\ref{tab:equiv_models}) on an Nvidia T4 16GB GPU. Latency measures the time of a forward pass through the model on the GPU, throughput is the corresponding number of samples per second given the batch size. The larger batch size is chosen to maximize the throughput for the T4.}
\label{tab:spherical_performance}
\begin{tabular}{lll}
Batch size & Latency (ms)  & Throughput (N/s) \\
\hline
1          & $111\pm{0.6}$ & $9.0\pm{0.04}$   \\
7          & $479\pm{2.2}$ & $14.6\pm{0.07}$
\end{tabular}
\end{table}

\begin{table}[t]
    \caption{Runtime latency and throughput for non-equivariant CNN model (200k CNN in Appendix Table~\ref{tab:equiv_models}) on an Nvidia T4 16GB GPU. Latency measures the time of a forward pass through the model on the GPU, throughput is the corresponding number of samples per second given the batch size. The larger batch size is chosen to maximize the throughput for the T4.}
\label{tab:CNN_performance}
\begin{tabular}{lll}
Batch size & Latency (ms)  & Throughput (N/s) \\
\hline
1 & $5.93 \pm{0.24} $ &  $169 \pm{5.8}$ \\
60 & $87.98 \pm{0.17}$ & $682 \pm{1.3}$
\end{tabular}
\end{table}

\begin{table}[t]
\centering
\caption{Training times for the S2CNN classification model and non-equivariant CNN model at matched accuracy on rotated spherical images. The S2CNN model is trained on non-rotated images whereas the CNN is trained on an augmented dataset with rotated images. A single Nvidia T4 16GB was used for training.}
\label{tab:classification_training_times}
\begin{tabular}{lll}
Model      & Accuracy & Training time \\
\hline
150k S2CNN & 97.64\%    & 15h           \\
5M CNN     & 97.49\%    & 26h
\end{tabular}
\end{table}

%%% Local Variables:
%%% mode: latex
%%% TeX-master: "mnist_paper"
%%% End:

%% file: imgs/spherical_classification_data_augmentation_1_digit.tex
% This file was created with tikzplotlib v0.9.12.
\begin{tikzpicture}

\definecolor{color0}{rgb}{0.917647058823529,0.917647058823529,0.949019607843137}
\definecolor{color1}{rgb}{0.505882352941176,0.447058823529412,0.698039215686274}
\definecolor{color2}{rgb}{0.298039215686275,0.447058823529412,0.690196078431373}
\definecolor{color3}{rgb}{0.333333333333333,0.658823529411765,0.407843137254902}
\definecolor{color4}{rgb}{0.768627450980392,0.305882352941176,0.32156862745098}

\begin{axis}[
axis background/.style={fill=color0},
axis line style={white},
legend cell align={left},
legend style={
  fill opacity=0.8,
  draw opacity=1,
  text opacity=1,
  at={(0.97,0.03)},
  anchor=south east,
  draw=none,
  fill=color0
},
tick align=outside,
tick pos=left,
x grid style={white},
xlabel={Number of training images},
xmajorgrids,
xmin=-1.5, xmax=251.5,
xtick style={color=white!15!black},
xtick={0,50,100,150,200,250},
xticklabels={0,50k,100k,150k,200k,250k},
y grid style={white},
ylabel={Test accuracy},
ymajorgrids,
ymin=80, ymax=100,
ytick style={color=white!15!black}
]
\addplot [very thick, color1, mark=*, mark size=3.5, mark options={solid}]
table {%
10 91.81
20 96.01
30 97.26
40 96.61
50 97.54
60 97.64
100 98.5
120 98.47
240 98.58
};
\addlegendentry{150k S2CNN}
\addplot [very thick, color2, mark=+, mark size=3.5, mark options={solid}]
table {%
10 85.16
20 86.77
30 87.26
40 87.41
50 90.03
60 89.46
80 91.48
100 91.5
120 94.43
240 95.65
};
\addlegendentry{114k CNN}
\addplot [very thick, color3, mark=+, mark size=3.5, mark options={solid}]
table {%
10 86.63
20 86.42
30 88.15
40 89.02
50 91.93
60 92.21
80 93.13
100 94.01
120 95.44
240 96.59
};
\addlegendentry{0.5M CNN}
\addplot [very thick, color4, mark=+, mark size=3.5, mark options={solid}]
table {%
10 86.33
20 87.07
30 88.08
40 89.78
50 91.16
60 91.81
80 93.15
100 94.01
120 95.5
240 97.05
};
\addlegendentry{2.5M CNN}
\end{axis}

\end{tikzpicture}

%% file: imgs/spherical_data_augmentation_non_rot_projected_on_grid_center_non_rot_eval.tex
% This file was created with tikzplotlib v0.9.12.
\begin{tikzpicture}

\definecolor{color0}{rgb}{0.917647058823529,0.917647058823529,0.949019607843137}
\definecolor{color1}{rgb}{0.505882352941176,0.447058823529412,0.698039215686274}
\definecolor{color2}{rgb}{0.298039215686275,0.447058823529412,0.690196078431373}
\definecolor{color3}{rgb}{0.333333333333333,0.658823529411765,0.407843137254902}
\definecolor{color4}{rgb}{0.768627450980392,0.305882352941176,0.32156862745098}

\begin{axis}[
axis background/.style={fill=color0},
axis line style={white},
legend cell align={left},
legend style={
  fill opacity=0.8,
  draw opacity=1,
  text opacity=1,
  at={(0.97,0.03)},
  anchor=south east,
  draw=none,
  fill=color0
},
tick align=outside,
tick pos=left,
x grid style={white},
xlabel={Number of unrotated training images},
xmajorgrids,
xmin=-1.5, xmax=251.5,
xtick style={color=white!15!black},
xtick={0,50,100,150,200,250},
xticklabels={0,50k,100k,150k,200k,250k},
y grid style={white},
ylabel={mIoU on Unrotated Images},
ymajorgrids,
ymin=0, ymax=100,
ytick style={color=white!15!black}
]
\addplot [very thick, color1, mark=*, mark size=3.5, mark options={solid}]
table {%
10 74.2677050380129
20 76.8594136096554
30 77.6652588021171
40 77.4544267995054
50 78.5149509435184
60 78.9408037774072
80 79.2532051489549
100 79.3389975715341
120 79.6200572759775
240 79.8134576137201
};
\addlegendentry{204k S2CNN}
\addplot [very thick, color2, mark=+, mark size=3.5, mark options={solid}]
table {%
10 73.58010400801
20 78.7023771991111
30 78.9041138752722
40 79.3057521588921
50 80.5108672546465
60 80.2258178334838
80 80.7878977638898
100 81.4627986946612
120 82.0625219880998
240 82.4408405267292
};
\addlegendentry{218k CNN}
\addplot [very thick, color3, mark=+, mark size=3.5, mark options={solid}]
table {%
10 78.2265176796211
20 80.535194868353
30 82.0018308336076
40 81.6911872451438
50 81.2831222653486
60 82.2505874229562
80 82.7425805840175
100 80.9219300755188
120 82.7111695574602
240 81.9464730390086
};
\addlegendentry{1M CNN}
\addplot [very thick, color4, mark=+, mark size=3.5, mark options={solid}]
table {%
10 78.0618706094855
20 81.3463741541022
30 83.3847572361571
40 84.2081055815584
50 84.6111215789895
60 84.4535715293519
80 85.2798405602758
100 85.8347079643411
120 86.1351850568458
240 87.1462695739364
};
\addlegendentry{5.5M CNN}
\end{axis}

\end{tikzpicture}

%% file: imgs/spherical_data_augmentation_non_rot_projected_on_grid_center_rot_eval.tex
% This file was created with tikzplotlib v0.9.12.
\begin{tikzpicture}

\definecolor{color0}{rgb}{0.917647058823529,0.917647058823529,0.949019607843137}
\definecolor{color1}{rgb}{0.505882352941176,0.447058823529412,0.698039215686274}
\definecolor{color2}{rgb}{0.298039215686275,0.447058823529412,0.690196078431373}
\definecolor{color3}{rgb}{0.333333333333333,0.658823529411765,0.407843137254902}
\definecolor{color4}{rgb}{0.768627450980392,0.305882352941176,0.32156862745098}

\begin{axis}[
axis background/.style={fill=color0},
axis line style={white},
legend cell align={left},
legend style={
  fill opacity=0.8,
  draw opacity=1,
  text opacity=1,
  at={(0.91,0.5)},
  anchor=east,
  draw=none,
  fill=color0
},
legend style={at={(0.98,0.65)}, anchor=north east},
tick align=outside,
tick pos=left,
x grid style={white},
xlabel={Number of unrotated training images},
xmajorgrids,
xmin=-1.5, xmax=251.5,
xtick style={color=white!15!black},
xtick={0,50,100,150,200,250},
xticklabels={0,50k,100k,150k,200k,250k},
y grid style={white},
ylabel={mIoU on Rotated Images},
ymajorgrids,
ymin=0, ymax=100,
ytick style={color=white!15!black}
]
\addplot [very thick, color1, mark=*, mark size=3.5, mark options={solid}]
table {%
10 75.1587116491286
20 77.2851588144363
30 77.8402581084608
40 77.7334137544592
50 78.4878156072444
60 78.5875963391494
80 78.5976902440102
100 78.7177660313421
120 78.7853338790605
240 78.9656905254994
};
\addlegendentry{204k S2CNN}
\addplot [very thick, color2, mark=+, mark size=3.5, mark options={solid}]
table {%
10 15.4105544086955
20 11.1467325010807
30 8.87570742592795
40 11.5564898197759
50 10.3677151806773
60 13.9558447331002
80 11.2701381964532
100 13.9038172230294
120 8.72533194444059
240 13.4127458878852
};
\addlegendentry{218k CNN}
\addplot [very thick, color3, mark=+, mark size=3.5, mark options={solid}]
table {%
10 15.5062010399999
20 9.55727127915249
30 10.7656904274824
40 10.9923507777807
50 11.0679791346708
60 14.6015120060682
80 10.749363031772
100 14.4085376092666
120 7.95947345375563
240 13.3799775522978
};
\addlegendentry{1M CNN}
\addplot [very thick, color4, mark=+, mark size=3.5, mark options={solid}]
table {%
10 13.2053186386201
20 10.14357604628
30 9.75132015740805
40 11.0343348734973
50 9.92196461811875
60 13.1360583893131
80 9.52646708318885
100 13.4541821036274
120 9.22203602195639
240 15.227933881533
};
\addlegendentry{5.5M CNN}
\end{axis}

\end{tikzpicture}

%% file: imgs/spherical_data_augmentation_4_digits.tex
% This file was created with tikzplotlib v0.9.12.
\begin{tikzpicture}

\definecolor{color0}{rgb}{0.917647058823529,0.917647058823529,0.949019607843137}
\definecolor{color1}{rgb}{0.505882352941176,0.447058823529412,0.698039215686274}
\definecolor{color2}{rgb}{0.298039215686275,0.447058823529412,0.690196078431373}
\definecolor{color3}{rgb}{0.333333333333333,0.658823529411765,0.407843137254902}
\definecolor{color4}{rgb}{0.768627450980392,0.305882352941176,0.32156862745098}

\begin{axis}[
axis background/.style={fill=color0},
axis line style={white},
legend cell align={left},
legend style={
  fill opacity=0.8,
  draw opacity=1,
  text opacity=1,
  at={(0.97,0.03)},
  anchor=south east,
  draw=none,
  fill=color0
},
tick align=outside,
tick pos=left,
x grid style={white},
xlabel={Number of training images},
xmajorgrids,
xmin=-1.5, xmax=251.5,
xtick style={color=white!15!black},
xtick={0,50,100,150,200,250},
xticklabels={0,50k,100k,150k,200k,250k},
y grid style={white},
ylabel={mIoU (non-background)},
ymajorgrids,
ymin=0, ymax=100,
ytick style={color=white!15!black}
]
\addplot [very thick, color1, mark=*, mark size=3.5, mark options={solid}]
table {%
10 73.6574362848322
20 76.0395833736555
30 76.7890055815408
40 77.5328788714405
50 77.020082794026
60 77.8381367155123
80 78.2197165462477
100 77.9684623342617
120 78.1636630549863
240 78.3261777832088
};
\addlegendentry{204k S2CNN}
\addplot [very thick, color2, mark=+, mark size=3.5, mark options={solid}]
table {%
10 32.7623520492889
20 39.785570381459
30 45.6672944820466
40 43.6034531633931
50 52.9660619529629
60 47.3842111139795
80 50.7463134189105
100 46.5693888014821
120 54.5243203801985
240 55.2213802508576
};
\addlegendentry{218k CNN}
\addplot [very thick, color3, mark=+, mark size=3.5, mark options={solid}]
table {%
10 35.9387544826403
20 46.3409641066674
30 50.8084672150865
40 48.5435259540383
50 58.9691012445147
60 54.1634577116196
80 59.2485369254248
100 55.0810140666171
120 61.8213733749629
240 62.4269677407072
};
\addlegendentry{1M CNN}
\addplot [very thick, color4, mark=+, mark size=3.5, mark options={solid}]
table {%
10 31.7295263691858
20 41.5629787039001
30 47.0335572651033
40 44.6008543255815
50 56.2212101792903
60 50.878548185644
80 57.4930188112352
100 52.2904181758365
120 60.3875023574351
240 61.5921850677986
};
\addlegendentry{5.5M CNN}
\end{axis}

\end{tikzpicture}

%% file: imgs/spherical_data_augmentation_fashion.tex
% This file was created with tikzplotlib v0.9.12.
\begin{tikzpicture}

\definecolor{color0}{rgb}{0.917647058823529,0.917647058823529,0.949019607843137}
\definecolor{color1}{rgb}{0.505882352941176,0.447058823529412,0.698039215686274}
\definecolor{color2}{rgb}{0.298039215686275,0.447058823529412,0.690196078431373}
\definecolor{color3}{rgb}{0.333333333333333,0.658823529411765,0.407843137254902}
\definecolor{color4}{rgb}{0.768627450980392,0.305882352941176,0.32156862745098}

\begin{axis}[
axis background/.style={fill=color0},
axis line style={white},
legend cell align={left},
legend style={at={(0.95,0.35)}, anchor=north east},
legend style={fill opacity=0.8, draw opacity=1, text opacity=1, draw=none, fill=color0},
tick align=outside,
tick pos=left,
x grid style={white},
xlabel={Number of training images},
xmajorgrids,
xmin=-1.5, xmax=251.5,
xtick style={color=white!15!black},
xtick={0,50,100,150,200,250},
xticklabels={0,50k,100k,150k,200k,250k},
y grid style={white},
ylabel={mIoU (non-background)},
ymajorgrids,
ymin=0, ymax=100,
ytick style={color=white!15!black}
]
\addplot [very thick, color1, mark=*, mark size=3.5, mark options={solid}]
table {%
10 59.8833690094129
20 67.5175774465214
30 66.5316322772944
40 71.4203271233446
50 70.2112386026664
60 71.3907155760861
80 70.8977203195638
100 72.9155460697253
120 73.2339063854297
240 74.0517217575614
};
\addlegendentry{204k S2CNN}
\addplot [very thick, color2, mark=+, mark size=3.5, mark options={solid}]
table {%
10 22.5082698002173
20 28.8614744187404
30 32.1149379389629
40 31.1702102926083
50 30.9537364928743
60 33.9032943442777
80 33.119978228583
100 33.3696622002457
120 36.8788733578608
240 35.2961755666738
};
\addlegendentry{218k CNN}
\addplot [very thick, color3, mark=+, mark size=3.5, mark options={solid}]
table {%
10 25.0082077407303
20 32.2951965242738
30 38.4856474495207
40 37.0713883220831
50 38.9962159729059
60 41.1757407279295
80 41.2057450235803
100 44.6504887109477
120 45.4971268810286
240 47.412395566588
};
\addlegendentry{1M CNN}
\addplot [very thick, color4, mark=+, mark size=3.5, mark options={solid}]
table {%
10 22.98938910436
20 32.5826745956069
30 37.1956010977744
40 36.7357742160776
50 38.1971730854776
60 39.9339397058471
80 39.6827163021231
100 44.1857118136206
120 45.7469042760396
240 47.968625028129
};
\addlegendentry{5.5M CNN}
\end{axis}

\end{tikzpicture}

%% file: imgs/spherical_classification_data_augmentation_1_digit_full.tex
% This file was created with tikzplotlib v0.9.12.
\begin{tikzpicture}

\definecolor{color0}{rgb}{0.917647058823529,0.917647058823529,0.949019607843137}
\definecolor{color1}{rgb}{0.505882352941176,0.447058823529412,0.698039215686274}
\definecolor{color2}{rgb}{0.298039215686275,0.447058823529412,0.690196078431373}
\definecolor{color3}{rgb}{0.333333333333333,0.658823529411765,0.407843137254902}
\definecolor{color4}{rgb}{0.768627450980392,0.305882352941176,0.32156862745098}

\begin{axis}[
axis background/.style={fill=color0},
axis line style={white},
legend cell align={left},
legend style={
  fill opacity=0.8,
  draw opacity=1,
  text opacity=1,
  at={(0.97,0.03)},
  anchor=south east,
  draw=none,
  fill=color0
},
tick align=outside,
tick pos=left,
x grid style={white},
xlabel={Number of training images},
xmajorgrids,
xmin=-60, xmax=1260,
xtick style={color=white!15!black},
xtick={0,200,400,600,800,1000,1200},
xticklabels={0,200k,400k,600k,800k,1.0M,1.2M},
y grid style={white},
ylabel={Test accuracy},
ymajorgrids,
ymin=80, ymax=100,
ytick style={color=white!15!black}
]
\path [draw=color1, line width=0.7pt, dash pattern=on 6.475pt off 2.8pt]
(axis cs:0,98.58)
--(axis cs:1200,98.58);

\addplot [very thick, color1, mark=*, mark size=3.5, mark options={solid}]
table {%
10 91.81
20 96.01
30 97.26
40 96.61
50 97.54
60 97.64
100 98.5
120 98.47
240 98.58
};
\addlegendentry{150k S2CNN}
\addplot [very thick, color2, mark=+, mark size=3.5, mark options={solid}]
table {%
10 85.16
20 86.77
30 87.26
40 87.41
50 90.03
60 89.46
80 91.48
100 91.5
120 94.43
240 95.65
600 95.86
1200 97.14
};
\addlegendentry{114k CNN}
\addplot [very thick, color3, mark=+, mark size=3.5, mark options={solid}]
table {%
10 86.33
20 87.07
30 88.08
40 89.78
50 91.16
60 91.81
80 93.15
100 94.01
120 95.5
240 97.05
600 97.36
1200 98.01
};
\addlegendentry{0.5M CNN}
\addplot [very thick, color4, mark=+, mark size=3.5, mark options={solid}]
table {%
10 86.63
20 86.42
30 88.15
40 89.02
50 91.93
60 92.21
80 93.13
100 94.01
120 95.44
240 96.59
600 97.49
1200 98.3
};
\addlegendentry{2.5M CNN}
\end{axis}

\end{tikzpicture}

%% file: imgs/spherical_data_augmentation_1_digit_saturation.tex
% This file was created with tikzplotlib v0.9.12.
\begin{tikzpicture}

\definecolor{color0}{rgb}{0.917647058823529,0.917647058823529,0.949019607843137}
\definecolor{color1}{rgb}{0.505882352941176,0.447058823529412,0.698039215686274}
\definecolor{color2}{rgb}{0.298039215686275,0.447058823529412,0.690196078431373}
\definecolor{color3}{rgb}{0.333333333333333,0.658823529411765,0.407843137254902}
\definecolor{color4}{rgb}{0.768627450980392,0.305882352941176,0.32156862745098}

\begin{axis}[
axis background/.style={fill=color0},
axis line style={white},
legend cell align={left},
legend style={
  fill opacity=0.8,
  draw opacity=1,
  text opacity=1,
  at={(0.97,0.03)},
  anchor=south east,
  draw=none,
  fill=color0
},
tick align=outside,
tick pos=left,
x grid style={white},
xlabel={Number of training images},
xmajorgrids,
xmin=-0.06, xmax=1.26,
xtick style={color=white!15!black},
xtick={0,0.2,0.4,0.6,0.8,1,1.2},
xticklabels={0,200k,400k,600k,800k,1.0M,1.2M},
y grid style={white},
ylabel={Non-background IoU},
ymajorgrids,
ymin=0, ymax=100,
ytick style={color=white!15!black}
]
\path [draw=color1, line width=0.7pt, dash pattern=on 6.475pt off 2.8pt]
(axis cs:0,79.0172031795914)
--(axis cs:1.2,79.0172031795914);

\addplot [very thick, color1, mark=*, mark size=3.5, mark options={solid}]
table {%
0.01 73.6672228087486
0.02 76.2396656837297
0.03 75.9440592307601
0.04 77.0006140071276
0.05 78.0121633327085
0.06 78.2821354107209
0.08 78.6005328484791
0.1 78.3440466722201
0.1 78.7591833919464
0.12 78.6653772927548
0.24 79.0172031795914
};
\addlegendentry{204k S2CNN}
\addplot [very thick, color2, mark=+, mark size=3.5, mark options={solid}]
table {%
0.01 33.9979413230564
0.02 35.2098953469084
0.03 36.6968152116464
0.04 36.3691463001431
0.05 36.8220698817118
0.06 37.2769826449922
0.08 40.2635431161527
0.1 40.2390027250844
0.12 47.0656347978214
0.24 49.6463416406717
0.6 49.4607500191341
1.2 49.3667478680225
};
\addlegendentry{218k CNN}
\addplot [very thick, color3, mark=+, mark size=3.5, mark options={solid}]
table {%
0.01 40.5212931678065
0.02 41.377917930675
0.03 43.251758735901
0.04 42.6359169287997
0.05 44.2412785935146
0.06 44.5472077883157
0.08 49.1886897951291
0.1 48.7415286074443
0.12 55.0321060520411
0.24 58.6775255398797
0.6 60.5768117291
1.2 61.4187406860649
};
\addlegendentry{1M CNN}
\addplot [very thick, color4, mark=+, mark size=3.5, mark options={solid}]
table {%
0.01 39.8560257098186
0.02 39.2542836381079
0.03 40.2511386267728
0.04 40.9095744779148
0.05 43.239519380981
0.06 42.5646759801113
0.08 46.8256498871211
0.1 47.1734684919231
0.12 54.0339057646098
0.24 57.5987507390389
0.6 56.9030047900288
1.2 59.1977968380171
};
\addlegendentry{5.5M CNN}
\end{axis}

\end{tikzpicture}

%% file: section_conclusions.tex
\section{Conclusions}
\label{sec:conclusions}

Our results indicate that equivariant models possess an inherent advantage over non-equivariant ones, which cannot be overcome by data augmentation when applied to tasks probing the full rotational equivariance of the spherical image data. In order to corroborate and generalize these findings, several extensions of the current study are natural to pursue. In particular, it would be interesting to consider non-equivariant models specifically adapted to the sphere.

Furthermore, even though our results indicate that the advantage of equivariance is not explained by low data complexity, it would be interesting to investigate data sets which are both richer in complexity and native to the sphere rather than projected onto it.

In terms of inference latency the CUDA implementation of the equivariant convolution in the Fourier domain, together with the large $\SO(3)$ tensors, is significantly slower than a traditional spatial convolution and the profiling shows where future optimizations should be targeted. With wider adoption it is likely that this situation would improve on multiple fronts.

 It is  appealing to think of symmetries as a fundamental design principle for network architectures. In this paper we use the symmetry of the sphere as a guiding principle for the network architecture. From this perspective it is natural to  consider the question of how to train neural networks in the case of other ``non-flat'' data manifolds, i.e. when the domain $\mathcal{M}$ is a (possibly curved) manifold. This research field is referred to as \emph{geometric deep learning}, an umbrella term first coined in  \cite{LeCunGeometric} (see~\cite{bronstein2021} and \cite{gerkenGeometricDeepLearning2021} for recent  reviews). The results of the present paper may thus be viewed as probing a small corner of the vast field of geometric deep learning.

Extending the exploration of inherent advantages of equivariance to other tasks, data manifolds and, possibly local, symmetry groups offers exciting prospects for future research.

\section*{Acknowledgments}
\noindent We are very grateful to Jimmy Aronsson for valuable discussions and helpful feedback on the text. We also thank the anonymous referees for their comments.

 The work of D.P. and O.C. is supported by the Wallenberg AI, Autonomous Systems and Software Program
(WASP) funded by the Knut and Alice Wallenberg Foundation. D.P. and J.G. are supported
by the Swedish Research Council, J.G. is also supported by the Knut and Alice Wallenberg Foundation and by the German Ministry for Education and Research (BMBF) under Grants 01IS14013A-E, 01GQ1115, 1GQ0850, 01IS18025A and 01IS18037A.

The computations were enabled by
resources provided by the Swedish National Infrastructure for
Computing (SNIC) at C3SE partially funded by the Swedish
Research Council through grant agreement no. 2018-05973.

%% file: appendix_proj_so3_s2.tex
\section{Mathematical properties of the final S2CNN layer}
\label{sec:projection-so3-s2}

In this appendix, we show the equivariance of the final S2CNN layer used for semantic segmentation \eqref{eq:6} and give an interpretation in terms of the projection \eqref{eq:10} from $G$ to $G/H$. For convenience, we reproduce \eqref{eq:6} here,
\begin{align}
  f^{\text{final}}(x)=\sum_{\ell=0}^{L}\sum_{m=-\ell}^{\ell}\sum_{n=-\ell}^{\ell}(\kappa\star f)^{\ell}_{mn}Y^{\ell}_{m}(x)\,.\label{eq:11}
\end{align}

\subsection{Proof of equivariance}
Using basic properties of Wigner matrices and spherical harmonics, we find for $R\in\SO(3)$
\begin{align}
  f^{\text{final}}(Rx) &= \sum_{\ell=0}^{L}\sum_{m,n=-\ell}^{\ell}(\kappa\star f)^{\ell}_{mn}Y^{\ell}_{m}(Rx)\\
  &\hspace{-2.5em}=\sum_{\ell=0}^{L}\sum_{m,n,k=-\ell}^{\ell}(\kappa\star f)^{\ell}_{mn}\overline{\mathcal{D}^{\ell}_{mk}(R)}Y^{\ell}_{k}(x)\label{eq:14}\\
  &\hspace{-2.5em}=\sum_{\ell=0}^{L}\sum_{m,n,k=-\ell}^{\ell}(\kappa\star f)^{\ell}_{mn}\mathcal{D}^{\ell}_{km}(R^{-1})Y^{\ell}_{k}(x)\,,\label{eq:12}
\end{align}
where in \eqref{eq:14} we used the transformation property of the spherical harmonics and in \eqref{eq:12} that the Wigner matrices form a unitary representation. For a summary of the necessary formulae and the conventions\footnote{We want to reiterate that we use the complex conjugate conventions for Wigner matrices in comparison to the reference.}, see, e.g.,  \cite{gerkenGeometricDeepLearning2021}.

Using the definition of the Fourier coefficients on $\SO(3)$, we obtain
\begin{align}
  &\sum_{m=-\ell}^{\ell}(\kappa\star f)^{\ell}_{mn}\mathcal{D}^{\ell}_{km}(R^{-1})\nonumber\\
  =&\sum_{m=-\ell}^{\ell}\frac{2\ell+1}{8\pi^{2}}\int_{\SO(3)}(\kappa\star f)(S) \mathcal{D}^{\ell}_{mn}(S)\mathcal{D}^{\ell}_{km}(R^{-1})\dd{S}\nonumber\\
  =&\frac{2\ell+1}{8\pi^{2}}\int_{\SO(3)}(\kappa\star f)(S) \mathcal{D}^{\ell}_{kn}(R^{-1}S)\dd{S}\nonumber\\
  =&\frac{2\ell+1}{8\pi^{2}}\int_{\SO(3)}(\kappa\star f)(RS') \mathcal{D}^{\ell}_{kn}(S')\dd{S'}\nonumber\\
  =&(\kappa\star L_{R}f)^{\ell}_{kn}\,,\label{eq:13}
\end{align}
where we have used the notation $(L_{R}(f))(Q)=f(RQ)$ and the equivariance of \eqref{eq:5} in the last step.

Plugging \eqref{eq:13} into \eqref{eq:12} completes the prove of equivariance of the last layer
\begin{align}
  f^{\text{final}}(Rx) = \sum_{\ell=0}^{L}\sum_{m,n=-\ell}^{\ell} (\kappa\star L_{R}f)^{\ell}_{mn}Y^{\ell}_{m}(x)\,.
\end{align}

\subsection{Projection from $\SO(3)$ to $S^{2}$}
To see the connection between \eqref{eq:11} and the integration over $H$ (in this case $H=\SO(2)$) in \eqref{eq:10}, we rewrite \eqref{eq:11} in position space
\begin{align}
  f^{\text{final}}(x)&=\sum_{\ell=0}^{L}\sum_{m=-\ell}^{\ell}\sum_{n=-\ell}^{\ell}(\kappa\star f)^{\ell}_{mn}Y^{\ell}_{m}(x)\\
  &\hspace{-3em}=\sum_{\ell=0}^{L}\sum_{m,n=-\ell}^{\ell}\int_{\SO(3)}(\kappa\star f)(S)\mathcal{D}^{\ell}_{mn}(S)Y^{\ell}_{m}(x)\dd{S}\nonumber\\
  &\hspace{-3em}=\sum_{\ell=0}^{L}\sum_{n=-\ell}^{\ell}\frac{2\ell+1}{8\pi^{2}}\int_{\SO(3)}\!(\kappa\star f)(S)Y^{\ell}_{n}(S^{-1}x)\dd{S}\,.\nonumber
\end{align}
Now, we factorize $S$ into an element $\alpha\in\SO(2)$ which stabilizes $x$ and an element $y\in S^{2}$. This can always be done uniquely. With this, we obtain
\begin{align}
  f^{\text{final}}(x)=\int_{S^{2}}\!\widetilde{(\kappa\star f)}(y)\sum_{\ell=0}^{L}\sum_{n=-\ell}^{\ell}\frac{2\ell+1}{8\pi^{2}}Y^{\ell}_{n}(y)\dd{y}\label{eq:15}
\end{align}
where we have defined
\begin{align}
  \widetilde{(\kappa\star f)}(y)=\int_{\SO(2)}\!(\kappa\star f)(y,\alpha)\dd{\alpha}\,.\label{eq:16}
\end{align}
From \eqref{eq:15}, we can conclude that \eqref{eq:11} can be understood as the projected version of $\kappa\star f$, \eqref{eq:16}, integrated against the function $\sum_{\ell=0}^{L}\sum_{n=-\ell}^{\ell}\frac{2\ell+1}{8\pi^{2}}Y^{\ell}_{n}$.

%%% Local Variables:
%%% mode: latex
%%% TeX-master: "mnist_paper"
%%% End:

%% file: appendix_model_generation.tex
\section{Random model generation}
\label{sec:spherical-model-gen}

In this appendix, we give further details about the procedure used to randomly generate the model architectures we used in our experiments.

We generated 20 random models for each desired parameter range, trained it to convergence, evaluated it on a holdout dataset and picked the best model according to the non-background mIoU.

It should be noted that the goal of the procedure is not to come up with a selection of models that perform optimally on the given task for a certain parameter budget, but rather to select architectures to fairly compare equivariant and non-equivariant models.

\subsection{Non-equivariant models}
\label{sec:non-equiv-model-gen}
\begin{table*}
  \centering
    \caption{Non-equivariant model architectures used in our experiments. The up- and down-sampling layers are defined in~\eqref{eq:2} and \eqref{eq:3}, respectively. The columns labeled as ``output'' contain the channel- and spatial dimensions of the output feature maps.}
  \label{tab:non_equiv_models}
  \begin{tabular}{rcccccc}
    \toprule
    Model & \multicolumn{2}{c}{218k CNN} & \multicolumn{2}{c}{1M CNN} & \multicolumn{2}{c}{5.5M CNN} \\
    & block & output & block & output & block & output \\
    \midrule
    & input & $1{\times}100^2$ & input  & $1{\times}100^2$ & input  & $1{\times}100^2$\\
    Down & $\conv(1{,}13{,}5{,}1)$ & $13{\times}96^2$ & $\conv(1{,}12{,}3{,}1)$ & $12{\times}98^2$ & $\conv(1{,}12{,}3{,}1)$ & $12{\times}98^2$\\
    & $\down(13{,}15{,}3{,}1)$ & $15{\times}94^2$ & $\down(12{,}13{,}3{,}1)$ & $13{\times}96^2$ & $\down(12{,}15{,}5,{,}1)$ & $15{\times}94^2$\\
    & $\down(15{,}22{,}9{,}1)$ & $22{\times}86^2$ & $\down(13{,}16{,}5{,}1)$ & $16{\times}92^2$ & $\down(15{,}16{,}3{,}1)$ & $16{\times}92^2$\\
    & $\down(22{,}31{,}7{,}1)$ & $31{\times}80^2$ & $\down(16{,}77{,}5{,}2)$ & $77{\times}92^2$ & $\down(16{,}85{,}7{,}2)$ & $85{\times}43^2$\\
    & $\down(31{,}141{,}3{,}2)$ & $141{\times}39^2$ & $\down(77{,}96{,}3{,}1)$ & $96{\times}42^2$& $\down(85{,}191{,}3{,}1)$ & $191{\times}39^2$\\
    & & & $\down(96{,}163{,}5{,}1)$ & $163{\times}38^2$ & $\down(191{,}191{,}3{,}1)$ & $191{\times}37^2$\\
    & & & & & $\down(191{,}1100{,}3{,}2)$ & $1100{\times}18^2$\\
    \midrule
    Up & $\up(141{,}31{,} 3{,} 2)$ & $31{\times}80^2$ & $\up(163{,}96{,}5{,}1)$& $96{\times}42^2$ & $\up(1100{,}191{,}3{,}2)$ & $191{\times}37^2$\\
    & $\up(31{,}22{,}7{,}1)$ & $22{\times}86^2$ & $\up(96{,}77{,}3{,}1)$ & $77{\times}44^2$ & $\up(191{,}141{,}3{,}1)$ & $141{\times}39^2$\\
    & $\up(22{,}15{,}9{,}1)$ & $15{\times}94^2$ & $\up(77{,}16{,}5{,}2)$ & $16{\times}92^2$ & $\up(141{,}85{,}5{,}1)$ & $85{\times}43^2$\\
    & $\up(15{,}13{,}3{,}1)$ & $13{\times}96^2$ & $\up(16{,}13{,}5{,}1)$ & $13{\times}96^2$ & $\up(85{,}16{,}7{,}2)$ & $16{\times}92^2$\\
    & $\tConv(13{,}11{,} 5{,} 1)$ & $11{\times}100^2$ & $\up(13{,}12{,}3{,}1)$ & $12{\times}98^2$ & $\up(16{,}15{,}3{,}1)$ & $15{\times}94^2$\\
    & & & $\tConv(12{,}11{,}3{,}1)$ & $11{\times}100^2$& $\up(15{,}12{,}5{,}1)$ & $12{\times}98^2$\\
    & & & & & $\tConv(12{,}11{,}3{,}1)$ & $11{\times}100^2$\\
    \midrule
    Params. & \multicolumn{2}{c}{\num{218144}} & \multicolumn{2}{c}{\num{1042013}} & \multicolumn{2}{c}{\num{5519335}} \\
    \bottomrule
  \end{tabular}
\end{table*}

As mentioned in the main text, the non-equivariant models consist of skipped convolutional downsampling and skipped transposed convolutional upsampling layers. Specifically, the downsampling layers are defined by
\begin{align}
  \down(n_{i},n_{o},k,s) =& \conv(n_{i},n_{o},k,s)\label{eq:2}\\
  &+\conv(n_{i},n_{o}{,}1{,}1)\circ \maxPool(k{,}s)\,,\nonumber
\end{align}
where $\conv(n_{i},n_{o},k,s)$ is a 2d convolution with $n_{i}$ input channels, $n_{o}$ output channels, kernel size $k$ and stride $s$, $\circ$ denotes function composition and $\maxPool(k,s)$ is a 2d max pooling operation with kernel size $k$ and stride $s$. The upsampling layers are defined by
\begin{align}
  \up(n_{i},n_{o},k,s) =& \tConv(n_{i},n_{o},k,s)\label{eq:3}\\
  &+\upsample(d)\circ\conv(n_{i},n_{o}{,}1{,}1)\,,\nonumber
\end{align}
where $\tConv(n_{i},n_{o},k,s)$ is a 2d transpose convolution with $n_{i}$ input channels, $n_{o}$ output channels, kernel size $k$ and stride $s$ and $\upsample(d)$ is a 2d upsampling layer to a feature tensor with spatial resolution $d\times d$ and nearest neighbor interpolation. In \eqref{eq:3}, $d$ is chosen such that the spatial dimensions of both summands match. After each down- and upsampling layer, we apply a ReLU nonlinearity.

We randomly select the number of layers and their channels, the kernel sizes and the strides in the following way: First, we select the depth of the network between 1 and $\lfloor \hat{p}/(2\cdot 10^{4}) \rfloor$, where $\hat{p}$ is the upper limit of the parameter range. Then, in order to obtain an hourglass-shape for the network, we select the size of the image dimension at the bottleneck between 2 and 30 and linearly interpolate from that to the size of the input (and output) image. These interpolated image sizes are the target dimensions $d^{\mathrm{target}}_{i}$ which we then try to approximate by choosing the kernel sizes and strides of the convolutions appropriately. This is done only for the downsampling layers as the upsampling layers are set to exactly mirror the downsampling architecture.

The kernel sizes $k_{i}$ are selected at random to be odd integers between 1 and 9. If the image in layer $i$ has smaller size than $9\times9$, we take instead the highest odd integer below the image dimension as the upper bound for the random choice. The strides $s_{i}$ are then computed from the target dimension $d_{i}^{\mathrm{target}}$, the input dimension $d_{i}^{\mathrm{in}}$ of the layer and the kernel size $k_{i}$ according to
\begin{align}
  s_{i}=\mathrm{round}\Big(\frac{d^{\mathrm{in}}_{i}-k_{i}}{d^{\mathrm{target}}_{i}-1}\Big)\,.\label{eq:1}
\end{align}
If the number generated by \eqref{eq:1} is 0, we set the stride to 1 instead. The actual output dimensions are then given by
\begin{align}
  d^{\mathrm{in}}_{i+1}=\frac{d^{\mathrm{in}}_{i}-k_{i}}{s_{i}}+1\,.
\end{align}

In order to keep the total number of features roughly constant across layers, we set the number of channels $N_{i}^{c}$ to
\begin{align}
  N_{i}^{c}=\left\lceil\frac{f}{(d^{\mathrm{in}}_{i+1})^{2}}\right\rceil\,,
\end{align}
where $f=11\cdot(2\cdot L)^{2}$ is the number of features in the final layer and $L$ is the bandwidth of the input (and output) data.

During the model generation process, architectures are generated according to the procedure summarized above and the total number of parameters for each architecture is computed. If the parameter count lies in the desired range, the model is accepted, otherwise it is rejected and a new model is generated.

All models are trained with batch size 32 and learning rate $10^{-3}$ using Adam on a segmentation task with one MNIST digit on the sphere and 60k rotated training samples until convergence and then evaluated. In all experiments, we use early stopping on the non-background mIoU metric and a maximum of 100 epochs.

In this way, for each desired parameter range, 20 models were trained and evaluated, we then picked the best performing models (according to the non-background mIoU) and used them for our experiments. The resulting architectures of the three non-equivariant models are summarized in Table~\ref{tab:non_equiv_models}.

\subsection{Equivariant models}
\label{sec:equiv-model-gen}
\begin{table*}
  \centering
    \caption{Equivariant model architectures used in our experiments. The notation for the layers is introduced in \eqref{eq:7}--\eqref{eq:9}. Note that the product of the \emph{input} bandwidth and $\hat{\beta}$ stays constant throughout the network. The columns labeled as ``output'' contain the channel- and spatial dimensions of the output feature maps.}
  \label{tab:equiv_models}
  \begin{tabular}{rcccc}
    \toprule
    Model & \multicolumn{2}{c}{204k S2CNN} & \multicolumn{2}{c}{820k S2CNN}\\
    & block & output & block & output \\
    \midrule
    & input & $1{\times}100^2$ & input & $1{\times}100^{2}$ \\
    Down & $\StwoSOthreeconv(1,11,42,0.1238\cdot\pi)$ & $11{\times}84^3$ & $\StwoSOthreeconv(1,13,43,0.1881\cdot\pi)$ & $13{\times}86^{3}$\\
    & $\SOthreeconv(11,12,35,0.1474\cdot\pi)$ & $12{\times}70^3$ & $\SOthreeconv(13,16,36,0.2187\cdot\pi)$ & $16{\times}72^{3}$\\
    & $\SOthreeconv(12,13,27,0.1768\cdot\pi)$ & $13{\times}54^3$& $\SOthreeconv(16,19,30,0.2613\cdot\pi)$ & $19{\times}60^{3}$\\
    & $\SOthreeconv(13,14,20,0.2292\cdot\pi)$ & $14{\times}40^3$& $\SOthreeconv(19,21,23,0.3135\cdot\pi)$ & $21{\times}46^{3}$\\
    & & & $\SOthreeconv(21,24,16,0.4089\cdot\pi)$ & $24{\times}32^{3}$\\
    & & & $\SOthreeconv(24,27,10,0.5878\cdot\pi)$ & $27{\times}20^{3}$\\
    \midrule
    Up & $\SOthreeconv(14,13,27,0.3095\cdot\pi)$ & $13{\times}54^3$ & $\SOthreeconv(27,24,16,0.9405\cdot\pi)$ & $24{\times}32^{3}$ \\
    & $\SOthreeconv(13,12,35,0.2292\cdot\pi)$ & $12{\times}70^3$& $\SOthreeconv(24,21,23,0.5878\cdot\pi)$ & $21{\times}46^{3}$\\
    & $\SOthreeconv(12,11,42,0.1768\cdot\pi)$ & $11{\times}84^3$& $\SOthreeconv(21,19,30,0.4089\cdot\pi)$ & $19{\times}60^{3}$\\
    & $\SOthreeStwoconv(11,11,50,0.1474\cdot\pi)$ & $11{\times}100^2$ & $\SOthreeconv(19,16,36,0.3135\cdot\pi)$ & $16{\times}72^{3}$\\
    & & & $\SOthreeconv(16,13,43,0.2613\cdot\pi)$ & $13{\times}86^{3}$\\
    & & & $\SOthreeStwoconv(13,11,50,0.2187\cdot\pi)$ & $11{\times}100^{2}$\\
    \midrule
    Params. & \multicolumn{2}{c}{\num{204073}} & \multicolumn{2}{c}{\num{820184}} \\
    \bottomrule
  \end{tabular}
\end{table*}
The equivariant models consist of the three layers described in Section~\ref{sec:equivariant-model}. We will denote the operation \eqref{eq:4} which takes input features on $S^{2}$ and returns output features on $\mathrm{SO}(3)$ by
\begin{align}
  \StwoSOthreeconv(n_{i},n_{o},b_{i},b_{o},\hat{\beta})\,,\label{eq:7}
\end{align}
where $n_{i}$ is the number of input channels, $n_{o}$ is the number of output channels, $b_{i}$ is the bandwidth of the input feature map, $b_{o}$ is the bandwidth of the output feature map and $\hat{\beta}$ is related to the kernel size as described further down. The operation \eqref{eq:5} which takes and returns feature maps on $\mathrm{SO}(3)$ is similarly denoted by
\begin{align}
  \SOthreeconv(n_{i},n_{o},b_{i},b_{o},\hat{\beta})\label{eq:8}
\end{align}
and the operation~\eqref{eq:6} with input feature maps on $\mathrm{SO}(3)$ and output feature maps on $S^{2}$ is denoted by
\begin{align}
  \SOthreeStwoconv(n_{i},n_{o},b_{i},b_{o},\hat{\beta})\,.\label{eq:9}
\end{align}
After each layer, we apply a ReLU activation function.

We also experimented with adding the equivariant skip connections provided by the original S2CNN implementation to the layers \eqref{eq:8} but found no performance gain, so we did not use them in the experiments.

For the spherical models, the procedure to generate architectures is similar to the non-equivariant case, but now the bandlimit replaces the role of the image dimension in the construction.

First, we select the depth and the bottleneck size in the same way as for the non-equivariant models. Then, we select the bandlimit at the bottleneck between 3 and 20 and compute a linear interpolation between the input bandlimit and the bandlimit at the bottleneck. Since the equivariant layers in the S2CNN architecture directly take the input- and output bandlimits as parameters, we only have to round those interpolated bandlimits to integers.

On top of the bandlimits, the spherical convolutional layers also depend on the grid on which the kernel is sampled. For this grid, we take a Driscoll-Healy grid \cite{DRISCOLL1994202} around the unit element in $\mathrm{SO}(3)$. The three Euler angles $\alpha$, $\beta$ and $\gamma$ of the grid points lie between 0 and $2\pi$ for $\alpha$ and $\gamma$ but $\beta$ takes values between 0 and some upper bound $\hat{\beta}$. This upper bound is responsible for the locality of the kernel and is analogous to the kernel size in standard CNN layers. However, since the bandlimit $L_{i}$ of the feature map determines how fine the grid for the feature map is, the effective kernel size is characterized by the product $\hat{\beta}\cdot L_{i}$. In order to keep this product fixed throughout the network, we set $\hat{\beta}_{i} = L / L_{i} \cdot \hat{\beta}^{\mathrm{ref}}$ where $L$ is the bandlimit of the input and $\hat{\beta}^{\mathrm{ref}}$ is a reference value that we pick at random between $0.02$ and $0.25$. Note that this construction implies that the number of trainable parameters in the spatial dimensions of the kernel is fixed throughout the network.

For the equivariant models, we randomly select the channel numbers by first sampling a maximum channel number at the bottleneck between 11 and 30 and then linearly interpolating between 11 (the channel number at the output) and this maximum.

Again, as in the non-equivariant case, we generate architectures according to this procedure and reject the ones whose parameter count lies outside the desired range. The models obtained in this way are trained which batch size 32 and learning rate \num{e-3} using Adam on a segmentation dataset of 60k unrotated training samples containing one MNIST digit on the sphere each until convergence and then evaluated. In all experiments, we use early stopping on the non-background mIoU metric and a maximum of 200~epochs.

As in the non-equivariant case, we generated, trained and evaluated 20 models in the desired parameter ranges and picked the best performing ones according to the non-background mIoU. These models were then used for the experiments.  The resulting architectures of the two equivariant models are summarized in Table~\ref{tab:equiv_models}.

%%% Local Variables:
%%% mode: latex
%%% TeX-master: "mnist_paper"
%%% End:

%% file: appendix_spherical_expmts.tex
\section{Details on  experiments}
\label{sec:spherical-expmts}
In this appendix, we give more details about the experimental setup for the spherical runs described in Section~\ref{sec:spherical_case}.

\subsection{Data generation}

As discussed in the main text, the input data to our models consist of MNIST digits and FashionMNIST items of clothing which were projected onto the sphere, labeled by their corresponding segmentation masks. Here we give more details on the data generation process.

For an input image with $n$ digits / items of clothing on the sphere, we first sampled $n$ images from MNIST or FashionMNIST and pasted them at a random position on a $60\times 60$ canvas. This canvas is then projected onto the sphere by projecting the spherical grid points onto the plane using a stereograhpic projection and performing a bilinear interpolation in the plane. To obtain a rotated input sample, we rotate the grid points with a random rotation matrix before projecting them.

We generate the segmentation mask for a single digit / item of clothing by considering all pixels with a grayscale value above a certain threshold as belonging to the target class and to the background class otherwise. These segmentation masks are then assembled into the $60\times 60$ canvas and projected onto the sphere. Instead of a bilinear interpolation, we use nearest-neighbor interpolation for the segmentation masks.

\begin{figure}
  \centering
  \includegraphics[width=0.99\columnwidth]{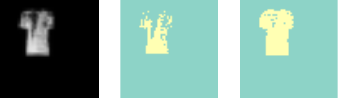}
  \caption{Examples of segmentation masks for FashionMNIST generated from the original (left) with threshold values of 150 (center) and 10 (right).}
  \label{fig:fashion_thresholds}
\end{figure}
For MNIST, a threshold value of 150 yielded good results. For FashionMNIST however, we use a value of 10 to capture finer details of the cloths, as illustrated in Figure~\ref{fig:fashion_thresholds}. Even lower values lead to a blurring along the edges in the segmentation mask.

For validation, we generated datasets in the same way as for training, but sampling \num{10000} data points from the test split of the MNIST dataset. 

\subsection{Performance variation due to data sampling}
\label{sec:perf-vari-data-sampl}
\begin{figure}[t]
  \centering
  \resizebox{0.49\textwidth}{!}{\input{imgs/spherical_data_augmentation_2_digits_reruns.tex}}
  \caption{\textit{Variation in Performance}. Performance for the non-equivariant 218k parameters model trained three times on rotated datasets of 2 MNIST digits on the sphere of various sizes.}
  \label{fig:2_digits_reruns}
\end{figure}
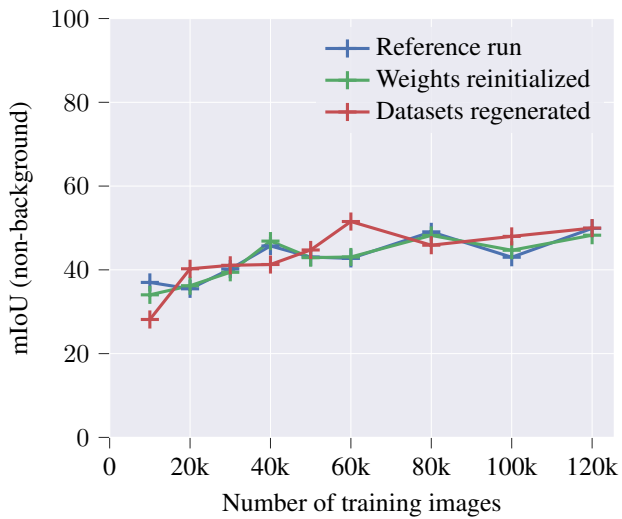

As mentioned in the main text, increases or decreases in performance with varying dataset sizes that occur completely in parallel for all models are due to sampling effects during the data generation process. This point is illustrated by the plot depicted in Figure~\ref{fig:2_digits_reruns}, where we trained the non-equivariant model with 218k parameters three times on rotated datasets of various sizes with 2 MNIST digits on the sphere. One run is for reference, one with randomly reinitialized weights before training and one was trained on a randomly regenerated dataset. We can see clearly that the weight reinitialization only had a minor influence on the model performance, but the peculiar irregularities of the performance increase with growing training dataset are within the variation of regenerating the dataset.

\subsection{Influence of projection point for non-rotated data}
\label{sec:infl-proj-point}
\begin{figure}[t]
  \centering
  \includegraphics[width=0.67\columnwidth]{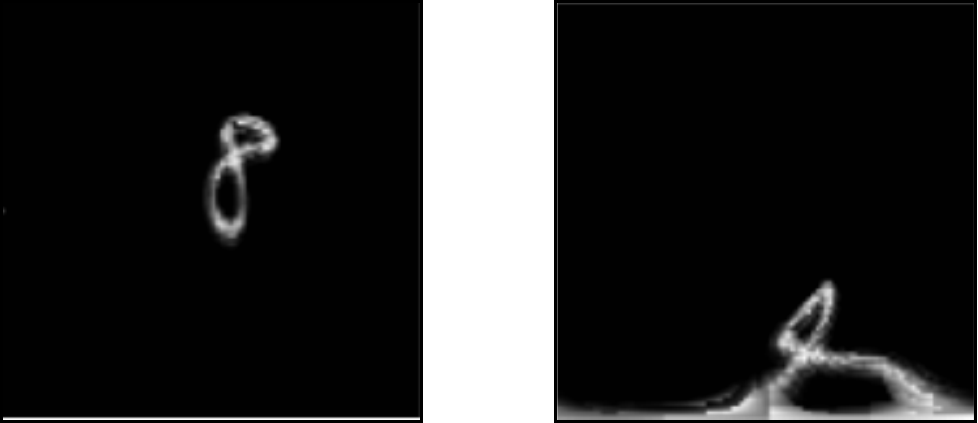}
  \caption{Example of the input data in the Driscoll--Healy grid for a digit projected onto the grid center (left) and onto the pole of the sphere (right).}
  \label{fig:proj_example}
\end{figure}
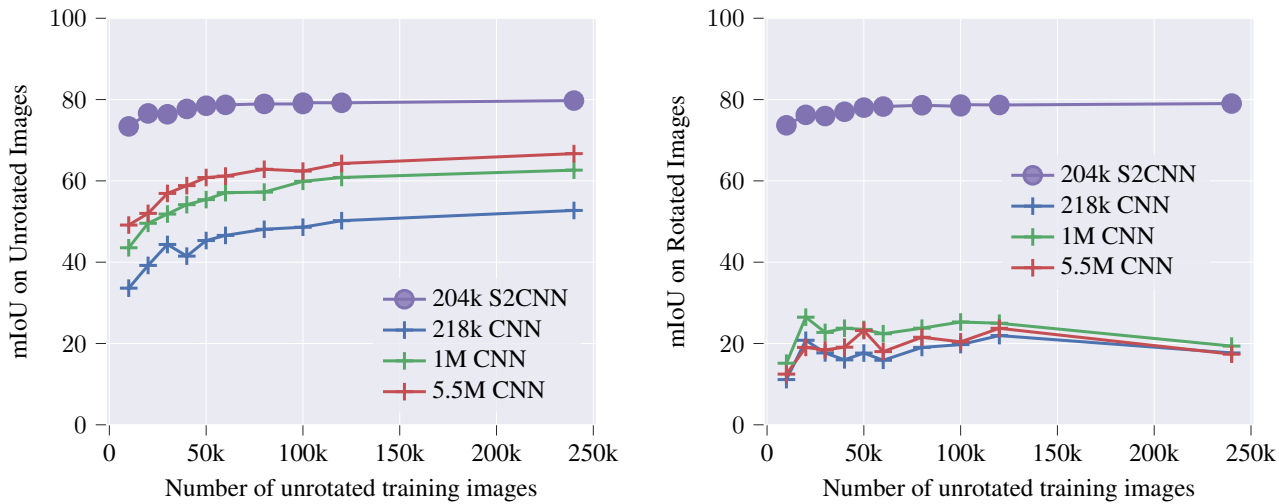
\begin{figure*}
  \centering
  \resizebox{0.49\textwidth}{!}{\input{imgs/spherical_data_augmentation_non_rot_non_rot_eval.tex}}
  \hfill
 \resizebox{0.49\textwidth}{!}{\input{imgs/spherical_data_augmentation_non_rot_rot_eval.tex}}
 \caption{\textit{Polar Training Images}. Performance of equivariant and non-equivariant models for various amounts of data augmentation for models trained on unrotated data, projected on the pole of the sphere. Performance is measured in terms of mIoU for the non-background classes. Left: Evaluated on unrotated test data. Right: evaluated on rotated test data.}
  \label{fig:data_augm_non_rot_pole_proj}
\end{figure*}
The spherical data we use in our experiments is given in the Driscoll-Healy grid which is an equispaced grid in spherical coordinates, i.e.\ in the azimuthal angle $\phi$ and polar angle $\theta$. Therefore, what the input data looks like on this grid depends crucially on the projection point relative to the grid. Figure~\ref{fig:proj_example} illustrates this with a digit projected into the center of the $\phi,\theta$ grid (i.e.\ on the equator, close to the $\phi=\pi$ line) and a digit projected onto the pole.

For the experiments for non-rotated data depicted in Figure~\ref{fig:data_augm_non_rot} of the main text, we projected the input images onto the grid center, to help the non-equivariant networks. For this projection point, slight variations of the digit positions correspond almost to translations in the Driscoll-Healy grid with respect to which the ordinary CNNs are equivariant. For comparison, Figure~\ref{fig:data_augm_non_rot_pole_proj} shows the results of training on data projected onto the pole of the sphere. Note that the performance on unrotated data is reduced considerably as compared to Figure~\ref{fig:data_augm_non_rot}, but slightly improved on rotated data. The reason is that slight positional variations across the pole lead to very non-linear deformations of the data represented in the Driscoll-Healy grid and the translational equivariance of the CNNs cannot help the training process. On the other hand, since the polar projections are most challenging, overall performance on rotated test data is improved (although of course still very poor).

Since the S2CNN models are equivariant with respect to all rotations on the sphere, their performance does not depend on the projection point of the digit, cf.\ Figure~\ref{fig:data_augm_non_rot_pole_proj} vs.\ Figure~\ref{fig:data_augm_non_rot}. Therefore, we trained the S2CNN models for all other experiments on images projected onto the pole.

\subsection{Sample predictions}
\label{sec:sample-predictions}
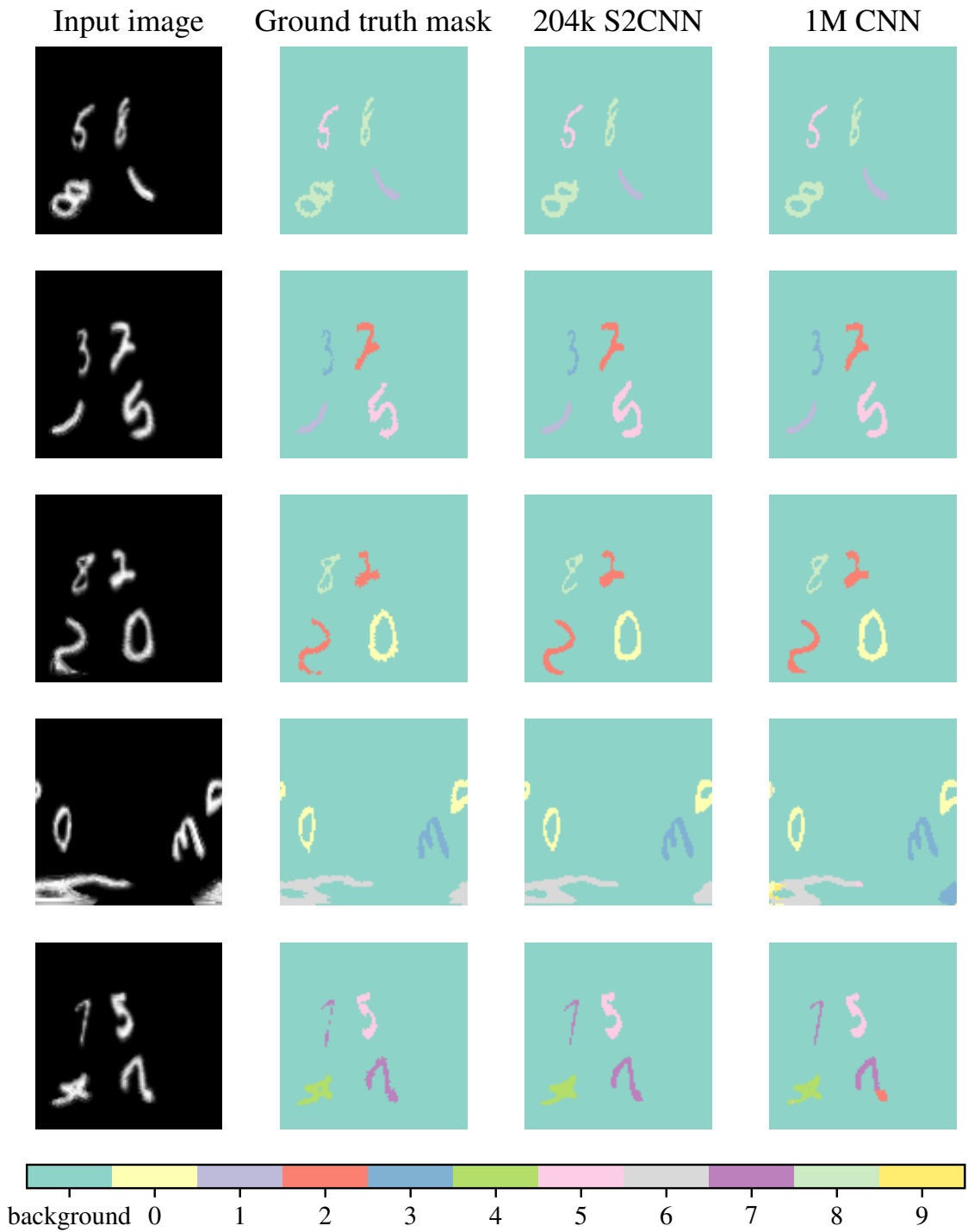
\begin{figure*}
  \centering
  \resizebox{0.9\textwidth}{!}{\input{imgs/sample_preds_part_1.pgf}}
  \caption{Sample predictions on the test dataset for the best equivariant model (240k S2CNN, trained on 240k samples) and non-equivariant model (1M CNN, trained on 600k samples) on the four-digits task depicted in Figure~\ref{fig:sphericalMNIST_4_digits}. The five samples were selected at random from the dataset and we depict here the raw data on the Driscoll--Healy grid. Five more samples are depicted in Figure~\ref{fig:sample_preds_2}.}
  \label{fig:sample_preds_1}
\end{figure*}

In Figures~\ref{fig:sample_preds_1} and \ref{fig:sample_preds_2}, we show ten random sample predictions from the best equivariant and non-equivariant models in the four-digit task.
\begin{figure*}
  \centering
  \resizebox{0.9\textwidth}{!}{\input{imgs/sample_preds_part_2.pgf}}
  \caption{Five more samples of model predictions, cf.~Figure~\ref{fig:sample_preds_1}.}
  \label{fig:sample_preds_2}
\end{figure*}

%%% Local Variables:
%%% mode: latex
%%% TeX-master: "mnist_paper"
%%% End:

%% file: imgs/spherical_data_augmentation_2_digits_reruns.tex
% This file was created with tikzplotlib v0.9.12.
\begin{tikzpicture}

\definecolor{color0}{rgb}{0.917647058823529,0.917647058823529,0.949019607843137}
\definecolor{color1}{rgb}{0.298039215686275,0.447058823529412,0.690196078431373}
\definecolor{color2}{rgb}{0.333333333333333,0.658823529411765,0.407843137254902}
\definecolor{color3}{rgb}{0.768627450980392,0.305882352941176,0.32156862745098}

\begin{axis}[
axis background/.style={fill=color0},
axis line style={white},
legend cell align={left},
legend style={fill opacity=0.8, draw opacity=1, text opacity=1, draw=none, fill=color0},
tick align=outside,
tick pos=left,
x grid style={white},
xlabel={Number of training images},
xmajorgrids,
xmin=0, xmax=125.5,
xtick style={color=white!15!black},
xtick={0,20,40,60,80,100,120},
xticklabels={0,20k,40k,60k,80k,100k,120k},
y grid style={white},
ylabel={mIoU (non-background)},
ymajorgrids,
ymin=0, ymax=100,
ytick style={color=white!15!black}
]
\addplot [very thick, color1, mark=+, mark size=3.5, mark options={solid}]
table {%
10 36.999541410654
20 35.4719139427761
30 40.1979044537547
40 45.7091823155197
50 43.0583567600674
60 42.731509863652
80 49.0717858592495
100 43.0071714185072
120 49.9036082320355
};
\addlegendentry{Reference run}
\addplot [very thick, color2, mark=+, mark size=3.5, mark options={solid}]
table {%
10 34.0363217492872
20 36.2153535219318
30 39.4180808776375
40 46.8659110910392
50 42.9096176663126
60 43.0718245835321
80 48.2215805971468
100 44.6800974756389
120 48.2712558576458
};
\addlegendentry{Weights reinitialized}
\addplot [very thick, color3, mark=+, mark size=3.5, mark options={solid}]
table {%
10 28.1521862084209
20 40.2530210345169
30 41.0847126381996
40 41.2770981830282
50 44.7472489127007
60 51.5356506872585
80 45.8847423425363
100 47.9881792168513
120 49.9737842268687
};
\addlegendentry{Datasets regenerated}
\end{axis}

\end{tikzpicture}

%% file: imgs/spherical_data_augmentation_non_rot_non_rot_eval.tex
% This file was created with tikzplotlib v0.9.12.
\begin{tikzpicture}

\definecolor{color0}{rgb}{0.917647058823529,0.917647058823529,0.949019607843137}
\definecolor{color1}{rgb}{0.505882352941176,0.447058823529412,0.698039215686274}
\definecolor{color2}{rgb}{0.298039215686275,0.447058823529412,0.690196078431373}
\definecolor{color3}{rgb}{0.333333333333333,0.658823529411765,0.407843137254902}
\definecolor{color4}{rgb}{0.768627450980392,0.305882352941176,0.32156862745098}

\begin{axis}[
axis background/.style={fill=color0},
axis line style={white},
legend cell align={left},
legend style={
  fill opacity=0.8,
  draw opacity=1,
  text opacity=1,
  at={(0.97,0.03)},
  anchor=south east,
  draw=none,
  fill=color0
},
tick align=outside,
tick pos=left,
x grid style={white},
xlabel={Number of unrotated training images},
xmajorgrids,
xmin=-1.5, xmax=251.5,
xtick style={color=white!15!black},
xtick={0,50,100,150,200,250},
xticklabels={0,50k,100k,150k,200k,250k},
y grid style={white},
ylabel={mIoU on Unrotated Images},
ymajorgrids,
ymin=0, ymax=100,
ytick style={color=white!15!black}
]
\addplot [very thick, color1, mark=*, mark size=3.5, mark options={solid}]
table {%
10 73.3974528914608
20 76.5968217526796
30 76.3819566065774
40 77.7036051787835
50 78.4796345077198
60 78.7009372486817
80 78.9322216422207
100 78.9037558811369
100 79.2380915782961
120 79.2007620844352
240 79.7527575005371
};
\addlegendentry{204k S2CNN}
\addplot [very thick, color2, mark=+, mark size=3.5, mark options={solid}]
table {%
10 33.6142997155828
20 39.2087944634741
30 44.3611360429465
40 41.4904151639397
50 45.3261435831514
60 46.5851298021677
80 48.0902225372996
100 48.6041536898522
120 50.2100768542436
240 52.7297273356198
};
\addlegendentry{218k CNN}
\addplot [very thick, color3, mark=+, mark size=3.5, mark options={solid}]
table {%
10 43.5503171625281
20 49.5661142884367
30 51.8393112195059
40 54.1516122905952
50 55.3929731408812
60 57.0981042334844
80 57.2326157895088
100 59.8884400736839
120 60.838538909474
240 62.6433851779529
};
\addlegendentry{1M CNN}
\addplot [very thick, color4, mark=+, mark size=3.5, mark options={solid}]
table {%
10 49.1335843902316
20 52.0059911691985
30 56.9114730251209
40 58.8349890721363
50 60.8256648268285
60 61.1761882580404
80 62.8526585104493
100 62.4117528990672
120 64.2820799851893
240 66.7026529005458
};
\addlegendentry{5.5M CNN}
\end{axis}

\end{tikzpicture}

%% file: imgs/spherical_data_augmentation_non_rot_rot_eval.tex
% This file was created with tikzplotlib v0.9.12.
\begin{tikzpicture}

\definecolor{color0}{rgb}{0.917647058823529,0.917647058823529,0.949019607843137}
\definecolor{color1}{rgb}{0.505882352941176,0.447058823529412,0.698039215686274}
\definecolor{color2}{rgb}{0.298039215686275,0.447058823529412,0.690196078431373}
\definecolor{color3}{rgb}{0.333333333333333,0.658823529411765,0.407843137254902}
\definecolor{color4}{rgb}{0.768627450980392,0.305882352941176,0.32156862745098}

\begin{axis}[
axis background/.style={fill=color0},
axis line style={white},
legend cell align={left},
legend style={
  fill opacity=0.8,
  draw opacity=1,
  text opacity=1,
  at={(0.91,0.5)},
  anchor=east,
  draw=none,
  fill=color0
},
tick align=outside,
tick pos=left,
x grid style={white},
xlabel={Number of unrotated training images},
xmajorgrids,
xmin=-1.5, xmax=251.5,
xtick style={color=white!15!black},
xtick={0,50,100,150,200,250},
xticklabels={0,50k,100k,150k,200k,250k},
y grid style={white},
ylabel={mIoU on Rotated Images},
ymajorgrids,
ymin=0, ymax=100,
ytick style={color=white!15!black}
]
\addplot [very thick, color1, mark=*, mark size=3.5, mark options={solid}]
table {%
10 73.6672228087486
20 76.2396656837297
30 75.9440592307601
40 77.0006140071276
50 78.0121633327085
60 78.2821354107209
80 78.6005328484791
100 78.3440466722201
100 78.7591833919464
120 78.6653772927548
240 79.0172031795914
};
\addlegendentry{204k S2CNN}
\addplot [very thick, color2, mark=+, mark size=3.5, mark options={solid}]
table {%
10 11.125275802043
20 20.7920309164621
30 17.6808207296639
40 15.9798926514498
50 17.6487589648848
60 15.8718983090195
80 19.0297573594044
100 19.7329351714989
120 21.9510599115003
240 17.6786745085189
};
\addlegendentry{218k CNN}
\addplot [very thick, color3, mark=+, mark size=3.5, mark options={solid}]
table {%
10 15.1485398445208
20 26.4705773510407
30 22.7274621554245
40 23.7518328222184
50 23.4576474385992
60 22.4008266401536
80 23.766361106333
100 25.2877050644026
120 24.9855421187989
240 19.3646375998322
};
\addlegendentry{1M CNN}
\addplot [very thick, color4, mark=+, mark size=3.5, mark options={solid}]
table {%
10 12.469889385026
20 19.0285200589813
30 18.4204909566414
40 19.0925597467302
50 23.189671089043
60 18.006331735596
80 21.5283216363026
100 20.4088356050101
120 23.7352566348186
240 17.372136374598
};
\addlegendentry{5.5M CNN}
\end{axis}

\end{tikzpicture}

%% file: imgs/sample_preds_part_1.pgf
%% Creator: Matplotlib, PGF backend
%%
%% To include the figure in your LaTeX document, write
%%   \input{<filename>.pgf}
%%
%% Make sure the required packages are loaded in your preamble
%%   \usepackage{pgf}
%%
%% Also ensure that all the required font packages are loaded; for instance,
%% the lmodern package is sometimes necessary when using math font.
%%   \usepackage{lmodern}
%%
%% Figures using additional raster images can only be included by \input if
%% they are in the same directory as the main LaTeX file. For loading figures
%% from other directories you can use the `import` package
%% and then include the figures with
%%   \import{<path to file>}{<filename>.pgf}
%%
%% Matplotlib used the following preamble
%%
\begingroup%
\makeatletter%
\begin{pgfpicture}%
\pgfpathrectangle{\pgfpointorigin}{\pgfqpoint{5.285820in}{6.656060in}}%
\pgfusepath{use as bounding box, clip}%
\begin{pgfscope}%
\pgfsetbuttcap%
\pgfsetmiterjoin%
\definecolor{currentfill}{rgb}{1.000000,1.000000,1.000000}%
\pgfsetfillcolor{currentfill}%
\pgfsetlinewidth{0.000000pt}%
\definecolor{currentstroke}{rgb}{1.000000,1.000000,1.000000}%
\pgfsetstrokecolor{currentstroke}%
\pgfsetdash{}{0pt}%
\pgfpathmoveto{\pgfqpoint{0.000000in}{0.000000in}}%
\pgfpathlineto{\pgfqpoint{5.285820in}{0.000000in}}%
\pgfpathlineto{\pgfqpoint{5.285820in}{6.656060in}}%
\pgfpathlineto{\pgfqpoint{0.000000in}{6.656060in}}%
\pgfpathlineto{\pgfqpoint{0.000000in}{0.000000in}}%
\pgfpathclose%
\pgfusepath{fill}%
\end{pgfscope}%
\begin{pgfscope}%
\pgfpathrectangle{\pgfqpoint{0.271449in}{5.369984in}}{\pgfqpoint{0.987003in}{0.987003in}}%
\pgfusepath{clip}%
\pgfsys@transformshift{0.271449in}{5.369984in}%
\pgftext[left,bottom]{\includegraphics[interpolate=true,width=0.990000in,height=0.990000in]{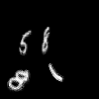}}%
\end{pgfscope}%
\begin{pgfscope}%
\definecolor{textcolor}{rgb}{0.000000,0.000000,0.000000}%
\pgfsetstrokecolor{textcolor}%
\pgfsetfillcolor{textcolor}%
\pgftext[x=0.764950in,y=6.440320in,,base]{\color{textcolor}\rmfamily\fontsize{12.000000}{14.400000}\selectfont Input image}%
\end{pgfscope}%
\begin{pgfscope}%
\pgfpathrectangle{\pgfqpoint{1.565362in}{5.369984in}}{\pgfqpoint{0.987003in}{0.987003in}}%
\pgfusepath{clip}%
\pgfsys@transformcm{0.990000}{0.000000}{0.000000}{-0.990000}{1.565362in}{6.359984in}%
\pgftext[left,bottom]{\includegraphics[interpolate=false,width=1.000000in,height=1.000000in]{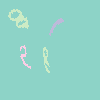}}%
\end{pgfscope}%
\begin{pgfscope}%
\definecolor{textcolor}{rgb}{0.000000,0.000000,0.000000}%
\pgfsetstrokecolor{textcolor}%
\pgfsetfillcolor{textcolor}%
\pgftext[x=2.058863in,y=6.440320in,,base]{\color{textcolor}\rmfamily\fontsize{12.000000}{14.400000}\selectfont Ground truth mask}%
\end{pgfscope}%
\begin{pgfscope}%
\pgfpathrectangle{\pgfqpoint{2.859275in}{5.369984in}}{\pgfqpoint{0.987003in}{0.987003in}}%
\pgfusepath{clip}%
\pgfsys@transformcm{0.990000}{0.000000}{0.000000}{-0.990000}{2.859275in}{6.359984in}%
\pgftext[left,bottom]{\includegraphics[interpolate=false,width=1.000000in,height=1.000000in]{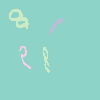}}%
\end{pgfscope}%
\begin{pgfscope}%
\definecolor{textcolor}{rgb}{0.000000,0.000000,0.000000}%
\pgfsetstrokecolor{textcolor}%
\pgfsetfillcolor{textcolor}%
\pgftext[x=3.352776in,y=6.440320in,,base]{\color{textcolor}\rmfamily\fontsize{12.000000}{14.400000}\selectfont 204k S2CNN}%
\end{pgfscope}%
\begin{pgfscope}%
\pgfpathrectangle{\pgfqpoint{4.153188in}{5.369984in}}{\pgfqpoint{0.987003in}{0.987003in}}%
\pgfusepath{clip}%
\pgfsys@transformcm{0.990000}{0.000000}{0.000000}{-0.990000}{4.153188in}{6.359984in}%
\pgftext[left,bottom]{\includegraphics[interpolate=false,width=1.000000in,height=1.000000in]{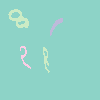}}%
\end{pgfscope}%
\begin{pgfscope}%
\definecolor{textcolor}{rgb}{0.000000,0.000000,0.000000}%
\pgfsetstrokecolor{textcolor}%
\pgfsetfillcolor{textcolor}%
\pgftext[x=4.646689in,y=6.440320in,,base]{\color{textcolor}\rmfamily\fontsize{12.000000}{14.400000}\selectfont 1M CNN}%
\end{pgfscope}%
\begin{pgfscope}%
\pgfpathrectangle{\pgfqpoint{0.271449in}{4.185581in}}{\pgfqpoint{0.987003in}{0.987003in}}%
\pgfusepath{clip}%
\pgfsys@transformshift{0.271449in}{4.185581in}%
\pgftext[left,bottom]{\includegraphics[interpolate=true,width=0.990000in,height=0.990000in]{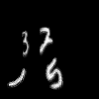}}%
\end{pgfscope}%
\begin{pgfscope}%
\pgfpathrectangle{\pgfqpoint{1.565362in}{4.185581in}}{\pgfqpoint{0.987003in}{0.987003in}}%
\pgfusepath{clip}%
\pgfsys@transformcm{0.990000}{0.000000}{0.000000}{-0.990000}{1.565362in}{5.175581in}%
\pgftext[left,bottom]{\includegraphics[interpolate=false,width=1.000000in,height=1.000000in]{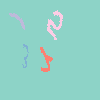}}%
\end{pgfscope}%
\begin{pgfscope}%
\pgfpathrectangle{\pgfqpoint{2.859275in}{4.185581in}}{\pgfqpoint{0.987003in}{0.987003in}}%
\pgfusepath{clip}%
\pgfsys@transformcm{0.990000}{0.000000}{0.000000}{-0.990000}{2.859275in}{5.175581in}%
\pgftext[left,bottom]{\includegraphics[interpolate=false,width=1.000000in,height=1.000000in]{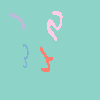}}%
\end{pgfscope}%
\begin{pgfscope}%
\pgfpathrectangle{\pgfqpoint{4.153188in}{4.185581in}}{\pgfqpoint{0.987003in}{0.987003in}}%
\pgfusepath{clip}%
\pgfsys@transformcm{0.990000}{0.000000}{0.000000}{-0.990000}{4.153188in}{5.175581in}%
\pgftext[left,bottom]{\includegraphics[interpolate=false,width=1.000000in,height=1.000000in]{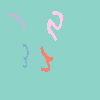}}%
\end{pgfscope}%
\begin{pgfscope}%
\pgfpathrectangle{\pgfqpoint{0.271449in}{3.001178in}}{\pgfqpoint{0.987003in}{0.987003in}}%
\pgfusepath{clip}%
\pgfsys@transformshift{0.271449in}{3.001178in}%
\pgftext[left,bottom]{\includegraphics[interpolate=true,width=0.990000in,height=0.990000in]{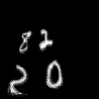}}%
\end{pgfscope}%
\begin{pgfscope}%
\pgfpathrectangle{\pgfqpoint{1.565362in}{3.001178in}}{\pgfqpoint{0.987003in}{0.987003in}}%
\pgfusepath{clip}%
\pgfsys@transformcm{0.990000}{0.000000}{0.000000}{-0.990000}{1.565362in}{3.991178in}%
\pgftext[left,bottom]{\includegraphics[interpolate=false,width=1.000000in,height=1.000000in]{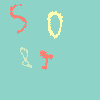}}%
\end{pgfscope}%
\begin{pgfscope}%
\pgfpathrectangle{\pgfqpoint{2.859275in}{3.001178in}}{\pgfqpoint{0.987003in}{0.987003in}}%
\pgfusepath{clip}%
\pgfsys@transformcm{0.990000}{0.000000}{0.000000}{-0.990000}{2.859275in}{3.991178in}%
\pgftext[left,bottom]{\includegraphics[interpolate=false,width=1.000000in,height=1.000000in]{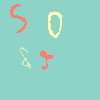}}%
\end{pgfscope}%
\begin{pgfscope}%
\pgfpathrectangle{\pgfqpoint{4.153188in}{3.001178in}}{\pgfqpoint{0.987003in}{0.987003in}}%
\pgfusepath{clip}%
\pgfsys@transformcm{0.990000}{0.000000}{0.000000}{-0.990000}{4.153188in}{3.991178in}%
\pgftext[left,bottom]{\includegraphics[interpolate=false,width=1.000000in,height=1.000000in]{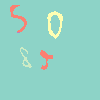}}%
\end{pgfscope}%
\begin{pgfscope}%
\pgfpathrectangle{\pgfqpoint{0.271449in}{1.816774in}}{\pgfqpoint{0.987003in}{0.987003in}}%
\pgfusepath{clip}%
\pgfsys@transformshift{0.271449in}{1.816774in}%
\pgftext[left,bottom]{\includegraphics[interpolate=true,width=0.990000in,height=0.990000in]{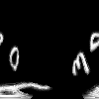}}%
\end{pgfscope}%
\begin{pgfscope}%
\pgfpathrectangle{\pgfqpoint{1.565362in}{1.816774in}}{\pgfqpoint{0.987003in}{0.987003in}}%
\pgfusepath{clip}%
\pgfsys@transformcm{0.990000}{0.000000}{0.000000}{-0.990000}{1.565362in}{2.806774in}%
\pgftext[left,bottom]{\includegraphics[interpolate=false,width=1.000000in,height=1.000000in]{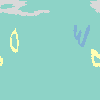}}%
\end{pgfscope}%
\begin{pgfscope}%
\pgfpathrectangle{\pgfqpoint{2.859275in}{1.816774in}}{\pgfqpoint{0.987003in}{0.987003in}}%
\pgfusepath{clip}%
\pgfsys@transformcm{0.990000}{0.000000}{0.000000}{-0.990000}{2.859275in}{2.806774in}%
\pgftext[left,bottom]{\includegraphics[interpolate=false,width=1.000000in,height=1.000000in]{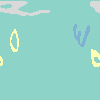}}%
\end{pgfscope}%
\begin{pgfscope}%
\pgfpathrectangle{\pgfqpoint{4.153188in}{1.816774in}}{\pgfqpoint{0.987003in}{0.987003in}}%
\pgfusepath{clip}%
\pgfsys@transformcm{0.990000}{0.000000}{0.000000}{-0.990000}{4.153188in}{2.806774in}%
\pgftext[left,bottom]{\includegraphics[interpolate=false,width=1.000000in,height=1.000000in]{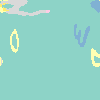}}%
\end{pgfscope}%
\begin{pgfscope}%
\pgfpathrectangle{\pgfqpoint{0.271449in}{0.632371in}}{\pgfqpoint{0.987003in}{0.987003in}}%
\pgfusepath{clip}%
\pgfsys@transformshift{0.271449in}{0.632371in}%
\pgftext[left,bottom]{\includegraphics[interpolate=true,width=0.990000in,height=0.990000in]{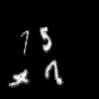}}%
\end{pgfscope}%
\begin{pgfscope}%
\pgfpathrectangle{\pgfqpoint{1.565362in}{0.632371in}}{\pgfqpoint{0.987003in}{0.987003in}}%
\pgfusepath{clip}%
\pgfsys@transformcm{0.990000}{0.000000}{0.000000}{-0.990000}{1.565362in}{1.622371in}%
\pgftext[left,bottom]{\includegraphics[interpolate=false,width=1.000000in,height=1.000000in]{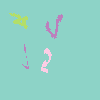}}%
\end{pgfscope}%
\begin{pgfscope}%
\pgfpathrectangle{\pgfqpoint{2.859275in}{0.632371in}}{\pgfqpoint{0.987003in}{0.987003in}}%
\pgfusepath{clip}%
\pgfsys@transformcm{0.990000}{0.000000}{0.000000}{-0.990000}{2.859275in}{1.622371in}%
\pgftext[left,bottom]{\includegraphics[interpolate=false,width=1.000000in,height=1.000000in]{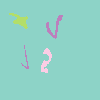}}%
\end{pgfscope}%
\begin{pgfscope}%
\pgfpathrectangle{\pgfqpoint{4.153188in}{0.632371in}}{\pgfqpoint{0.987003in}{0.987003in}}%
\pgfusepath{clip}%
\pgfsys@transformcm{0.990000}{0.000000}{0.000000}{-0.990000}{4.153188in}{1.622371in}%
\pgftext[left,bottom]{\includegraphics[interpolate=false,width=1.000000in,height=1.000000in]{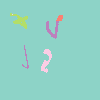}}%
\end{pgfscope}%
\begin{pgfscope}%
\pgfsetbuttcap%
\pgfsetmiterjoin%
\definecolor{currentfill}{rgb}{1.000000,1.000000,1.000000}%
\pgfsetfillcolor{currentfill}%
\pgfsetlinewidth{0.000000pt}%
\definecolor{currentstroke}{rgb}{0.000000,0.000000,0.000000}%
\pgfsetstrokecolor{currentstroke}%
\pgfsetstrokeopacity{0.000000}%
\pgfsetdash{}{0pt}%
\pgfpathmoveto{\pgfqpoint{0.225820in}{0.320679in}}%
\pgfpathlineto{\pgfqpoint{5.185820in}{0.320679in}}%
\pgfpathlineto{\pgfqpoint{5.185820in}{0.444679in}}%
\pgfpathlineto{\pgfqpoint{0.225820in}{0.444679in}}%
\pgfpathlineto{\pgfqpoint{0.225820in}{0.320679in}}%
\pgfpathclose%
\pgfusepath{fill}%
\end{pgfscope}%
\begin{pgfscope}%
\pgfpathrectangle{\pgfqpoint{0.225820in}{0.320679in}}{\pgfqpoint{4.960000in}{0.124000in}}%
\pgfusepath{clip}%
\pgfsetbuttcap%
\pgfsetmiterjoin%
\definecolor{currentfill}{rgb}{1.000000,1.000000,1.000000}%
\pgfsetfillcolor{currentfill}%
\pgfsetlinewidth{0.010037pt}%
\definecolor{currentstroke}{rgb}{1.000000,1.000000,1.000000}%
\pgfsetstrokecolor{currentstroke}%
\pgfsetdash{}{0pt}%
\pgfusepath{stroke,fill}%
\end{pgfscope}%
\begin{pgfscope}%
\pgfpathrectangle{\pgfqpoint{0.225820in}{0.320679in}}{\pgfqpoint{4.960000in}{0.124000in}}%
\pgfusepath{clip}%
\pgfsetbuttcap%
\pgfsetroundjoin%
\definecolor{currentfill}{rgb}{0.552941,0.827451,0.780392}%
\pgfsetfillcolor{currentfill}%
\pgfsetlinewidth{0.000000pt}%
\definecolor{currentstroke}{rgb}{0.000000,0.000000,0.000000}%
\pgfsetstrokecolor{currentstroke}%
\pgfsetdash{}{0pt}%
\pgfpathmoveto{\pgfqpoint{0.225820in}{0.320679in}}%
\pgfpathlineto{\pgfqpoint{0.225820in}{0.444679in}}%
\pgfpathlineto{\pgfqpoint{0.676729in}{0.444679in}}%
\pgfpathlineto{\pgfqpoint{0.676729in}{0.320679in}}%
\pgfpathlineto{\pgfqpoint{0.225820in}{0.320679in}}%
\pgfusepath{fill}%
\end{pgfscope}%
\begin{pgfscope}%
\pgfpathrectangle{\pgfqpoint{0.225820in}{0.320679in}}{\pgfqpoint{4.960000in}{0.124000in}}%
\pgfusepath{clip}%
\pgfsetbuttcap%
\pgfsetroundjoin%
\definecolor{currentfill}{rgb}{1.000000,1.000000,0.701961}%
\pgfsetfillcolor{currentfill}%
\pgfsetlinewidth{0.000000pt}%
\definecolor{currentstroke}{rgb}{0.000000,0.000000,0.000000}%
\pgfsetstrokecolor{currentstroke}%
\pgfsetdash{}{0pt}%
\pgfpathmoveto{\pgfqpoint{0.676729in}{0.320679in}}%
\pgfpathlineto{\pgfqpoint{0.676729in}{0.444679in}}%
\pgfpathlineto{\pgfqpoint{1.127638in}{0.444679in}}%
\pgfpathlineto{\pgfqpoint{1.127638in}{0.320679in}}%
\pgfpathlineto{\pgfqpoint{0.676729in}{0.320679in}}%
\pgfusepath{fill}%
\end{pgfscope}%
\begin{pgfscope}%
\pgfpathrectangle{\pgfqpoint{0.225820in}{0.320679in}}{\pgfqpoint{4.960000in}{0.124000in}}%
\pgfusepath{clip}%
\pgfsetbuttcap%
\pgfsetroundjoin%
\definecolor{currentfill}{rgb}{0.745098,0.729412,0.854902}%
\pgfsetfillcolor{currentfill}%
\pgfsetlinewidth{0.000000pt}%
\definecolor{currentstroke}{rgb}{0.000000,0.000000,0.000000}%
\pgfsetstrokecolor{currentstroke}%
\pgfsetdash{}{0pt}%
\pgfpathmoveto{\pgfqpoint{1.127638in}{0.320679in}}%
\pgfpathlineto{\pgfqpoint{1.127638in}{0.444679in}}%
\pgfpathlineto{\pgfqpoint{1.578547in}{0.444679in}}%
\pgfpathlineto{\pgfqpoint{1.578547in}{0.320679in}}%
\pgfpathlineto{\pgfqpoint{1.127638in}{0.320679in}}%
\pgfusepath{fill}%
\end{pgfscope}%
\begin{pgfscope}%
\pgfpathrectangle{\pgfqpoint{0.225820in}{0.320679in}}{\pgfqpoint{4.960000in}{0.124000in}}%
\pgfusepath{clip}%
\pgfsetbuttcap%
\pgfsetroundjoin%
\definecolor{currentfill}{rgb}{0.984314,0.501961,0.447059}%
\pgfsetfillcolor{currentfill}%
\pgfsetlinewidth{0.000000pt}%
\definecolor{currentstroke}{rgb}{0.000000,0.000000,0.000000}%
\pgfsetstrokecolor{currentstroke}%
\pgfsetdash{}{0pt}%
\pgfpathmoveto{\pgfqpoint{1.578547in}{0.320679in}}%
\pgfpathlineto{\pgfqpoint{1.578547in}{0.444679in}}%
\pgfpathlineto{\pgfqpoint{2.029456in}{0.444679in}}%
\pgfpathlineto{\pgfqpoint{2.029456in}{0.320679in}}%
\pgfpathlineto{\pgfqpoint{1.578547in}{0.320679in}}%
\pgfusepath{fill}%
\end{pgfscope}%
\begin{pgfscope}%
\pgfpathrectangle{\pgfqpoint{0.225820in}{0.320679in}}{\pgfqpoint{4.960000in}{0.124000in}}%
\pgfusepath{clip}%
\pgfsetbuttcap%
\pgfsetroundjoin%
\definecolor{currentfill}{rgb}{0.501961,0.694118,0.827451}%
\pgfsetfillcolor{currentfill}%
\pgfsetlinewidth{0.000000pt}%
\definecolor{currentstroke}{rgb}{0.000000,0.000000,0.000000}%
\pgfsetstrokecolor{currentstroke}%
\pgfsetdash{}{0pt}%
\pgfpathmoveto{\pgfqpoint{2.029456in}{0.320679in}}%
\pgfpathlineto{\pgfqpoint{2.029456in}{0.444679in}}%
\pgfpathlineto{\pgfqpoint{2.480365in}{0.444679in}}%
\pgfpathlineto{\pgfqpoint{2.480365in}{0.320679in}}%
\pgfpathlineto{\pgfqpoint{2.029456in}{0.320679in}}%
\pgfusepath{fill}%
\end{pgfscope}%
\begin{pgfscope}%
\pgfpathrectangle{\pgfqpoint{0.225820in}{0.320679in}}{\pgfqpoint{4.960000in}{0.124000in}}%
\pgfusepath{clip}%
\pgfsetbuttcap%
\pgfsetroundjoin%
\definecolor{currentfill}{rgb}{0.701961,0.870588,0.411765}%
\pgfsetfillcolor{currentfill}%
\pgfsetlinewidth{0.000000pt}%
\definecolor{currentstroke}{rgb}{0.000000,0.000000,0.000000}%
\pgfsetstrokecolor{currentstroke}%
\pgfsetdash{}{0pt}%
\pgfpathmoveto{\pgfqpoint{2.480365in}{0.320679in}}%
\pgfpathlineto{\pgfqpoint{2.480365in}{0.444679in}}%
\pgfpathlineto{\pgfqpoint{2.931274in}{0.444679in}}%
\pgfpathlineto{\pgfqpoint{2.931274in}{0.320679in}}%
\pgfpathlineto{\pgfqpoint{2.480365in}{0.320679in}}%
\pgfusepath{fill}%
\end{pgfscope}%
\begin{pgfscope}%
\pgfpathrectangle{\pgfqpoint{0.225820in}{0.320679in}}{\pgfqpoint{4.960000in}{0.124000in}}%
\pgfusepath{clip}%
\pgfsetbuttcap%
\pgfsetroundjoin%
\definecolor{currentfill}{rgb}{0.988235,0.803922,0.898039}%
\pgfsetfillcolor{currentfill}%
\pgfsetlinewidth{0.000000pt}%
\definecolor{currentstroke}{rgb}{0.000000,0.000000,0.000000}%
\pgfsetstrokecolor{currentstroke}%
\pgfsetdash{}{0pt}%
\pgfpathmoveto{\pgfqpoint{2.931274in}{0.320679in}}%
\pgfpathlineto{\pgfqpoint{2.931274in}{0.444679in}}%
\pgfpathlineto{\pgfqpoint{3.382183in}{0.444679in}}%
\pgfpathlineto{\pgfqpoint{3.382183in}{0.320679in}}%
\pgfpathlineto{\pgfqpoint{2.931274in}{0.320679in}}%
\pgfusepath{fill}%
\end{pgfscope}%
\begin{pgfscope}%
\pgfpathrectangle{\pgfqpoint{0.225820in}{0.320679in}}{\pgfqpoint{4.960000in}{0.124000in}}%
\pgfusepath{clip}%
\pgfsetbuttcap%
\pgfsetroundjoin%
\definecolor{currentfill}{rgb}{0.850980,0.850980,0.850980}%
\pgfsetfillcolor{currentfill}%
\pgfsetlinewidth{0.000000pt}%
\definecolor{currentstroke}{rgb}{0.000000,0.000000,0.000000}%
\pgfsetstrokecolor{currentstroke}%
\pgfsetdash{}{0pt}%
\pgfpathmoveto{\pgfqpoint{3.382183in}{0.320679in}}%
\pgfpathlineto{\pgfqpoint{3.382183in}{0.444679in}}%
\pgfpathlineto{\pgfqpoint{3.833092in}{0.444679in}}%
\pgfpathlineto{\pgfqpoint{3.833092in}{0.320679in}}%
\pgfpathlineto{\pgfqpoint{3.382183in}{0.320679in}}%
\pgfusepath{fill}%
\end{pgfscope}%
\begin{pgfscope}%
\pgfpathrectangle{\pgfqpoint{0.225820in}{0.320679in}}{\pgfqpoint{4.960000in}{0.124000in}}%
\pgfusepath{clip}%
\pgfsetbuttcap%
\pgfsetroundjoin%
\definecolor{currentfill}{rgb}{0.737255,0.501961,0.741176}%
\pgfsetfillcolor{currentfill}%
\pgfsetlinewidth{0.000000pt}%
\definecolor{currentstroke}{rgb}{0.000000,0.000000,0.000000}%
\pgfsetstrokecolor{currentstroke}%
\pgfsetdash{}{0pt}%
\pgfpathmoveto{\pgfqpoint{3.833092in}{0.320679in}}%
\pgfpathlineto{\pgfqpoint{3.833092in}{0.444679in}}%
\pgfpathlineto{\pgfqpoint{4.284001in}{0.444679in}}%
\pgfpathlineto{\pgfqpoint{4.284001in}{0.320679in}}%
\pgfpathlineto{\pgfqpoint{3.833092in}{0.320679in}}%
\pgfusepath{fill}%
\end{pgfscope}%
\begin{pgfscope}%
\pgfpathrectangle{\pgfqpoint{0.225820in}{0.320679in}}{\pgfqpoint{4.960000in}{0.124000in}}%
\pgfusepath{clip}%
\pgfsetbuttcap%
\pgfsetroundjoin%
\definecolor{currentfill}{rgb}{0.800000,0.921569,0.772549}%
\pgfsetfillcolor{currentfill}%
\pgfsetlinewidth{0.000000pt}%
\definecolor{currentstroke}{rgb}{0.000000,0.000000,0.000000}%
\pgfsetstrokecolor{currentstroke}%
\pgfsetdash{}{0pt}%
\pgfpathmoveto{\pgfqpoint{4.284001in}{0.320679in}}%
\pgfpathlineto{\pgfqpoint{4.284001in}{0.444679in}}%
\pgfpathlineto{\pgfqpoint{4.734910in}{0.444679in}}%
\pgfpathlineto{\pgfqpoint{4.734910in}{0.320679in}}%
\pgfpathlineto{\pgfqpoint{4.284001in}{0.320679in}}%
\pgfusepath{fill}%
\end{pgfscope}%
\begin{pgfscope}%
\pgfpathrectangle{\pgfqpoint{0.225820in}{0.320679in}}{\pgfqpoint{4.960000in}{0.124000in}}%
\pgfusepath{clip}%
\pgfsetbuttcap%
\pgfsetroundjoin%
\definecolor{currentfill}{rgb}{1.000000,0.929412,0.435294}%
\pgfsetfillcolor{currentfill}%
\pgfsetlinewidth{0.000000pt}%
\definecolor{currentstroke}{rgb}{0.000000,0.000000,0.000000}%
\pgfsetstrokecolor{currentstroke}%
\pgfsetdash{}{0pt}%
\pgfpathmoveto{\pgfqpoint{4.734910in}{0.320679in}}%
\pgfpathlineto{\pgfqpoint{4.734910in}{0.444679in}}%
\pgfpathlineto{\pgfqpoint{5.185820in}{0.444679in}}%
\pgfpathlineto{\pgfqpoint{5.185820in}{0.320679in}}%
\pgfpathlineto{\pgfqpoint{4.734910in}{0.320679in}}%
\pgfusepath{fill}%
\end{pgfscope}%
\begin{pgfscope}%
\pgfsetbuttcap%
\pgfsetroundjoin%
\definecolor{currentfill}{rgb}{0.000000,0.000000,0.000000}%
\pgfsetfillcolor{currentfill}%
\pgfsetlinewidth{0.803000pt}%
\definecolor{currentstroke}{rgb}{0.000000,0.000000,0.000000}%
\pgfsetstrokecolor{currentstroke}%
\pgfsetdash{}{0pt}%
\pgfsys@defobject{currentmarker}{\pgfqpoint{0.000000in}{-0.048611in}}{\pgfqpoint{0.000000in}{0.000000in}}{%
\pgfpathmoveto{\pgfqpoint{0.000000in}{0.000000in}}%
\pgfpathlineto{\pgfqpoint{0.000000in}{-0.048611in}}%
\pgfusepath{stroke,fill}%
}%
\begin{pgfscope}%
\pgfsys@transformshift{0.451274in}{0.320679in}%
\pgfsys@useobject{currentmarker}{}%
\end{pgfscope}%
\end{pgfscope}%
\begin{pgfscope}%
\definecolor{textcolor}{rgb}{0.000000,0.000000,0.000000}%
\pgfsetstrokecolor{textcolor}%
\pgfsetfillcolor{textcolor}%
\pgftext[x=0.451274in,y=0.223457in,,top]{\color{textcolor}\rmfamily\fontsize{10.000000}{12.000000}\selectfont background}%
\end{pgfscope}%
\begin{pgfscope}%
\pgfsetbuttcap%
\pgfsetroundjoin%
\definecolor{currentfill}{rgb}{0.000000,0.000000,0.000000}%
\pgfsetfillcolor{currentfill}%
\pgfsetlinewidth{0.803000pt}%
\definecolor{currentstroke}{rgb}{0.000000,0.000000,0.000000}%
\pgfsetstrokecolor{currentstroke}%
\pgfsetdash{}{0pt}%
\pgfsys@defobject{currentmarker}{\pgfqpoint{0.000000in}{-0.048611in}}{\pgfqpoint{0.000000in}{0.000000in}}{%
\pgfpathmoveto{\pgfqpoint{0.000000in}{0.000000in}}%
\pgfpathlineto{\pgfqpoint{0.000000in}{-0.048611in}}%
\pgfusepath{stroke,fill}%
}%
\begin{pgfscope}%
\pgfsys@transformshift{0.902183in}{0.320679in}%
\pgfsys@useobject{currentmarker}{}%
\end{pgfscope}%
\end{pgfscope}%
\begin{pgfscope}%
\definecolor{textcolor}{rgb}{0.000000,0.000000,0.000000}%
\pgfsetstrokecolor{textcolor}%
\pgfsetfillcolor{textcolor}%
\pgftext[x=0.902183in,y=0.223457in,,top]{\color{textcolor}\rmfamily\fontsize{10.000000}{12.000000}\selectfont 0}%
\end{pgfscope}%
\begin{pgfscope}%
\pgfsetbuttcap%
\pgfsetroundjoin%
\definecolor{currentfill}{rgb}{0.000000,0.000000,0.000000}%
\pgfsetfillcolor{currentfill}%
\pgfsetlinewidth{0.803000pt}%
\definecolor{currentstroke}{rgb}{0.000000,0.000000,0.000000}%
\pgfsetstrokecolor{currentstroke}%
\pgfsetdash{}{0pt}%
\pgfsys@defobject{currentmarker}{\pgfqpoint{0.000000in}{-0.048611in}}{\pgfqpoint{0.000000in}{0.000000in}}{%
\pgfpathmoveto{\pgfqpoint{0.000000in}{0.000000in}}%
\pgfpathlineto{\pgfqpoint{0.000000in}{-0.048611in}}%
\pgfusepath{stroke,fill}%
}%
\begin{pgfscope}%
\pgfsys@transformshift{1.353092in}{0.320679in}%
\pgfsys@useobject{currentmarker}{}%
\end{pgfscope}%
\end{pgfscope}%
\begin{pgfscope}%
\definecolor{textcolor}{rgb}{0.000000,0.000000,0.000000}%
\pgfsetstrokecolor{textcolor}%
\pgfsetfillcolor{textcolor}%
\pgftext[x=1.353092in,y=0.223457in,,top]{\color{textcolor}\rmfamily\fontsize{10.000000}{12.000000}\selectfont 1}%
\end{pgfscope}%
\begin{pgfscope}%
\pgfsetbuttcap%
\pgfsetroundjoin%
\definecolor{currentfill}{rgb}{0.000000,0.000000,0.000000}%
\pgfsetfillcolor{currentfill}%
\pgfsetlinewidth{0.803000pt}%
\definecolor{currentstroke}{rgb}{0.000000,0.000000,0.000000}%
\pgfsetstrokecolor{currentstroke}%
\pgfsetdash{}{0pt}%
\pgfsys@defobject{currentmarker}{\pgfqpoint{0.000000in}{-0.048611in}}{\pgfqpoint{0.000000in}{0.000000in}}{%
\pgfpathmoveto{\pgfqpoint{0.000000in}{0.000000in}}%
\pgfpathlineto{\pgfqpoint{0.000000in}{-0.048611in}}%
\pgfusepath{stroke,fill}%
}%
\begin{pgfscope}%
\pgfsys@transformshift{1.804001in}{0.320679in}%
\pgfsys@useobject{currentmarker}{}%
\end{pgfscope}%
\end{pgfscope}%
\begin{pgfscope}%
\definecolor{textcolor}{rgb}{0.000000,0.000000,0.000000}%
\pgfsetstrokecolor{textcolor}%
\pgfsetfillcolor{textcolor}%
\pgftext[x=1.804001in,y=0.223457in,,top]{\color{textcolor}\rmfamily\fontsize{10.000000}{12.000000}\selectfont 2}%
\end{pgfscope}%
\begin{pgfscope}%
\pgfsetbuttcap%
\pgfsetroundjoin%
\definecolor{currentfill}{rgb}{0.000000,0.000000,0.000000}%
\pgfsetfillcolor{currentfill}%
\pgfsetlinewidth{0.803000pt}%
\definecolor{currentstroke}{rgb}{0.000000,0.000000,0.000000}%
\pgfsetstrokecolor{currentstroke}%
\pgfsetdash{}{0pt}%
\pgfsys@defobject{currentmarker}{\pgfqpoint{0.000000in}{-0.048611in}}{\pgfqpoint{0.000000in}{0.000000in}}{%
\pgfpathmoveto{\pgfqpoint{0.000000in}{0.000000in}}%
\pgfpathlineto{\pgfqpoint{0.000000in}{-0.048611in}}%
\pgfusepath{stroke,fill}%
}%
\begin{pgfscope}%
\pgfsys@transformshift{2.254910in}{0.320679in}%
\pgfsys@useobject{currentmarker}{}%
\end{pgfscope}%
\end{pgfscope}%
\begin{pgfscope}%
\definecolor{textcolor}{rgb}{0.000000,0.000000,0.000000}%
\pgfsetstrokecolor{textcolor}%
\pgfsetfillcolor{textcolor}%
\pgftext[x=2.254910in,y=0.223457in,,top]{\color{textcolor}\rmfamily\fontsize{10.000000}{12.000000}\selectfont 3}%
\end{pgfscope}%
\begin{pgfscope}%
\pgfsetbuttcap%
\pgfsetroundjoin%
\definecolor{currentfill}{rgb}{0.000000,0.000000,0.000000}%
\pgfsetfillcolor{currentfill}%
\pgfsetlinewidth{0.803000pt}%
\definecolor{currentstroke}{rgb}{0.000000,0.000000,0.000000}%
\pgfsetstrokecolor{currentstroke}%
\pgfsetdash{}{0pt}%
\pgfsys@defobject{currentmarker}{\pgfqpoint{0.000000in}{-0.048611in}}{\pgfqpoint{0.000000in}{0.000000in}}{%
\pgfpathmoveto{\pgfqpoint{0.000000in}{0.000000in}}%
\pgfpathlineto{\pgfqpoint{0.000000in}{-0.048611in}}%
\pgfusepath{stroke,fill}%
}%
\begin{pgfscope}%
\pgfsys@transformshift{2.705820in}{0.320679in}%
\pgfsys@useobject{currentmarker}{}%
\end{pgfscope}%
\end{pgfscope}%
\begin{pgfscope}%
\definecolor{textcolor}{rgb}{0.000000,0.000000,0.000000}%
\pgfsetstrokecolor{textcolor}%
\pgfsetfillcolor{textcolor}%
\pgftext[x=2.705820in,y=0.223457in,,top]{\color{textcolor}\rmfamily\fontsize{10.000000}{12.000000}\selectfont 4}%
\end{pgfscope}%
\begin{pgfscope}%
\pgfsetbuttcap%
\pgfsetroundjoin%
\definecolor{currentfill}{rgb}{0.000000,0.000000,0.000000}%
\pgfsetfillcolor{currentfill}%
\pgfsetlinewidth{0.803000pt}%
\definecolor{currentstroke}{rgb}{0.000000,0.000000,0.000000}%
\pgfsetstrokecolor{currentstroke}%
\pgfsetdash{}{0pt}%
\pgfsys@defobject{currentmarker}{\pgfqpoint{0.000000in}{-0.048611in}}{\pgfqpoint{0.000000in}{0.000000in}}{%
\pgfpathmoveto{\pgfqpoint{0.000000in}{0.000000in}}%
\pgfpathlineto{\pgfqpoint{0.000000in}{-0.048611in}}%
\pgfusepath{stroke,fill}%
}%
\begin{pgfscope}%
\pgfsys@transformshift{3.156729in}{0.320679in}%
\pgfsys@useobject{currentmarker}{}%
\end{pgfscope}%
\end{pgfscope}%
\begin{pgfscope}%
\definecolor{textcolor}{rgb}{0.000000,0.000000,0.000000}%
\pgfsetstrokecolor{textcolor}%
\pgfsetfillcolor{textcolor}%
\pgftext[x=3.156729in,y=0.223457in,,top]{\color{textcolor}\rmfamily\fontsize{10.000000}{12.000000}\selectfont 5}%
\end{pgfscope}%
\begin{pgfscope}%
\pgfsetbuttcap%
\pgfsetroundjoin%
\definecolor{currentfill}{rgb}{0.000000,0.000000,0.000000}%
\pgfsetfillcolor{currentfill}%
\pgfsetlinewidth{0.803000pt}%
\definecolor{currentstroke}{rgb}{0.000000,0.000000,0.000000}%
\pgfsetstrokecolor{currentstroke}%
\pgfsetdash{}{0pt}%
\pgfsys@defobject{currentmarker}{\pgfqpoint{0.000000in}{-0.048611in}}{\pgfqpoint{0.000000in}{0.000000in}}{%
\pgfpathmoveto{\pgfqpoint{0.000000in}{0.000000in}}%
\pgfpathlineto{\pgfqpoint{0.000000in}{-0.048611in}}%
\pgfusepath{stroke,fill}%
}%
\begin{pgfscope}%
\pgfsys@transformshift{3.607638in}{0.320679in}%
\pgfsys@useobject{currentmarker}{}%
\end{pgfscope}%
\end{pgfscope}%
\begin{pgfscope}%
\definecolor{textcolor}{rgb}{0.000000,0.000000,0.000000}%
\pgfsetstrokecolor{textcolor}%
\pgfsetfillcolor{textcolor}%
\pgftext[x=3.607638in,y=0.223457in,,top]{\color{textcolor}\rmfamily\fontsize{10.000000}{12.000000}\selectfont 6}%
\end{pgfscope}%
\begin{pgfscope}%
\pgfsetbuttcap%
\pgfsetroundjoin%
\definecolor{currentfill}{rgb}{0.000000,0.000000,0.000000}%
\pgfsetfillcolor{currentfill}%
\pgfsetlinewidth{0.803000pt}%
\definecolor{currentstroke}{rgb}{0.000000,0.000000,0.000000}%
\pgfsetstrokecolor{currentstroke}%
\pgfsetdash{}{0pt}%
\pgfsys@defobject{currentmarker}{\pgfqpoint{0.000000in}{-0.048611in}}{\pgfqpoint{0.000000in}{0.000000in}}{%
\pgfpathmoveto{\pgfqpoint{0.000000in}{0.000000in}}%
\pgfpathlineto{\pgfqpoint{0.000000in}{-0.048611in}}%
\pgfusepath{stroke,fill}%
}%
\begin{pgfscope}%
\pgfsys@transformshift{4.058547in}{0.320679in}%
\pgfsys@useobject{currentmarker}{}%
\end{pgfscope}%
\end{pgfscope}%
\begin{pgfscope}%
\definecolor{textcolor}{rgb}{0.000000,0.000000,0.000000}%
\pgfsetstrokecolor{textcolor}%
\pgfsetfillcolor{textcolor}%
\pgftext[x=4.058547in,y=0.223457in,,top]{\color{textcolor}\rmfamily\fontsize{10.000000}{12.000000}\selectfont 7}%
\end{pgfscope}%
\begin{pgfscope}%
\pgfsetbuttcap%
\pgfsetroundjoin%
\definecolor{currentfill}{rgb}{0.000000,0.000000,0.000000}%
\pgfsetfillcolor{currentfill}%
\pgfsetlinewidth{0.803000pt}%
\definecolor{currentstroke}{rgb}{0.000000,0.000000,0.000000}%
\pgfsetstrokecolor{currentstroke}%
\pgfsetdash{}{0pt}%
\pgfsys@defobject{currentmarker}{\pgfqpoint{0.000000in}{-0.048611in}}{\pgfqpoint{0.000000in}{0.000000in}}{%
\pgfpathmoveto{\pgfqpoint{0.000000in}{0.000000in}}%
\pgfpathlineto{\pgfqpoint{0.000000in}{-0.048611in}}%
\pgfusepath{stroke,fill}%
}%
\begin{pgfscope}%
\pgfsys@transformshift{4.509456in}{0.320679in}%
\pgfsys@useobject{currentmarker}{}%
\end{pgfscope}%
\end{pgfscope}%
\begin{pgfscope}%
\definecolor{textcolor}{rgb}{0.000000,0.000000,0.000000}%
\pgfsetstrokecolor{textcolor}%
\pgfsetfillcolor{textcolor}%
\pgftext[x=4.509456in,y=0.223457in,,top]{\color{textcolor}\rmfamily\fontsize{10.000000}{12.000000}\selectfont 8}%
\end{pgfscope}%
\begin{pgfscope}%
\pgfsetbuttcap%
\pgfsetroundjoin%
\definecolor{currentfill}{rgb}{0.000000,0.000000,0.000000}%
\pgfsetfillcolor{currentfill}%
\pgfsetlinewidth{0.803000pt}%
\definecolor{currentstroke}{rgb}{0.000000,0.000000,0.000000}%
\pgfsetstrokecolor{currentstroke}%
\pgfsetdash{}{0pt}%
\pgfsys@defobject{currentmarker}{\pgfqpoint{0.000000in}{-0.048611in}}{\pgfqpoint{0.000000in}{0.000000in}}{%
\pgfpathmoveto{\pgfqpoint{0.000000in}{0.000000in}}%
\pgfpathlineto{\pgfqpoint{0.000000in}{-0.048611in}}%
\pgfusepath{stroke,fill}%
}%
\begin{pgfscope}%
\pgfsys@transformshift{4.960365in}{0.320679in}%
\pgfsys@useobject{currentmarker}{}%
\end{pgfscope}%
\end{pgfscope}%
\begin{pgfscope}%
\definecolor{textcolor}{rgb}{0.000000,0.000000,0.000000}%
\pgfsetstrokecolor{textcolor}%
\pgfsetfillcolor{textcolor}%
\pgftext[x=4.960365in,y=0.223457in,,top]{\color{textcolor}\rmfamily\fontsize{10.000000}{12.000000}\selectfont 9}%
\end{pgfscope}%
\begin{pgfscope}%
\pgfsetrectcap%
\pgfsetmiterjoin%
\pgfsetlinewidth{0.803000pt}%
\definecolor{currentstroke}{rgb}{0.000000,0.000000,0.000000}%
\pgfsetstrokecolor{currentstroke}%
\pgfsetdash{}{0pt}%
\pgfpathmoveto{\pgfqpoint{0.225820in}{0.320679in}}%
\pgfpathlineto{\pgfqpoint{0.225820in}{0.382679in}}%
\pgfpathlineto{\pgfqpoint{0.225820in}{0.444679in}}%
\pgfpathlineto{\pgfqpoint{5.185820in}{0.444679in}}%
\pgfpathlineto{\pgfqpoint{5.185820in}{0.382679in}}%
\pgfpathlineto{\pgfqpoint{5.185820in}{0.320679in}}%
\pgfpathlineto{\pgfqpoint{0.225820in}{0.320679in}}%
\pgfpathclose%
\pgfusepath{stroke}%
\end{pgfscope}%
\end{pgfpicture}%
\makeatother%
\endgroup%

%% file: imgs/sample_preds_part_2.pgf
%% Creator: Matplotlib, PGF backend
%%
%% To include the figure in your LaTeX document, write
%%   \input{<filename>.pgf}
%%
%% Make sure the required packages are loaded in your preamble
%%   \usepackage{pgf}
%%
%% Also ensure that all the required font packages are loaded; for instance,
%% the lmodern package is sometimes necessary when using math font.
%%   \usepackage{lmodern}
%%
%% Figures using additional raster images can only be included by \input if
%% they are in the same directory as the main LaTeX file. For loading figures
%% from other directories you can use the `import` package
%% and then include the figures with
%%   \import{<path to file>}{<filename>.pgf}
%%
%% Matplotlib used the following preamble
%%
\begingroup%
\makeatletter%
\begin{pgfpicture}%
\pgfpathrectangle{\pgfpointorigin}{\pgfqpoint{5.285820in}{6.656060in}}%
\pgfusepath{use as bounding box, clip}%
\begin{pgfscope}%
\pgfsetbuttcap%
\pgfsetmiterjoin%
\definecolor{currentfill}{rgb}{1.000000,1.000000,1.000000}%
\pgfsetfillcolor{currentfill}%
\pgfsetlinewidth{0.000000pt}%
\definecolor{currentstroke}{rgb}{1.000000,1.000000,1.000000}%
\pgfsetstrokecolor{currentstroke}%
\pgfsetdash{}{0pt}%
\pgfpathmoveto{\pgfqpoint{0.000000in}{0.000000in}}%
\pgfpathlineto{\pgfqpoint{5.285820in}{0.000000in}}%
\pgfpathlineto{\pgfqpoint{5.285820in}{6.656060in}}%
\pgfpathlineto{\pgfqpoint{0.000000in}{6.656060in}}%
\pgfpathlineto{\pgfqpoint{0.000000in}{0.000000in}}%
\pgfpathclose%
\pgfusepath{fill}%
\end{pgfscope}%
\begin{pgfscope}%
\pgfpathrectangle{\pgfqpoint{0.271449in}{5.369984in}}{\pgfqpoint{0.987003in}{0.987003in}}%
\pgfusepath{clip}%
\pgfsys@transformshift{0.271449in}{5.369984in}%
\pgftext[left,bottom]{\includegraphics[interpolate=true,width=0.990000in,height=0.990000in]{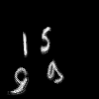}}%
\end{pgfscope}%
\begin{pgfscope}%
\definecolor{textcolor}{rgb}{0.000000,0.000000,0.000000}%
\pgfsetstrokecolor{textcolor}%
\pgfsetfillcolor{textcolor}%
\pgftext[x=0.764950in,y=6.440320in,,base]{\color{textcolor}\rmfamily\fontsize{12.000000}{14.400000}\selectfont Input image}%
\end{pgfscope}%
\begin{pgfscope}%
\pgfpathrectangle{\pgfqpoint{1.565362in}{5.369984in}}{\pgfqpoint{0.987003in}{0.987003in}}%
\pgfusepath{clip}%
\pgfsys@transformcm{0.990000}{0.000000}{0.000000}{-0.990000}{1.565362in}{6.359984in}%
\pgftext[left,bottom]{\includegraphics[interpolate=false,width=1.000000in,height=1.000000in]{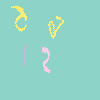}}%
\end{pgfscope}%
\begin{pgfscope}%
\definecolor{textcolor}{rgb}{0.000000,0.000000,0.000000}%
\pgfsetstrokecolor{textcolor}%
\pgfsetfillcolor{textcolor}%
\pgftext[x=2.058863in,y=6.440320in,,base]{\color{textcolor}\rmfamily\fontsize{12.000000}{14.400000}\selectfont Ground truth mask}%
\end{pgfscope}%
\begin{pgfscope}%
\pgfpathrectangle{\pgfqpoint{2.859275in}{5.369984in}}{\pgfqpoint{0.987003in}{0.987003in}}%
\pgfusepath{clip}%
\pgfsys@transformcm{0.990000}{0.000000}{0.000000}{-0.990000}{2.859275in}{6.359984in}%
\pgftext[left,bottom]{\includegraphics[interpolate=false,width=1.000000in,height=1.000000in]{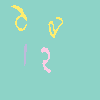}}%
\end{pgfscope}%
\begin{pgfscope}%
\definecolor{textcolor}{rgb}{0.000000,0.000000,0.000000}%
\pgfsetstrokecolor{textcolor}%
\pgfsetfillcolor{textcolor}%
\pgftext[x=3.352776in,y=6.440320in,,base]{\color{textcolor}\rmfamily\fontsize{12.000000}{14.400000}\selectfont 204k S2CNN}%
\end{pgfscope}%
\begin{pgfscope}%
\pgfpathrectangle{\pgfqpoint{4.153188in}{5.369984in}}{\pgfqpoint{0.987003in}{0.987003in}}%
\pgfusepath{clip}%
\pgfsys@transformcm{0.990000}{0.000000}{0.000000}{-0.990000}{4.153188in}{6.359984in}%
\pgftext[left,bottom]{\includegraphics[interpolate=false,width=1.000000in,height=1.000000in]{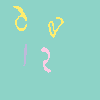}}%
\end{pgfscope}%
\begin{pgfscope}%
\definecolor{textcolor}{rgb}{0.000000,0.000000,0.000000}%
\pgfsetstrokecolor{textcolor}%
\pgfsetfillcolor{textcolor}%
\pgftext[x=4.646689in,y=6.440320in,,base]{\color{textcolor}\rmfamily\fontsize{12.000000}{14.400000}\selectfont 1M CNN}%
\end{pgfscope}%
\begin{pgfscope}%
\pgfpathrectangle{\pgfqpoint{0.271449in}{4.185581in}}{\pgfqpoint{0.987003in}{0.987003in}}%
\pgfusepath{clip}%
\pgfsys@transformshift{0.271449in}{4.185581in}%
\pgftext[left,bottom]{\includegraphics[interpolate=true,width=0.990000in,height=0.990000in]{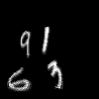}}%
\end{pgfscope}%
\begin{pgfscope}%
\pgfpathrectangle{\pgfqpoint{1.565362in}{4.185581in}}{\pgfqpoint{0.987003in}{0.987003in}}%
\pgfusepath{clip}%
\pgfsys@transformcm{0.990000}{0.000000}{0.000000}{-0.990000}{1.565362in}{5.175581in}%
\pgftext[left,bottom]{\includegraphics[interpolate=false,width=1.000000in,height=1.000000in]{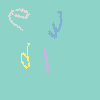}}%
\end{pgfscope}%
\begin{pgfscope}%
\pgfpathrectangle{\pgfqpoint{2.859275in}{4.185581in}}{\pgfqpoint{0.987003in}{0.987003in}}%
\pgfusepath{clip}%
\pgfsys@transformcm{0.990000}{0.000000}{0.000000}{-0.990000}{2.859275in}{5.175581in}%
\pgftext[left,bottom]{\includegraphics[interpolate=false,width=1.000000in,height=1.000000in]{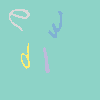}}%
\end{pgfscope}%
\begin{pgfscope}%
\pgfpathrectangle{\pgfqpoint{4.153188in}{4.185581in}}{\pgfqpoint{0.987003in}{0.987003in}}%
\pgfusepath{clip}%
\pgfsys@transformcm{0.990000}{0.000000}{0.000000}{-0.990000}{4.153188in}{5.175581in}%
\pgftext[left,bottom]{\includegraphics[interpolate=false,width=1.000000in,height=1.000000in]{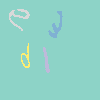}}%
\end{pgfscope}%
\begin{pgfscope}%
\pgfpathrectangle{\pgfqpoint{0.271449in}{3.001178in}}{\pgfqpoint{0.987003in}{0.987003in}}%
\pgfusepath{clip}%
\pgfsys@transformshift{0.271449in}{3.001178in}%
\pgftext[left,bottom]{\includegraphics[interpolate=true,width=0.990000in,height=0.990000in]{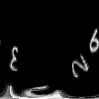}}%
\end{pgfscope}%
\begin{pgfscope}%
\pgfpathrectangle{\pgfqpoint{1.565362in}{3.001178in}}{\pgfqpoint{0.987003in}{0.987003in}}%
\pgfusepath{clip}%
\pgfsys@transformcm{0.990000}{0.000000}{0.000000}{-0.990000}{1.565362in}{3.991178in}%
\pgftext[left,bottom]{\includegraphics[interpolate=false,width=1.000000in,height=1.000000in]{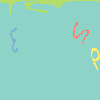}}%
\end{pgfscope}%
\begin{pgfscope}%
\pgfpathrectangle{\pgfqpoint{2.859275in}{3.001178in}}{\pgfqpoint{0.987003in}{0.987003in}}%
\pgfusepath{clip}%
\pgfsys@transformcm{0.990000}{0.000000}{0.000000}{-0.990000}{2.859275in}{3.991178in}%
\pgftext[left,bottom]{\includegraphics[interpolate=false,width=1.000000in,height=1.000000in]{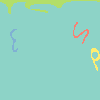}}%
\end{pgfscope}%
\begin{pgfscope}%
\pgfpathrectangle{\pgfqpoint{4.153188in}{3.001178in}}{\pgfqpoint{0.987003in}{0.987003in}}%
\pgfusepath{clip}%
\pgfsys@transformcm{0.990000}{0.000000}{0.000000}{-0.990000}{4.153188in}{3.991178in}%
\pgftext[left,bottom]{\includegraphics[interpolate=false,width=1.000000in,height=1.000000in]{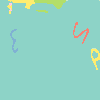}}%
\end{pgfscope}%
\begin{pgfscope}%
\pgfpathrectangle{\pgfqpoint{0.271449in}{1.816774in}}{\pgfqpoint{0.987003in}{0.987003in}}%
\pgfusepath{clip}%
\pgfsys@transformshift{0.271449in}{1.816774in}%
\pgftext[left,bottom]{\includegraphics[interpolate=true,width=0.990000in,height=0.990000in]{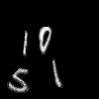}}%
\end{pgfscope}%
\begin{pgfscope}%
\pgfpathrectangle{\pgfqpoint{1.565362in}{1.816774in}}{\pgfqpoint{0.987003in}{0.987003in}}%
\pgfusepath{clip}%
\pgfsys@transformcm{0.990000}{0.000000}{0.000000}{-0.990000}{1.565362in}{2.806774in}%
\pgftext[left,bottom]{\includegraphics[interpolate=false,width=1.000000in,height=1.000000in]{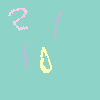}}%
\end{pgfscope}%
\begin{pgfscope}%
\pgfpathrectangle{\pgfqpoint{2.859275in}{1.816774in}}{\pgfqpoint{0.987003in}{0.987003in}}%
\pgfusepath{clip}%
\pgfsys@transformcm{0.990000}{0.000000}{0.000000}{-0.990000}{2.859275in}{2.806774in}%
\pgftext[left,bottom]{\includegraphics[interpolate=false,width=1.000000in,height=1.000000in]{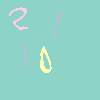}}%
\end{pgfscope}%
\begin{pgfscope}%
\pgfpathrectangle{\pgfqpoint{4.153188in}{1.816774in}}{\pgfqpoint{0.987003in}{0.987003in}}%
\pgfusepath{clip}%
\pgfsys@transformcm{0.990000}{0.000000}{0.000000}{-0.990000}{4.153188in}{2.806774in}%
\pgftext[left,bottom]{\includegraphics[interpolate=false,width=1.000000in,height=1.000000in]{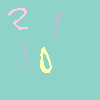}}%
\end{pgfscope}%
\begin{pgfscope}%
\pgfpathrectangle{\pgfqpoint{0.271449in}{0.632371in}}{\pgfqpoint{0.987003in}{0.987003in}}%
\pgfusepath{clip}%
\pgfsys@transformshift{0.271449in}{0.632371in}%
\pgftext[left,bottom]{\includegraphics[interpolate=true,width=0.990000in,height=0.990000in]{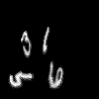}}%
\end{pgfscope}%
\begin{pgfscope}%
\pgfpathrectangle{\pgfqpoint{1.565362in}{0.632371in}}{\pgfqpoint{0.987003in}{0.987003in}}%
\pgfusepath{clip}%
\pgfsys@transformcm{0.990000}{0.000000}{0.000000}{-0.990000}{1.565362in}{1.622371in}%
\pgftext[left,bottom]{\includegraphics[interpolate=false,width=1.000000in,height=1.000000in]{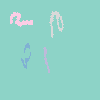}}%
\end{pgfscope}%
\begin{pgfscope}%
\pgfpathrectangle{\pgfqpoint{2.859275in}{0.632371in}}{\pgfqpoint{0.987003in}{0.987003in}}%
\pgfusepath{clip}%
\pgfsys@transformcm{0.990000}{0.000000}{0.000000}{-0.990000}{2.859275in}{1.622371in}%
\pgftext[left,bottom]{\includegraphics[interpolate=false,width=1.000000in,height=1.000000in]{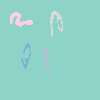}}%
\end{pgfscope}%
\begin{pgfscope}%
\pgfpathrectangle{\pgfqpoint{4.153188in}{0.632371in}}{\pgfqpoint{0.987003in}{0.987003in}}%
\pgfusepath{clip}%
\pgfsys@transformcm{0.990000}{0.000000}{0.000000}{-0.990000}{4.153188in}{1.622371in}%
\pgftext[left,bottom]{\includegraphics[interpolate=false,width=1.000000in,height=1.000000in]{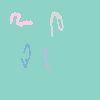}}%
\end{pgfscope}%
\begin{pgfscope}%
\pgfsetbuttcap%
\pgfsetmiterjoin%
\definecolor{currentfill}{rgb}{1.000000,1.000000,1.000000}%
\pgfsetfillcolor{currentfill}%
\pgfsetlinewidth{0.000000pt}%
\definecolor{currentstroke}{rgb}{0.000000,0.000000,0.000000}%
\pgfsetstrokecolor{currentstroke}%
\pgfsetstrokeopacity{0.000000}%
\pgfsetdash{}{0pt}%
\pgfpathmoveto{\pgfqpoint{0.225820in}{0.320679in}}%
\pgfpathlineto{\pgfqpoint{5.185820in}{0.320679in}}%
\pgfpathlineto{\pgfqpoint{5.185820in}{0.444679in}}%
\pgfpathlineto{\pgfqpoint{0.225820in}{0.444679in}}%
\pgfpathlineto{\pgfqpoint{0.225820in}{0.320679in}}%
\pgfpathclose%
\pgfusepath{fill}%
\end{pgfscope}%
\begin{pgfscope}%
\pgfpathrectangle{\pgfqpoint{0.225820in}{0.320679in}}{\pgfqpoint{4.960000in}{0.124000in}}%
\pgfusepath{clip}%
\pgfsetbuttcap%
\pgfsetmiterjoin%
\definecolor{currentfill}{rgb}{1.000000,1.000000,1.000000}%
\pgfsetfillcolor{currentfill}%
\pgfsetlinewidth{0.010037pt}%
\definecolor{currentstroke}{rgb}{1.000000,1.000000,1.000000}%
\pgfsetstrokecolor{currentstroke}%
\pgfsetdash{}{0pt}%
\pgfusepath{stroke,fill}%
\end{pgfscope}%
\begin{pgfscope}%
\pgfpathrectangle{\pgfqpoint{0.225820in}{0.320679in}}{\pgfqpoint{4.960000in}{0.124000in}}%
\pgfusepath{clip}%
\pgfsetbuttcap%
\pgfsetroundjoin%
\definecolor{currentfill}{rgb}{0.552941,0.827451,0.780392}%
\pgfsetfillcolor{currentfill}%
\pgfsetlinewidth{0.000000pt}%
\definecolor{currentstroke}{rgb}{0.000000,0.000000,0.000000}%
\pgfsetstrokecolor{currentstroke}%
\pgfsetdash{}{0pt}%
\pgfpathmoveto{\pgfqpoint{0.225820in}{0.320679in}}%
\pgfpathlineto{\pgfqpoint{0.225820in}{0.444679in}}%
\pgfpathlineto{\pgfqpoint{0.676729in}{0.444679in}}%
\pgfpathlineto{\pgfqpoint{0.676729in}{0.320679in}}%
\pgfpathlineto{\pgfqpoint{0.225820in}{0.320679in}}%
\pgfusepath{fill}%
\end{pgfscope}%
\begin{pgfscope}%
\pgfpathrectangle{\pgfqpoint{0.225820in}{0.320679in}}{\pgfqpoint{4.960000in}{0.124000in}}%
\pgfusepath{clip}%
\pgfsetbuttcap%
\pgfsetroundjoin%
\definecolor{currentfill}{rgb}{1.000000,1.000000,0.701961}%
\pgfsetfillcolor{currentfill}%
\pgfsetlinewidth{0.000000pt}%
\definecolor{currentstroke}{rgb}{0.000000,0.000000,0.000000}%
\pgfsetstrokecolor{currentstroke}%
\pgfsetdash{}{0pt}%
\pgfpathmoveto{\pgfqpoint{0.676729in}{0.320679in}}%
\pgfpathlineto{\pgfqpoint{0.676729in}{0.444679in}}%
\pgfpathlineto{\pgfqpoint{1.127638in}{0.444679in}}%
\pgfpathlineto{\pgfqpoint{1.127638in}{0.320679in}}%
\pgfpathlineto{\pgfqpoint{0.676729in}{0.320679in}}%
\pgfusepath{fill}%
\end{pgfscope}%
\begin{pgfscope}%
\pgfpathrectangle{\pgfqpoint{0.225820in}{0.320679in}}{\pgfqpoint{4.960000in}{0.124000in}}%
\pgfusepath{clip}%
\pgfsetbuttcap%
\pgfsetroundjoin%
\definecolor{currentfill}{rgb}{0.745098,0.729412,0.854902}%
\pgfsetfillcolor{currentfill}%
\pgfsetlinewidth{0.000000pt}%
\definecolor{currentstroke}{rgb}{0.000000,0.000000,0.000000}%
\pgfsetstrokecolor{currentstroke}%
\pgfsetdash{}{0pt}%
\pgfpathmoveto{\pgfqpoint{1.127638in}{0.320679in}}%
\pgfpathlineto{\pgfqpoint{1.127638in}{0.444679in}}%
\pgfpathlineto{\pgfqpoint{1.578547in}{0.444679in}}%
\pgfpathlineto{\pgfqpoint{1.578547in}{0.320679in}}%
\pgfpathlineto{\pgfqpoint{1.127638in}{0.320679in}}%
\pgfusepath{fill}%
\end{pgfscope}%
\begin{pgfscope}%
\pgfpathrectangle{\pgfqpoint{0.225820in}{0.320679in}}{\pgfqpoint{4.960000in}{0.124000in}}%
\pgfusepath{clip}%
\pgfsetbuttcap%
\pgfsetroundjoin%
\definecolor{currentfill}{rgb}{0.984314,0.501961,0.447059}%
\pgfsetfillcolor{currentfill}%
\pgfsetlinewidth{0.000000pt}%
\definecolor{currentstroke}{rgb}{0.000000,0.000000,0.000000}%
\pgfsetstrokecolor{currentstroke}%
\pgfsetdash{}{0pt}%
\pgfpathmoveto{\pgfqpoint{1.578547in}{0.320679in}}%
\pgfpathlineto{\pgfqpoint{1.578547in}{0.444679in}}%
\pgfpathlineto{\pgfqpoint{2.029456in}{0.444679in}}%
\pgfpathlineto{\pgfqpoint{2.029456in}{0.320679in}}%
\pgfpathlineto{\pgfqpoint{1.578547in}{0.320679in}}%
\pgfusepath{fill}%
\end{pgfscope}%
\begin{pgfscope}%
\pgfpathrectangle{\pgfqpoint{0.225820in}{0.320679in}}{\pgfqpoint{4.960000in}{0.124000in}}%
\pgfusepath{clip}%
\pgfsetbuttcap%
\pgfsetroundjoin%
\definecolor{currentfill}{rgb}{0.501961,0.694118,0.827451}%
\pgfsetfillcolor{currentfill}%
\pgfsetlinewidth{0.000000pt}%
\definecolor{currentstroke}{rgb}{0.000000,0.000000,0.000000}%
\pgfsetstrokecolor{currentstroke}%
\pgfsetdash{}{0pt}%
\pgfpathmoveto{\pgfqpoint{2.029456in}{0.320679in}}%
\pgfpathlineto{\pgfqpoint{2.029456in}{0.444679in}}%
\pgfpathlineto{\pgfqpoint{2.480365in}{0.444679in}}%
\pgfpathlineto{\pgfqpoint{2.480365in}{0.320679in}}%
\pgfpathlineto{\pgfqpoint{2.029456in}{0.320679in}}%
\pgfusepath{fill}%
\end{pgfscope}%
\begin{pgfscope}%
\pgfpathrectangle{\pgfqpoint{0.225820in}{0.320679in}}{\pgfqpoint{4.960000in}{0.124000in}}%
\pgfusepath{clip}%
\pgfsetbuttcap%
\pgfsetroundjoin%
\definecolor{currentfill}{rgb}{0.701961,0.870588,0.411765}%
\pgfsetfillcolor{currentfill}%
\pgfsetlinewidth{0.000000pt}%
\definecolor{currentstroke}{rgb}{0.000000,0.000000,0.000000}%
\pgfsetstrokecolor{currentstroke}%
\pgfsetdash{}{0pt}%
\pgfpathmoveto{\pgfqpoint{2.480365in}{0.320679in}}%
\pgfpathlineto{\pgfqpoint{2.480365in}{0.444679in}}%
\pgfpathlineto{\pgfqpoint{2.931274in}{0.444679in}}%
\pgfpathlineto{\pgfqpoint{2.931274in}{0.320679in}}%
\pgfpathlineto{\pgfqpoint{2.480365in}{0.320679in}}%
\pgfusepath{fill}%
\end{pgfscope}%
\begin{pgfscope}%
\pgfpathrectangle{\pgfqpoint{0.225820in}{0.320679in}}{\pgfqpoint{4.960000in}{0.124000in}}%
\pgfusepath{clip}%
\pgfsetbuttcap%
\pgfsetroundjoin%
\definecolor{currentfill}{rgb}{0.988235,0.803922,0.898039}%
\pgfsetfillcolor{currentfill}%
\pgfsetlinewidth{0.000000pt}%
\definecolor{currentstroke}{rgb}{0.000000,0.000000,0.000000}%
\pgfsetstrokecolor{currentstroke}%
\pgfsetdash{}{0pt}%
\pgfpathmoveto{\pgfqpoint{2.931274in}{0.320679in}}%
\pgfpathlineto{\pgfqpoint{2.931274in}{0.444679in}}%
\pgfpathlineto{\pgfqpoint{3.382183in}{0.444679in}}%
\pgfpathlineto{\pgfqpoint{3.382183in}{0.320679in}}%
\pgfpathlineto{\pgfqpoint{2.931274in}{0.320679in}}%
\pgfusepath{fill}%
\end{pgfscope}%
\begin{pgfscope}%
\pgfpathrectangle{\pgfqpoint{0.225820in}{0.320679in}}{\pgfqpoint{4.960000in}{0.124000in}}%
\pgfusepath{clip}%
\pgfsetbuttcap%
\pgfsetroundjoin%
\definecolor{currentfill}{rgb}{0.850980,0.850980,0.850980}%
\pgfsetfillcolor{currentfill}%
\pgfsetlinewidth{0.000000pt}%
\definecolor{currentstroke}{rgb}{0.000000,0.000000,0.000000}%
\pgfsetstrokecolor{currentstroke}%
\pgfsetdash{}{0pt}%
\pgfpathmoveto{\pgfqpoint{3.382183in}{0.320679in}}%
\pgfpathlineto{\pgfqpoint{3.382183in}{0.444679in}}%
\pgfpathlineto{\pgfqpoint{3.833092in}{0.444679in}}%
\pgfpathlineto{\pgfqpoint{3.833092in}{0.320679in}}%
\pgfpathlineto{\pgfqpoint{3.382183in}{0.320679in}}%
\pgfusepath{fill}%
\end{pgfscope}%
\begin{pgfscope}%
\pgfpathrectangle{\pgfqpoint{0.225820in}{0.320679in}}{\pgfqpoint{4.960000in}{0.124000in}}%
\pgfusepath{clip}%
\pgfsetbuttcap%
\pgfsetroundjoin%
\definecolor{currentfill}{rgb}{0.737255,0.501961,0.741176}%
\pgfsetfillcolor{currentfill}%
\pgfsetlinewidth{0.000000pt}%
\definecolor{currentstroke}{rgb}{0.000000,0.000000,0.000000}%
\pgfsetstrokecolor{currentstroke}%
\pgfsetdash{}{0pt}%
\pgfpathmoveto{\pgfqpoint{3.833092in}{0.320679in}}%
\pgfpathlineto{\pgfqpoint{3.833092in}{0.444679in}}%
\pgfpathlineto{\pgfqpoint{4.284001in}{0.444679in}}%
\pgfpathlineto{\pgfqpoint{4.284001in}{0.320679in}}%
\pgfpathlineto{\pgfqpoint{3.833092in}{0.320679in}}%
\pgfusepath{fill}%
\end{pgfscope}%
\begin{pgfscope}%
\pgfpathrectangle{\pgfqpoint{0.225820in}{0.320679in}}{\pgfqpoint{4.960000in}{0.124000in}}%
\pgfusepath{clip}%
\pgfsetbuttcap%
\pgfsetroundjoin%
\definecolor{currentfill}{rgb}{0.800000,0.921569,0.772549}%
\pgfsetfillcolor{currentfill}%
\pgfsetlinewidth{0.000000pt}%
\definecolor{currentstroke}{rgb}{0.000000,0.000000,0.000000}%
\pgfsetstrokecolor{currentstroke}%
\pgfsetdash{}{0pt}%
\pgfpathmoveto{\pgfqpoint{4.284001in}{0.320679in}}%
\pgfpathlineto{\pgfqpoint{4.284001in}{0.444679in}}%
\pgfpathlineto{\pgfqpoint{4.734910in}{0.444679in}}%
\pgfpathlineto{\pgfqpoint{4.734910in}{0.320679in}}%
\pgfpathlineto{\pgfqpoint{4.284001in}{0.320679in}}%
\pgfusepath{fill}%
\end{pgfscope}%
\begin{pgfscope}%
\pgfpathrectangle{\pgfqpoint{0.225820in}{0.320679in}}{\pgfqpoint{4.960000in}{0.124000in}}%
\pgfusepath{clip}%
\pgfsetbuttcap%
\pgfsetroundjoin%
\definecolor{currentfill}{rgb}{1.000000,0.929412,0.435294}%
\pgfsetfillcolor{currentfill}%
\pgfsetlinewidth{0.000000pt}%
\definecolor{currentstroke}{rgb}{0.000000,0.000000,0.000000}%
\pgfsetstrokecolor{currentstroke}%
\pgfsetdash{}{0pt}%
\pgfpathmoveto{\pgfqpoint{4.734910in}{0.320679in}}%
\pgfpathlineto{\pgfqpoint{4.734910in}{0.444679in}}%
\pgfpathlineto{\pgfqpoint{5.185820in}{0.444679in}}%
\pgfpathlineto{\pgfqpoint{5.185820in}{0.320679in}}%
\pgfpathlineto{\pgfqpoint{4.734910in}{0.320679in}}%
\pgfusepath{fill}%
\end{pgfscope}%
\begin{pgfscope}%
\pgfsetbuttcap%
\pgfsetroundjoin%
\definecolor{currentfill}{rgb}{0.000000,0.000000,0.000000}%
\pgfsetfillcolor{currentfill}%
\pgfsetlinewidth{0.803000pt}%
\definecolor{currentstroke}{rgb}{0.000000,0.000000,0.000000}%
\pgfsetstrokecolor{currentstroke}%
\pgfsetdash{}{0pt}%
\pgfsys@defobject{currentmarker}{\pgfqpoint{0.000000in}{-0.048611in}}{\pgfqpoint{0.000000in}{0.000000in}}{%
\pgfpathmoveto{\pgfqpoint{0.000000in}{0.000000in}}%
\pgfpathlineto{\pgfqpoint{0.000000in}{-0.048611in}}%
\pgfusepath{stroke,fill}%
}%
\begin{pgfscope}%
\pgfsys@transformshift{0.451274in}{0.320679in}%
\pgfsys@useobject{currentmarker}{}%
\end{pgfscope}%
\end{pgfscope}%
\begin{pgfscope}%
\definecolor{textcolor}{rgb}{0.000000,0.000000,0.000000}%
\pgfsetstrokecolor{textcolor}%
\pgfsetfillcolor{textcolor}%
\pgftext[x=0.451274in,y=0.223457in,,top]{\color{textcolor}\rmfamily\fontsize{10.000000}{12.000000}\selectfont background}%
\end{pgfscope}%
\begin{pgfscope}%
\pgfsetbuttcap%
\pgfsetroundjoin%
\definecolor{currentfill}{rgb}{0.000000,0.000000,0.000000}%
\pgfsetfillcolor{currentfill}%
\pgfsetlinewidth{0.803000pt}%
\definecolor{currentstroke}{rgb}{0.000000,0.000000,0.000000}%
\pgfsetstrokecolor{currentstroke}%
\pgfsetdash{}{0pt}%
\pgfsys@defobject{currentmarker}{\pgfqpoint{0.000000in}{-0.048611in}}{\pgfqpoint{0.000000in}{0.000000in}}{%
\pgfpathmoveto{\pgfqpoint{0.000000in}{0.000000in}}%
\pgfpathlineto{\pgfqpoint{0.000000in}{-0.048611in}}%
\pgfusepath{stroke,fill}%
}%
\begin{pgfscope}%
\pgfsys@transformshift{0.902183in}{0.320679in}%
\pgfsys@useobject{currentmarker}{}%
\end{pgfscope}%
\end{pgfscope}%
\begin{pgfscope}%
\definecolor{textcolor}{rgb}{0.000000,0.000000,0.000000}%
\pgfsetstrokecolor{textcolor}%
\pgfsetfillcolor{textcolor}%
\pgftext[x=0.902183in,y=0.223457in,,top]{\color{textcolor}\rmfamily\fontsize{10.000000}{12.000000}\selectfont 0}%
\end{pgfscope}%
\begin{pgfscope}%
\pgfsetbuttcap%
\pgfsetroundjoin%
\definecolor{currentfill}{rgb}{0.000000,0.000000,0.000000}%
\pgfsetfillcolor{currentfill}%
\pgfsetlinewidth{0.803000pt}%
\definecolor{currentstroke}{rgb}{0.000000,0.000000,0.000000}%
\pgfsetstrokecolor{currentstroke}%
\pgfsetdash{}{0pt}%
\pgfsys@defobject{currentmarker}{\pgfqpoint{0.000000in}{-0.048611in}}{\pgfqpoint{0.000000in}{0.000000in}}{%
\pgfpathmoveto{\pgfqpoint{0.000000in}{0.000000in}}%
\pgfpathlineto{\pgfqpoint{0.000000in}{-0.048611in}}%
\pgfusepath{stroke,fill}%
}%
\begin{pgfscope}%
\pgfsys@transformshift{1.353092in}{0.320679in}%
\pgfsys@useobject{currentmarker}{}%
\end{pgfscope}%
\end{pgfscope}%
\begin{pgfscope}%
\definecolor{textcolor}{rgb}{0.000000,0.000000,0.000000}%
\pgfsetstrokecolor{textcolor}%
\pgfsetfillcolor{textcolor}%
\pgftext[x=1.353092in,y=0.223457in,,top]{\color{textcolor}\rmfamily\fontsize{10.000000}{12.000000}\selectfont 1}%
\end{pgfscope}%
\begin{pgfscope}%
\pgfsetbuttcap%
\pgfsetroundjoin%
\definecolor{currentfill}{rgb}{0.000000,0.000000,0.000000}%
\pgfsetfillcolor{currentfill}%
\pgfsetlinewidth{0.803000pt}%
\definecolor{currentstroke}{rgb}{0.000000,0.000000,0.000000}%
\pgfsetstrokecolor{currentstroke}%
\pgfsetdash{}{0pt}%
\pgfsys@defobject{currentmarker}{\pgfqpoint{0.000000in}{-0.048611in}}{\pgfqpoint{0.000000in}{0.000000in}}{%
\pgfpathmoveto{\pgfqpoint{0.000000in}{0.000000in}}%
\pgfpathlineto{\pgfqpoint{0.000000in}{-0.048611in}}%
\pgfusepath{stroke,fill}%
}%
\begin{pgfscope}%
\pgfsys@transformshift{1.804001in}{0.320679in}%
\pgfsys@useobject{currentmarker}{}%
\end{pgfscope}%
\end{pgfscope}%
\begin{pgfscope}%
\definecolor{textcolor}{rgb}{0.000000,0.000000,0.000000}%
\pgfsetstrokecolor{textcolor}%
\pgfsetfillcolor{textcolor}%
\pgftext[x=1.804001in,y=0.223457in,,top]{\color{textcolor}\rmfamily\fontsize{10.000000}{12.000000}\selectfont 2}%
\end{pgfscope}%
\begin{pgfscope}%
\pgfsetbuttcap%
\pgfsetroundjoin%
\definecolor{currentfill}{rgb}{0.000000,0.000000,0.000000}%
\pgfsetfillcolor{currentfill}%
\pgfsetlinewidth{0.803000pt}%
\definecolor{currentstroke}{rgb}{0.000000,0.000000,0.000000}%
\pgfsetstrokecolor{currentstroke}%
\pgfsetdash{}{0pt}%
\pgfsys@defobject{currentmarker}{\pgfqpoint{0.000000in}{-0.048611in}}{\pgfqpoint{0.000000in}{0.000000in}}{%
\pgfpathmoveto{\pgfqpoint{0.000000in}{0.000000in}}%
\pgfpathlineto{\pgfqpoint{0.000000in}{-0.048611in}}%
\pgfusepath{stroke,fill}%
}%
\begin{pgfscope}%
\pgfsys@transformshift{2.254910in}{0.320679in}%
\pgfsys@useobject{currentmarker}{}%
\end{pgfscope}%
\end{pgfscope}%
\begin{pgfscope}%
\definecolor{textcolor}{rgb}{0.000000,0.000000,0.000000}%
\pgfsetstrokecolor{textcolor}%
\pgfsetfillcolor{textcolor}%
\pgftext[x=2.254910in,y=0.223457in,,top]{\color{textcolor}\rmfamily\fontsize{10.000000}{12.000000}\selectfont 3}%
\end{pgfscope}%
\begin{pgfscope}%
\pgfsetbuttcap%
\pgfsetroundjoin%
\definecolor{currentfill}{rgb}{0.000000,0.000000,0.000000}%
\pgfsetfillcolor{currentfill}%
\pgfsetlinewidth{0.803000pt}%
\definecolor{currentstroke}{rgb}{0.000000,0.000000,0.000000}%
\pgfsetstrokecolor{currentstroke}%
\pgfsetdash{}{0pt}%
\pgfsys@defobject{currentmarker}{\pgfqpoint{0.000000in}{-0.048611in}}{\pgfqpoint{0.000000in}{0.000000in}}{%
\pgfpathmoveto{\pgfqpoint{0.000000in}{0.000000in}}%
\pgfpathlineto{\pgfqpoint{0.000000in}{-0.048611in}}%
\pgfusepath{stroke,fill}%
}%
\begin{pgfscope}%
\pgfsys@transformshift{2.705820in}{0.320679in}%
\pgfsys@useobject{currentmarker}{}%
\end{pgfscope}%
\end{pgfscope}%
\begin{pgfscope}%
\definecolor{textcolor}{rgb}{0.000000,0.000000,0.000000}%
\pgfsetstrokecolor{textcolor}%
\pgfsetfillcolor{textcolor}%
\pgftext[x=2.705820in,y=0.223457in,,top]{\color{textcolor}\rmfamily\fontsize{10.000000}{12.000000}\selectfont 4}%
\end{pgfscope}%
\begin{pgfscope}%
\pgfsetbuttcap%
\pgfsetroundjoin%
\definecolor{currentfill}{rgb}{0.000000,0.000000,0.000000}%
\pgfsetfillcolor{currentfill}%
\pgfsetlinewidth{0.803000pt}%
\definecolor{currentstroke}{rgb}{0.000000,0.000000,0.000000}%
\pgfsetstrokecolor{currentstroke}%
\pgfsetdash{}{0pt}%
\pgfsys@defobject{currentmarker}{\pgfqpoint{0.000000in}{-0.048611in}}{\pgfqpoint{0.000000in}{0.000000in}}{%
\pgfpathmoveto{\pgfqpoint{0.000000in}{0.000000in}}%
\pgfpathlineto{\pgfqpoint{0.000000in}{-0.048611in}}%
\pgfusepath{stroke,fill}%
}%
\begin{pgfscope}%
\pgfsys@transformshift{3.156729in}{0.320679in}%
\pgfsys@useobject{currentmarker}{}%
\end{pgfscope}%
\end{pgfscope}%
\begin{pgfscope}%
\definecolor{textcolor}{rgb}{0.000000,0.000000,0.000000}%
\pgfsetstrokecolor{textcolor}%
\pgfsetfillcolor{textcolor}%
\pgftext[x=3.156729in,y=0.223457in,,top]{\color{textcolor}\rmfamily\fontsize{10.000000}{12.000000}\selectfont 5}%
\end{pgfscope}%
\begin{pgfscope}%
\pgfsetbuttcap%
\pgfsetroundjoin%
\definecolor{currentfill}{rgb}{0.000000,0.000000,0.000000}%
\pgfsetfillcolor{currentfill}%
\pgfsetlinewidth{0.803000pt}%
\definecolor{currentstroke}{rgb}{0.000000,0.000000,0.000000}%
\pgfsetstrokecolor{currentstroke}%
\pgfsetdash{}{0pt}%
\pgfsys@defobject{currentmarker}{\pgfqpoint{0.000000in}{-0.048611in}}{\pgfqpoint{0.000000in}{0.000000in}}{%
\pgfpathmoveto{\pgfqpoint{0.000000in}{0.000000in}}%
\pgfpathlineto{\pgfqpoint{0.000000in}{-0.048611in}}%
\pgfusepath{stroke,fill}%
}%
\begin{pgfscope}%
\pgfsys@transformshift{3.607638in}{0.320679in}%
\pgfsys@useobject{currentmarker}{}%
\end{pgfscope}%
\end{pgfscope}%
\begin{pgfscope}%
\definecolor{textcolor}{rgb}{0.000000,0.000000,0.000000}%
\pgfsetstrokecolor{textcolor}%
\pgfsetfillcolor{textcolor}%
\pgftext[x=3.607638in,y=0.223457in,,top]{\color{textcolor}\rmfamily\fontsize{10.000000}{12.000000}\selectfont 6}%
\end{pgfscope}%
\begin{pgfscope}%
\pgfsetbuttcap%
\pgfsetroundjoin%
\definecolor{currentfill}{rgb}{0.000000,0.000000,0.000000}%
\pgfsetfillcolor{currentfill}%
\pgfsetlinewidth{0.803000pt}%
\definecolor{currentstroke}{rgb}{0.000000,0.000000,0.000000}%
\pgfsetstrokecolor{currentstroke}%
\pgfsetdash{}{0pt}%
\pgfsys@defobject{currentmarker}{\pgfqpoint{0.000000in}{-0.048611in}}{\pgfqpoint{0.000000in}{0.000000in}}{%
\pgfpathmoveto{\pgfqpoint{0.000000in}{0.000000in}}%
\pgfpathlineto{\pgfqpoint{0.000000in}{-0.048611in}}%
\pgfusepath{stroke,fill}%
}%
\begin{pgfscope}%
\pgfsys@transformshift{4.058547in}{0.320679in}%
\pgfsys@useobject{currentmarker}{}%
\end{pgfscope}%
\end{pgfscope}%
\begin{pgfscope}%
\definecolor{textcolor}{rgb}{0.000000,0.000000,0.000000}%
\pgfsetstrokecolor{textcolor}%
\pgfsetfillcolor{textcolor}%
\pgftext[x=4.058547in,y=0.223457in,,top]{\color{textcolor}\rmfamily\fontsize{10.000000}{12.000000}\selectfont 7}%
\end{pgfscope}%
\begin{pgfscope}%
\pgfsetbuttcap%
\pgfsetroundjoin%
\definecolor{currentfill}{rgb}{0.000000,0.000000,0.000000}%
\pgfsetfillcolor{currentfill}%
\pgfsetlinewidth{0.803000pt}%
\definecolor{currentstroke}{rgb}{0.000000,0.000000,0.000000}%
\pgfsetstrokecolor{currentstroke}%
\pgfsetdash{}{0pt}%
\pgfsys@defobject{currentmarker}{\pgfqpoint{0.000000in}{-0.048611in}}{\pgfqpoint{0.000000in}{0.000000in}}{%
\pgfpathmoveto{\pgfqpoint{0.000000in}{0.000000in}}%
\pgfpathlineto{\pgfqpoint{0.000000in}{-0.048611in}}%
\pgfusepath{stroke,fill}%
}%
\begin{pgfscope}%
\pgfsys@transformshift{4.509456in}{0.320679in}%
\pgfsys@useobject{currentmarker}{}%
\end{pgfscope}%
\end{pgfscope}%
\begin{pgfscope}%
\definecolor{textcolor}{rgb}{0.000000,0.000000,0.000000}%
\pgfsetstrokecolor{textcolor}%
\pgfsetfillcolor{textcolor}%
\pgftext[x=4.509456in,y=0.223457in,,top]{\color{textcolor}\rmfamily\fontsize{10.000000}{12.000000}\selectfont 8}%
\end{pgfscope}%
\begin{pgfscope}%
\pgfsetbuttcap%
\pgfsetroundjoin%
\definecolor{currentfill}{rgb}{0.000000,0.000000,0.000000}%
\pgfsetfillcolor{currentfill}%
\pgfsetlinewidth{0.803000pt}%
\definecolor{currentstroke}{rgb}{0.000000,0.000000,0.000000}%
\pgfsetstrokecolor{currentstroke}%
\pgfsetdash{}{0pt}%
\pgfsys@defobject{currentmarker}{\pgfqpoint{0.000000in}{-0.048611in}}{\pgfqpoint{0.000000in}{0.000000in}}{%
\pgfpathmoveto{\pgfqpoint{0.000000in}{0.000000in}}%
\pgfpathlineto{\pgfqpoint{0.000000in}{-0.048611in}}%
\pgfusepath{stroke,fill}%
}%
\begin{pgfscope}%
\pgfsys@transformshift{4.960365in}{0.320679in}%
\pgfsys@useobject{currentmarker}{}%
\end{pgfscope}%
\end{pgfscope}%
\begin{pgfscope}%
\definecolor{textcolor}{rgb}{0.000000,0.000000,0.000000}%
\pgfsetstrokecolor{textcolor}%
\pgfsetfillcolor{textcolor}%
\pgftext[x=4.960365in,y=0.223457in,,top]{\color{textcolor}\rmfamily\fontsize{10.000000}{12.000000}\selectfont 9}%
\end{pgfscope}%
\begin{pgfscope}%
\pgfsetrectcap%
\pgfsetmiterjoin%
\pgfsetlinewidth{0.803000pt}%
\definecolor{currentstroke}{rgb}{0.000000,0.000000,0.000000}%
\pgfsetstrokecolor{currentstroke}%
\pgfsetdash{}{0pt}%
\pgfpathmoveto{\pgfqpoint{0.225820in}{0.320679in}}%
\pgfpathlineto{\pgfqpoint{0.225820in}{0.382679in}}%
\pgfpathlineto{\pgfqpoint{0.225820in}{0.444679in}}%
\pgfpathlineto{\pgfqpoint{5.185820in}{0.444679in}}%
\pgfpathlineto{\pgfqpoint{5.185820in}{0.382679in}}%
\pgfpathlineto{\pgfqpoint{5.185820in}{0.320679in}}%
\pgfpathlineto{\pgfqpoint{0.225820in}{0.320679in}}%
\pgfpathclose%
\pgfusepath{stroke}%
\end{pgfscope}%
\end{pgfpicture}%
\makeatother%
\endgroup%

%% file: appendix_profiling.tex
\section{Inference latency profiling}
\label{sec:latency_profiling}

Table~\ref{tab:latency_per_layer} shows the inference latency per layer for the equivariant semantic segmentation model. Here the bulk of the time is spent in the upsampling layers and in particular the last $\SO(3)$ to $S^{2}$ convolution takes up almost half of the total inference time. Comparing with Table~\ref{tab:equiv_models} this also coincides with the largest $\SO(3)$ tensors being processed.

In Table~\ref{tab:latency_per_op} the fraction of inference time is instead shown per operation. Here it is clear that a large portion is spent in the custom CUDA implementation for the complex matrix multiplication of $\SO{3}$ tensors, a possible avenue for optimization reducing overall inference latency. Most of the remaining time is spent in the Fast Fourier Transform (SO3\_fft\_real) and the inverse FFT (SO3\_ifft\_real).

\begin{table}[t!]
    \centering
        \caption{Latency per layer and fraction of total time for the equivariant semantic segmentation model (204k S2CNN in Appendix Table~\ref{tab:equiv_models}) on an Nvidia T4 16GB GPU as measured by NSight systems with NVTX for batch size 1. Latency measures the time of a forward pass through the layer on the GPU.}
    \label{tab:latency_per_layer}
    \begin{tabular}{llrr}
          & Layer & Latency (ms) &  Fraction (\%) \\
         \hline
  Down & S2SO3conv &  6.7 &       5.0 \\
Down & SO3conv & 16.1 &      11.8 \\
Down & SO3conv &  7.3 &       5.3 \\
Down & SO3conv &  3.1 &       2.3 \\
Up & SO3conv &  6.8 &       5.0 \\
Up & SO3conv & 15.3 &      11.2 \\
Up & SO3conv & 26.4 &      19.4 \\
 Up & SO3S2conv & 54.5 &      40.0 \\
\end{tabular}
\end{table}

\begin{table}[t!]
    \centering
    \caption{Fraction of time spent per operation ($>1\%$) for the equivariant semantic segmentation model (204k S2CNN in Appendix Table~\ref{tab:equiv_models}) on an Nvidia T4 16GB GPU as measured by NSight systems for batch size 1.}
    \label{tab:latency_per_op}
    \begin{tabular}{llr}
           Module & Op & Fraction of time (\%) \\
           \hline
s2cnn &  \_cuda\_SO3\_mm &       45.2 \\
s2cnn & SO3\_ifft\_real &       15.6 \\
s2cnn &  SO3\_fft\_real &       14.4 \\
pytorch &       einsum &       12.2 \\
pytorch &   contiguous &        8.4 \\
%   \rule{0pt}{1ex}
   \hline
   &&  95.7 \\
\end{tabular}
\end{table}